\PassOptionsToPackage{monochrome}{xcolor}
\documentclass[authoryear, review,12pt]{elsarticle}

\usepackage{amssymb}
\usepackage{amsmath}
\usepackage{makecell}
\usepackage{longtable}
\usepackage{adjustbox}
\usepackage{array,caption,threeparttable}
\usepackage{rotating}
\usepackage{multirow}
\usepackage{xcolor,colortbl}
\usepackage{soul}
\usepackage[labelfont=bf]{caption}
\usepackage[labelsep=period]{caption}
\usepackage{float}
\usepackage[colorlinks=true,linkcolor=blue,citecolor=blue]{hyperref}
\usepackage{cleveref}
\let\oldcref\cref
\def\cref#1{\color{blue}\oldcref{#1}}

\usepackage{fncylab}
\usepackage{booktabs}
\usepackage{rotating}
\usepackage{adjustbox}
\usepackage{calc}

\newcolumntype{S}{>{\centering\arraybackslash}m{0.8cm}}
\newcolumntype{M}{>{\centering\arraybackslash}m{1cm}}
\newcolumntype{G}{>{\centering\arraybackslash}m{1.3cm}}

\usepackage{lineno}
\usepackage{subcaption}

\begin{document}

\begin{frontmatter}

\title{$\mathrm{F^2Depth}$: Self-supervised Indoor Monocular Depth Estimation via Optical Flow Consistency and Feature Map Synthesis}

\author[label1]{Xiaotong Guo}
\ead{guoxiaotong0420@buaa.edu.cn}
\author[label1,label2,label3]{Huijie Zhao\corref{cor1}}
\ead{hjzhao@buaa.edu.cn}
\cortext[cor1]{Corresponding author.}
\author[label4]{Shuwei Shao}
\ead{swshao@buaa.edu.cn}
\author[label1]{Xudong Li}
\ead{xdli@buaa.edu.cn}
\author[label2]{Baochang Zhang}
\ead{bczhang@buaa.edu.cn}
\affiliation[label1]{organization={School of Instrumentation and Optoelectronic Engineering, Key Laboratory of Precision Opto-mechatronics Technology, Ministry of Education, Beihang University},
            addressline={No. 37 Xueyuan Road, Haidian District}, 
            city={Beijing},
            postcode={100191}, 
            country={China}}
\affiliation[label2]{organization={Institute of Artificial Intelligence, Beihang University},
            addressline={No. 37 Xueyuan Road, Haidian District}, 
            city={Beijing},
            postcode={100191}, 
            country={China}}
\affiliation[label3]{organization={Qingdao Research Institute of Beihang University},
            city={Qingdao},
            postcode={266101}, 
            country={China}}
\affiliation[label4]{organization={School of Automation Science and Electrical Engineering, Beihang University},
            addressline={No. 37 Xueyuan Road, Haidian District}, 
            city={Beijing},
            postcode={100191}, 
            country={China}}

\begin{abstract}
Self-supervised monocular depth estimation methods have been increasingly given much attention due to the benefit of not requiring \textcolor{red}{large, labelled} datasets. Such self-supervised methods require high-quality salient features and consequently suffer from severe performance drop for indoor scenes, where \textcolor{red}{low-textured} regions dominant in the scenes are almost indiscriminative. To address the issue, we propose a self-supervised indoor monocular depth estimation framework called $\mathrm{F^2Depth}$. A self-supervised optical flow estimation network is introduced to supervise depth learning. To improve optical flow estimation performance in \textcolor{red}{low-textured} areas, only some patches of points with more discriminative features are adopted for finetuning based on our well-designed patch-based photometric loss. The finetuned optical flow estimation network generates high-accuracy optical flow as a supervisory signal for depth estimation. Correspondingly, an optical flow consistency loss is designed. Multi-scale feature maps produced by finetuned optical flow estimation network perform warping to compute feature map synthesis loss as another supervisory signal for depth learning. Experimental results on the NYU Depth V2 dataset demonstrate the effectiveness of the framework and our proposed losses. To evaluate the generalization ability of our $\mathrm{F^2Depth}$, we collect a Campus Indoor depth dataset composed of approximately 1500 points selected from 99 images in 18 scenes. \textcolor{red}{Zero-shot generalization experiments on 7-Scenes dataset and Campus Indoor achieve $\delta_1$ accuracy of 75.8\% and 76.0\% respectively.} The accuracy results show that our model can generalize well to monocular images captured in unknown indoor scenes.
\end{abstract}

\end{frontmatter}

\section{Introduction}
\textcolor{red}{Depth estimation is an important basis of scene spatial structure perception.} Monocular depth estimation aims to predict scene depths from a single image, with a wide range of applications such as 3D reconstruction, scene understanding, environmental perception, \textcolor{red}{automatic driving, obstacle avoidance and robotic grasps}. Supervised depth estimation methods \citep{Agarwal2023, Patil2022, Yin2019, Yuan2022, Zhang2023} can achieve excellent performance but require massive \textcolor{red}{labelled} data during training. In order to get rid of the dependence on \textcolor{red}{labelled} data, self-supervised methods have attracted increasing attention. Many self-supervised methods \citep{Bian2021_RN52, Garg2016, Godard2017, Godard2019, Ren2017, Wang2018_RN51, Xie2016, Yin2018, Zhou2017} perform well on outdoor datasets such as KITTI \citep{Uhrig2017} and Cityscapes \citep{Cordts2016}. However, these methods cannot achieve satisfactory results on indoor datasets such as NYU Depth V2 \citep{Silberman2012} \textcolor{red}{and 7-Scenes} \citep{Shotton2013}.

A great challenge for self-supervised indoor depth estimation is that there are many low-textured regions such as walls, floors and tables. The main supervisory signal of self-supervised methods is based on view synthesis \citep{Wang2018_RN51} which aims to make the appearance of warped image and that of real image as consistent as possible. Thus, similar pixels lack of salient features can easily cause incorrect matching. It is difficult to predict depths of these regions only depending on photometric loss. $\mathrm{P^2Net}$ \citep{Yu2020} and StructDepth \citep{Li2021} exploited the geometric regularity of low-textured areas to design new loss functions together with the photometric loss as supervision. 

Optical flow describes motion of pixels between images. Accurate optical flow estimation can \textcolor{red}{serve} many higher-level computer vision tasks, such as monocular depth estimation. Unsupervised optical flow estimation methods are trained based on view synthesis, similar to depth estimation. Among existing unsupervised optical flow estimation methods, ARFlow \citep{Liu2020} can achieve a better balance between accuracy and computational complexity. However, it performs poorly in low-textured regions. TrianFlow \citep{Zhao2020} performed simultaneous indoor depth and pose learning using optical flow to estimate camera ego-motion, but achieved limited accuracy. Moving Indoor \citep{Zhou2019} utilized unsupervised optical flow estimation network SF-Net to generate optical flow as supervision for depth learning. Yet all pixels are used to compute photometric loss of SF-Net, resulting in limited performance on \textcolor{red}{low-textured} regions. The overall precision of Moving Indoor \citep{Zhou2019} is also not satisfactory.

We propose a self-supervised depth estimation framework via optical flow learning, called $\mathrm{F^2Depth}$. A large number of virtual datasets are available to pretrain optical flow estimation networks, which facilitates a more accurate pixel motion calculation for supervising depth learning. The optical flow estimation network is first finetuned to supervise depth learning. To improve the performance of optical flow learning on \textcolor{red}{low-textured} regions, a patch-based photometric loss is designed. Patches of points with discriminative features are adopted to calculate photometric loss. Here the generated optical flow from the depth estimation network is called rigid flow like previous methods \citep{Zhou2019, Zou2018}. The finetuned optical flow estimation network provides optical flow to supervise estimated rigid flow. Furthermore, RGB indoor images contain only three channels and many \textcolor{red}{low-textured} regions. Yet their feature maps usually contain far more feature dimensions and fewer regions with low gradient variance. Thus, feature maps can provide more reliable supervision for depth learning. A multi-scale feature map synthesis loss with similar principle to view synthesis is proposed. Target image’s finetuned feature maps are warped by rigid flow to compute the \textcolor{red}{feature map synthesis} loss. 

Different from Moving Indoor \citep{Zhou2019}, only some patches of pixels with salient features are used to calculate photometric loss for optical flow estimation network in $\mathrm{F^2Depth}$ while Moving Indoor \citep{Zhou2019} adopts all pixels. In addition to optical flow consistency loss, we also design multi-scale feature map synthesis loss for depth learning. ResNet18s \citep{He2016} taking images as input are used for predicting relative pose between images in our $\mathrm{F^2Depth}$, while Moving Indoor \citep{Zhou2019} uses optical flow to estimate pose.

To evaluate the generalization ability of the proposed method, a Campus Indoor depth dataset is collected. The dataset contains 99 monocular images captured in 18 indoor scenes. 13-16 points with ground truth are selected in each image. Unlike NYU Depth V2 dataset captured with an RGB-D camera, the ground truth of every point’s depth is obtained with a laser rangefinder in the Campus Indoor dataset. The extensive experiments on the NYU Depth V2, \textcolor{red}{7-Scenes} and the Campus Indoor dataset prove the effectiveness and generalization ability of $\mathrm{F^2Depth}$.

The contributions of the paper are as follows: 

1) We propose a self-supervised indoor monocular depth estimation framework via optical flow learning with well-designed constraints to better train a depth estimation network. 

2) We optimize the photometric loss of optical flow estimation network, which utilizes some patches of pixels \textcolor{red}{centred} at extracted key points instead of all pixels. Thus, the influence of massive similar pixels in low-textured areas can be avoided. 

3) We use optical flow generated from finetuned flow estimation network as a supervisory signal, to make the rigid flow as consistent as possible with that. The multi-scale feature maps of images calculated by finetuned flow estimation network are further used as another supervisory signal.

4) The extensive depth estimation experiments on the NYU Depth V2 dataset validate the effectiveness of introducing optical flow learning. We also collect a Campus Indoor dataset on which we perform zero-shot generalization experiments with the model trained on NYU Depth V2. Our $\mathrm{F^2Depth}$ shows good generalization abilities \textcolor{red}{on the public 7-Scenes dataset and Campus Indoor.}

\section{Related work}
\subsection{Supervised indoor monocular depth estimation}
Ground truths of depth are required in supervised indoor monocular depth estimation methods. Most early methods \citep{Eigen2015, Eigen2014, Fu2018, Laina2016, Li2017, Liu2015} treated depth estimation as a pixel-wise regression problem. Some methods \citep{Eigen2015, Eigen2014} combined multi-scale features to estimate depths in a coarse-to-fine way. Due to the continuity of depth, a deep convolutional neural fields framework \citep{Liu2015} based on continuous conditional random field was proposed. In \citep{Li2017}, a two-streamed framework was proposed. A depth gradient stream branch was fused with a depth stream branch to estimate depth. A recent method \citep{Zhang2023} increased entropy for regression with higher-entropy feature spaces and achieved excellent indoor depth estimation performance. \textcolor{red}{IEBins \citep{shao2024iebins} reframed depth estimation as a classification-regression problem and proposed iterative elastic bins to search for high-accuracy depth.}

As the attention mechanism has been widely used in computer vision tasks, many works \citep{Agarwal2022, Agarwal2023, Bhat2021, Bhat2022, Jun2022, Yuan2022} tended to replace commonly used \textcolor{red}{Convolutional Neural Network (CNN)} modules such as ResNet \citep{He2016} with Transformer \citep{Vaswani2017}. \cite{Piccinelli2023} introduced an attention-based internal discretization bottleneck module to learn implicitly without any explicit constraints. \textcolor{red}{AiT \citep{ning2023all} leveraged a general tokenizer to model the outputs of depth estimation as discrete tokens. URCDC-Depth \citep{shao2023urcdc} introduced uncertainty rectified cross-distillation to make Transformer branch and the CNN branch learn from each other.}

Benefiting from high regularity of real indoor scenes, many of recent methods adopted geometric constraints to improve their performance. Some works \citep{Hu2019, Yin2019} used normal loss as one of loss terms. More methods introduced plane related priors to improve accuracy since planes are common in indoor scenes. P3Depth \citep{Patil2022} took advantage of piecewise planarity prior and predicted better at occlusion boundaries. PlaneNet \citep{Liu2018} reconstructed piece-wise planes through an end-to-end \textcolor{red}{Deep Neural Network (DNN)}. PlaneReg \citep{Yu2019} also performed piece-wise planar 3D models reconstruction via proposal-free instance segmentation and associative embedding model.

\subsection{Self-supervised indoor monocular depth estimation}
Self-supervised depth estimation methods do not require \textcolor{red}{labelled} training data. The design of loss functions is mainly based on multi-view geometry \citep{Wang2018_RN51}. Photometric consistency loss between multi-view images is the most commonly used loss term. However, the performance of self-supervised methods is generally worse than that of supervised methods. 

SfMLearner \citep{Zhou2017} is the first self-supervised monocular video depth estimation method using warping-based view synthesis as supervisory signal. Many methods proposed later are based on SfMLearner. Similar to view synthesis, \cite{Zhan2018} performed a warp for features to compute deep feature-based reconstruction loss but without experiments on indoor datasets. Monodepth2 \citep{Godard2019} minimized per-pixel reprojection loss to robustly address the occlusion issue and introduced auto-masking loss to avoid the influence of moving objects. Monodepth2 also achieved great performance on outdoor datasets such as \textcolor{red}{KITTI but} performed not satisfactorily on indoor datasets. To better predict depth for low-textured areas in indoor scenes, $\mathrm{P^2Net}$ \citep{Yu2020} adopted patch-match loss and plane-regularization prior based on Monodepth2 architecture. StructDepth \citep{Li2021} also took advantage of structural regularities, introduced the Manhattan normal constraint and the co-planar constraint. However, whether the constraints are valid will be affected by depth prediction results during training. Following their previous work SC-Depth V1 \citep{Bian2021_RN52} designed for outdoor scenes, SC-Depth V2 \citep{Bian2021_RN54} proposed an Auto-Rectify Network to overcome the problem of complex camera motion in indoor scenes. In order to minimize the impact of camera rotation, MonoIndoor \citep{Ji2021} added a residual pose estimation module apart from the initial pose estimation. \textcolor{red}{Based on MonoIndoor, MonoIndoor++ \citep{Li2022} performed coordinate convolutional encoding to provide coordinates guidance for the residual pose estimation module.} DistDepth \citep{Wu2022} performed structure distillation from the expert DPT \citep{Ranftl2021, Ranftl2020} and achieved state-of-the-art performance. Since DPT is a supervised method, DistDepth cannot completely get rid of the dependence on \textcolor{red}{labelled} data. \textcolor{red}{GasMono\citep{Zhao2023} estimates pose coarsely with the structure-from-motion software COLMAP\citep{schonberger2016colmap} and leverages Transformer to enhance feature learning of low-texture areas. ADPDepth \citep{Song2023} designed a PCAtt module to extract multiscale spatial information instead of local features.}

Accurate optical flow estimation can provide accurate pixel matching pairs between adjacent frames for depth estimation. Some methods improve their performance by utilizing optical flow. GeoNet and DF-Net \citep{Yin2018, Zou2018} learnt depth and optical flow simultaneously on outdoor datasets. TrianFlow \citep{Zhao2020} performed joint indoor depth and pose learning using optical flow correspondence. \textcolor{red}{\citep{Xu2023} unifies flow and depth estimation through directly comparing feature similarities.} Moving Indoor \citep{Zhou2019} utilized optical flow produced by flow estimation network SF-Net as a supervisory signal, to supervise rigid flow generated from DepthNet and PoseNet. Inspired by Moving Indoor \citep{Zhou2019}, we design our monocular depth estimation framework via optical flow learning. Different from Moving Indoor \citep{Zhou2019}, we propose a patch-based photometric loss for optical flow learning while SF-Net calculates photometric loss using all pixels. Besides optical flow consistency loss, multi-scale feature map synthesis loss is also designed to supervise depth learning. $\mathrm{F^2Depth}$ adopts ResNet18s \citep{He2016} to compute relative pose. The ResNet18s take images as inputs while Moving Indoor \citep{Zhou2019} uses optical flow to predict pose.

\subsection{Unsupervised optical flow estimation}
FlowNet \citep{Dosovitskiy2015, Ilg2017} is a classic supervised optical flow estimation network, and many subsequent unsupervised methods are based on it. Similar to self-supervised depth estimation methods, photometric loss is the mainly used loss term in unsupervised optical flow estimation. Apart from photometric loss, UnFlow \citep{Meister2018} introduced a bidirectional census loss to learn bidirectional optical flow. To handle the occlusion issue, \cite{Wang2018_RN94} modeled occlusion explicitly and incorporated it into the loss.

PWC-Net \citep{Sun2018} proposed a learnable feature pyramid and achieved smaller network size and higher accuracy than FlowNet 2.0 \citep{Ilg2017}. Based on PWC-Net architecture, \cite{Janai2018} proposed to learn optical flow and occlusion over multiple frames to strengthen photometric loss. \cite{Zhong2019} introduced epipolar geometry constraint to address repetitive textures and occlusion issues. \textcolor{red}{MDFlow \citep{kong2022mdflow} introduced mutual knowledge distillation of teacher and student network to avoid mismatch. BrightFlow \citep{marsal2023brightflow} performed optical flow and brightness changes estimation jointly to improve robustness to brightness changes.} ARFlow \citep{Liu2020} added a forward branch of data augmentation to be more robust and used a \textcolor{red}{highly shared} flow decoder to be lightweight. However, ARFlow cannot achieve satisfactory performance on indoor datasets since there exist many low-textured areas in indoor scenes.

\section{Method}
\begin{figure}[!htb]
    \centering
    \includegraphics[scale=0.4]{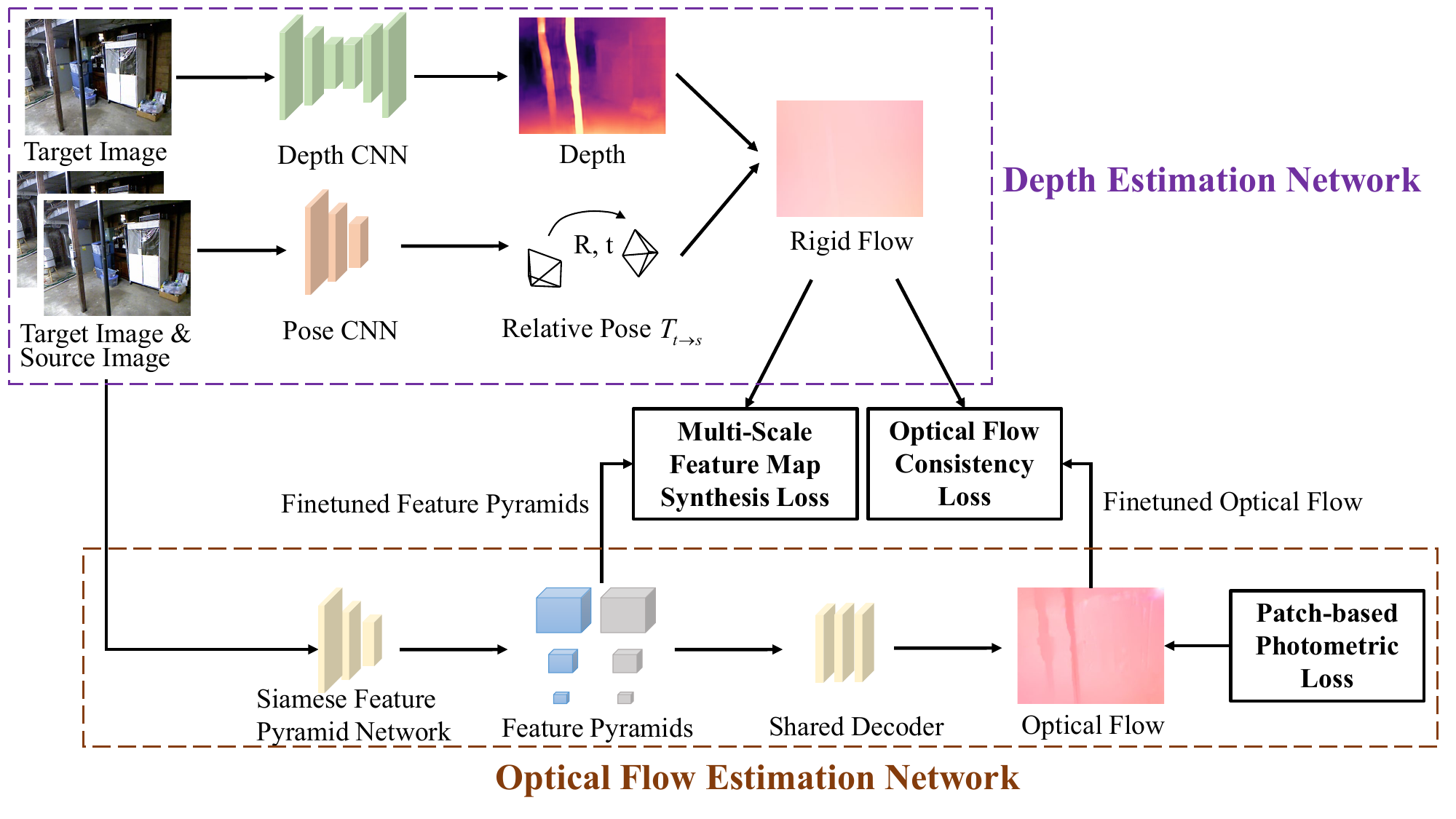}
    \caption{\textcolor{red}{The pipeline of $\mathrm{F^2Depth}$. The optical flow estimation network is first finetuned with the patch-based photometric loss. The finetuned feature pyramids are warped with rigid flow to compute feature map synthesis loss for depth estimation network. The optical flow consistency loss aims to make rigid flow consistent with finetuned optical flow.}}
    \label{fig:fig_1}
    
\end{figure}

In this section, we will introduce how our self-supervised indoor monocular depth estimation framework $\mathrm{F^2Depth}$ works. \textcolor{red}{The pipeline of our $\mathrm{F^2Depth}$ is shown in} \autoref{fig:fig_1}. An optical flow estimation network is first finetuned to supervise monocular depth learning. The photometric loss of optical flow estimation network is optimized using patches of pixels with discriminative features for finetuning. The weights of finetuned optical flow estimation network are frozen. The finetuned optical flow estimation network provides optical flow and multi-scale feature maps as two supervisory signals respectively. The optical flow consistency loss aims to make finetuned optical flow and rigid flow as consistent as possible. The multi-scale feature map synthesis loss aims to minimize the differences between warped feature maps and real feature maps. All the improvements we make aim to obtain more accurate pixel motions in low-textured areas so that the overall depth estimation performance can be improved.

In Section~\ref{sec3.1}, the improvement we make to the photometric loss of optical flow estimation network is explained. Section~\ref{sec3.2}~and~\ref{sec3.3} introduce the proposed optical flow consistency loss and multi-scale feature map synthesis loss respectively. The overall loss functions are illustrated in Section~\ref{sec3.4}.

\subsection{Patch-based photometric loss of optical flow estimation network}\label{sec3.1}

\begin{figure}[!htb]
    \centering
    \includegraphics[scale=0.4]{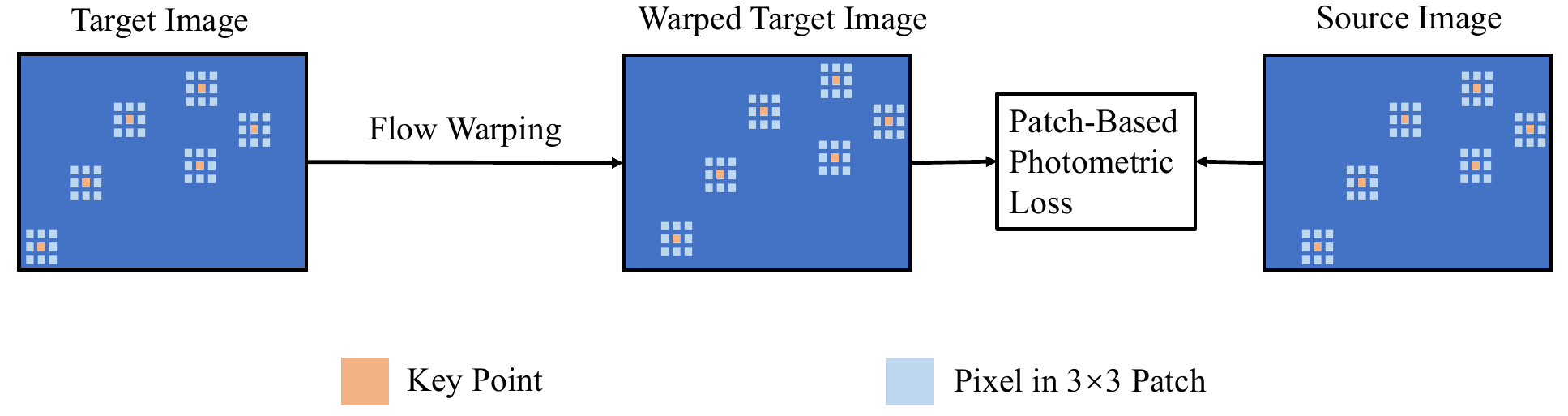}
    \caption{Patch-based photometric loss of optical flow estimation network}
    \label{fig:fig_2}
\end{figure}

The original optical flow estimation network adopts pixel-wise photometric loss, which can achieve good performance on outdoor datasets such as KITTI and the animated film dataset MPI Sintel \citep{Butler2012, Wulff2012}. However, it is difficult for the optical flow estimation network to be directly applied to indoor scenes due to the presence of many low-textured areas. To solve the low-texture issue, we propose the patch-based photometric loss. We only use some non-occluded pixels distributed in patches for photometric loss calculation, instead of using all non-occluded pixels, as shown in \autoref{fig:fig_2}. 

We first extract key points with Direct Sparse Odometry (DSO) \citep{Engel2017}, a classical feature extraction method commonly used in Simultaneous Localization and Mapping (SLAM). Like key points extracted, pixels around key points also have significant gradient variance. Therefore, pixels in a 3×3 patch \textcolor{red}{centred} at every key point are used to calculate photometric loss. This can avoid pixels in low-textured regions influencing the training process. The pixels in a patch can be written as
\begin{equation}\label{eq_P}
P=\{(x+x_k,y+y_k),x_k\in \{-N,0,N\},y_k\in \{-N,0,N\}\}
\end{equation}
where $(x,y)$ is the coordinate of the key point, and $N$ is set to 2. For the predicted optical flow ${f_{t \to s}}$ from the target image to the source image, the patch-based photometric loss is formulated as
\begin{equation}\label{eq_L_patch}
   {L_{patch}}({f_{t \to s}}) = \sum\limits_{{P_i}} {F({I_s}({P_i}),\widehat {{I_t}}({P_i}))}
\end{equation}
where $F(\cdot)$ is a function for computing differences between pixels. To reduce the impact of illumination changes on photometric consistency, we use ternary census transform loss to describe similarity between images like previous works \citep{Meister2018, Zou2018}. ${I_s}({P_i})$ denotes the intensity of a pixel in a patch of the source image, ${\widehat I_t}({P_i})$ denotes the warped pixel intensity that can be calculated by
\begin{equation}
    \widehat {{I_t}}({P_i}) = {I_t}({P_i} - {f_{t \to s}})
\end{equation}
where ${I_t}({P_i})$ denotes the intensity of a pixel in a patch of the target image.

\subsection{Optical flow consistency loss}\label{sec3.2}

Like most monocular depth estimation methods, our depth estimation network has two main modules: DepthCNN that outputs target image’s depth and PoseCNN that predicts pose between adjacent images. The rigid flow from target image to source image is defined as
\begin{equation}\label{eq_f_rigid}
{f_{rigid}}({p_t}) = {p_s} - {p_t}
\end{equation}
where $p_t$ and $p_s$ denote the corresponding pixel coordinates in target image and source image respectively. The projection relationship between target image’s predicted depth and 2D pixel homogeneous coordinates in images can be written as
\begin{equation}\label{eq_ps}
{p_s} = K{T_{t \to s}}{D_t}({p_t}){K^{ - 1}}{p_t}
\end{equation}
where $K$ denotes camera intrinsic parameters matrix, $T_{t\to s}$ denotes the predicted relative pose from target image to source image, $D_t(p_t)$ denotes the predicted depth of target image. From Eq.~\ref{eq_f_rigid} and \ref{eq_ps}, we can see that accurate rigid flow is critical for calculating accurate corresponding point pairs to guarantee accurate depth estimation to a certain extent. To obtain more precise rigid flow, we set a constraint that rigid flow should be in consistency with optical flow generated from finetuned flow estimation network. The corresponding loss term of this constraint is as follows:
\begin{equation}
{L_{rigid}} = \left| {{f_{rigid}}({p_t}) - {f_{flow}}({p_t})} \right|
\end{equation}
where ${f_{flow}}({p_t})$ stands for the optical flow produced by the flow estimation network. In training, we use frame 0 and frame +1 as the target image and source image. 

\subsection{Multi-scale feature map synthesis loss}\label{sec3.3}

\begin{figure}[!htb]
    \centering
    \includegraphics[scale=0.4]{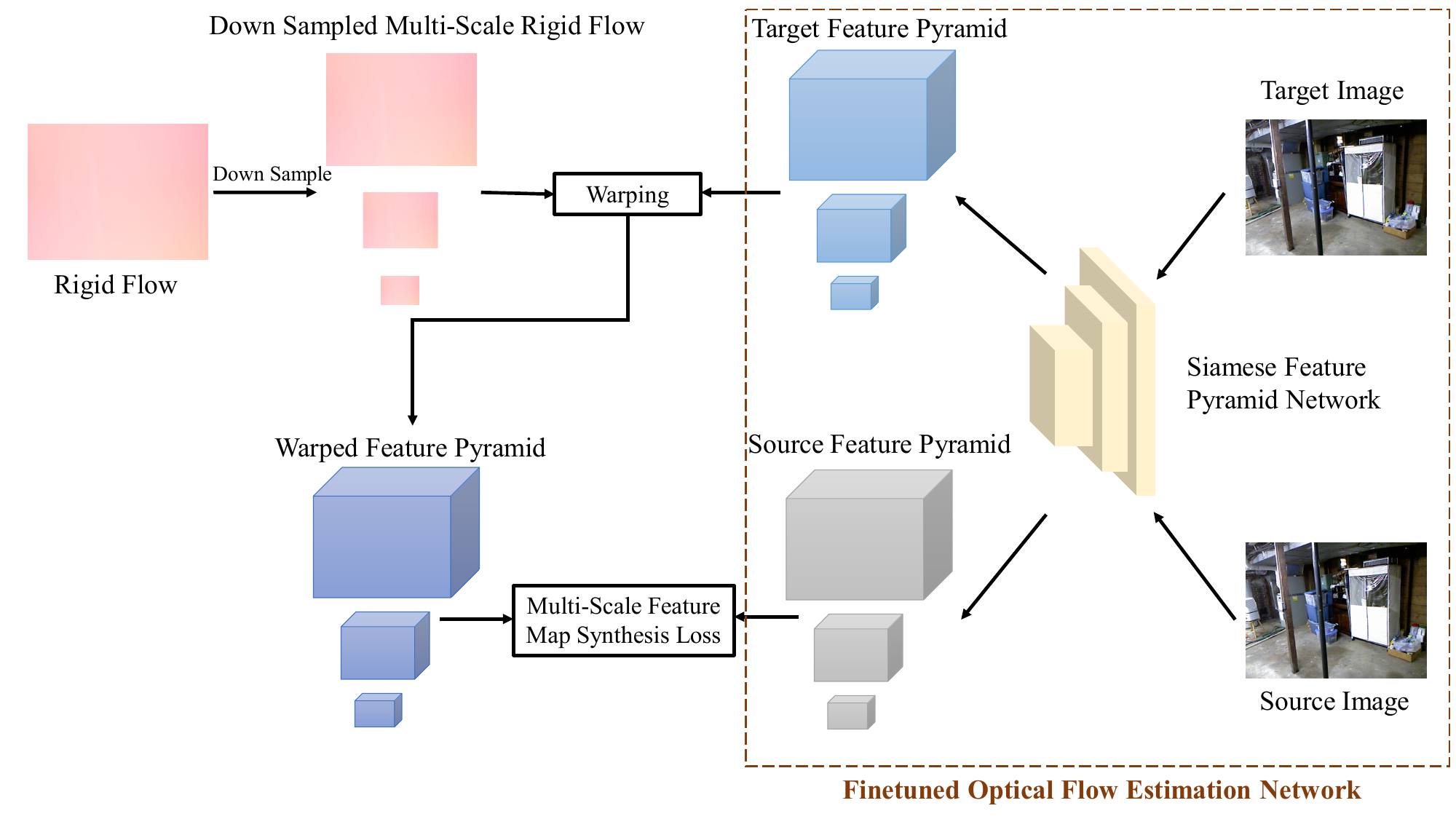}
    \caption{Multi-scale feature map synthesis loss}
    \label{fig:fig_3}
\end{figure}

We introduce a multi-scale feature map synthesis loss together with the optical flow consistency loss to supervise depth learning. The multi-scale feature maps of images are generated by the flow estimation network through a \textcolor{red}{Siamese} feature pyramid network \citep{Chopra2005}. Compared with RGB images with only three channels, their feature maps usually contain far more feature dimensions and fewer regions with low gradient variance. The multi-scale feature map synthesis loss is similar to photometric loss based on view synthesis. The target image’s multi-scale feature maps are warped using down sampled rigid flow of corresponding scales. The warped feature maps should be as consistent as possible with real source image’s feature map. In this way, rigid flow can be estimated more accurately so that performance of depth estimation can be improved. The feature channels of different scales are 16, 32, 64, 96, 128, 192. The detailed principle of the loss is shown in \autoref{fig:fig_3}
The feature map synthesis loss is calculated using L1 loss:
\begin{equation}
L_{feature} = \left| {{F_{src}}({p_t}) - \hat F({p_t})} \right|
\end{equation}
where ${F_{src}}({p_t})$ denotes the feature map of the source image, ${\hat F({p_t})}$ denotes the warped feature map of target image that can be calculated as
\begin{equation}
    \widehat F({p_t}) = {F_{tgt}}({p_t} - {f_{t \to s}})
\end{equation}
where ${F_{tgt}}({p_t})$ denotes the feature map of the target image.

\subsection{Overall loss functions}\label{sec3.4}

For the optical flow estimation network, the overall loss function is composed of patch-based photometric loss ${L_{patch}}$  and smoothness loss ${L_{sm}}$ : 
\begin{equation}
L_{flow} = {L_{patch}}({f_{t \to s}}) + \lambda {L_{sm}}({f_{t \to s}})
\end{equation}
where ${f_{t \to s}}$ denotes the predicted optical flow from target image to source image.   $\lambda $ is set to 50 in our experiments. The smoothness loss term ${L_{sm}}$  is defined as
\begin{equation}
{L_{sm}}({f_{t \to s}}) = |{\partial _x}{f_{t \to s}}|\exp ( - \left| {{\partial _x}{I_t}} \right|) + |{\partial _y}{f_{t \to s}}|\exp ( - \left| {{\partial _y}{I_t}} \right|)
\end{equation}
where ${I_t}$  denotes the target image. The smoothness loss term guarantees the prediction results of neighboring pixels are similar without significant gradient change. 

For the depth estimation network, the overall loss function is formulated as
\begin{equation}
L_{depth} = {L_{ph}} + {\lambda _1}{L_{sm}} + {\lambda _2}{L_{spp}} + {L_{rigid}} + {\lambda _3}{L_{feature}}
\end{equation}
where ${L_{ph}}$  denotes the patch-based multi-view photometric consistency loss,  ${L_{sm}}$ denotes the smoothness loss and  ${L_{spp}}$ denotes the planar consistency loss. These three loss terms are defined exactly the same as $\mathrm{P^2Net}$ \citep{Yu2020}. ${L_{rigid}}$  and ${L_{feature}}$  are the proposed optical flow consistency loss and multi-scale feature map synthesis loss.  ${\lambda _1}$ is set to 0.001, ${\lambda _2}$  is set to 0.05,  ${\lambda _3}$ is set to 3 in our experiments.

\section{Experiments}

\subsection{Implementation details}

\textbf{Datasets:} The optical flow estimation network is finetuned and evaluated on the public NYU Depth V2 dataset \citep{Silberman2012}. We train and evaluate the depth estimation network on NYU Depth V2 dataset. \textcolor{red}{The zero-shot generalization experiments are conducted on the 7-Scenes dataset} \citep{Shotton2013} \textcolor{red}{and the Campus Indoor dataset collected by us.}

NYU Depth V2 is a dataset of 582 indoor scenes recorded by the Microsoft Kinect. We train the networks using the same train split consisting of 283 scenes as previous works \citep{Yu2020, Zhou2019}, and evaluate the depth estimation network on the official test set composed of 654 densely \textcolor{red}{labelled} images. The training set contains 21483 images that have been undistorted and sampled at intervals of 10 frames.

\textcolor{red}{7-Scenes is a public dataset containing 7 indoor scenes. There are several video sequences in each scene. We perform zero-shot generalization experiments on the official test split of 17000 images.}

\begin{figure}[!htb]
    \centering
    \includegraphics[width=0.45\textwidth]{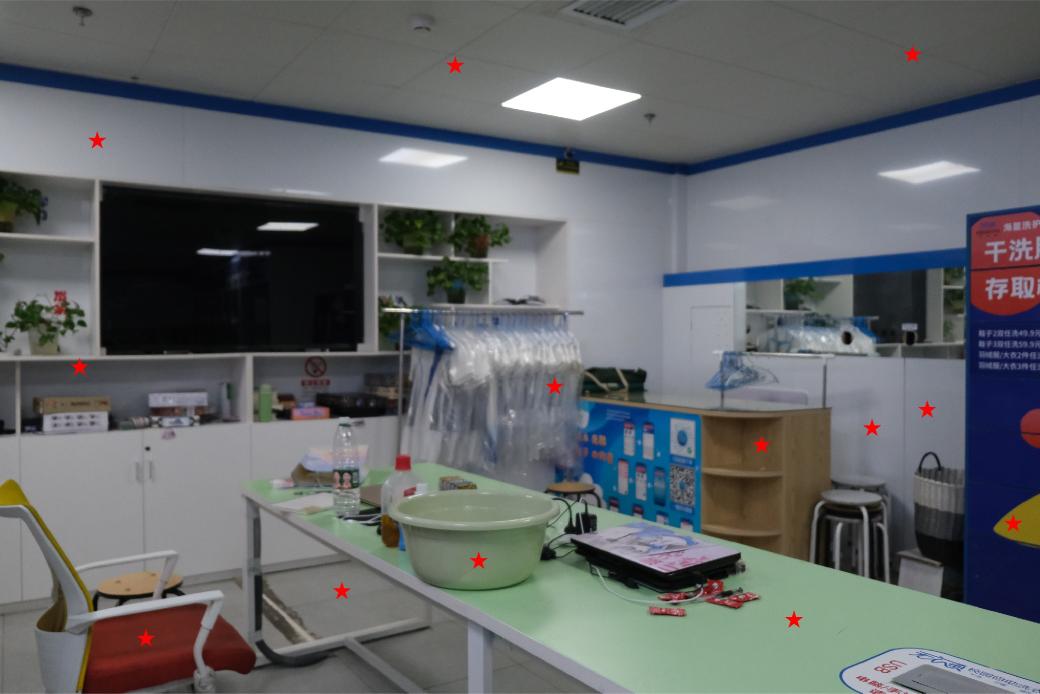}
    \includegraphics[width=0.45\textwidth]{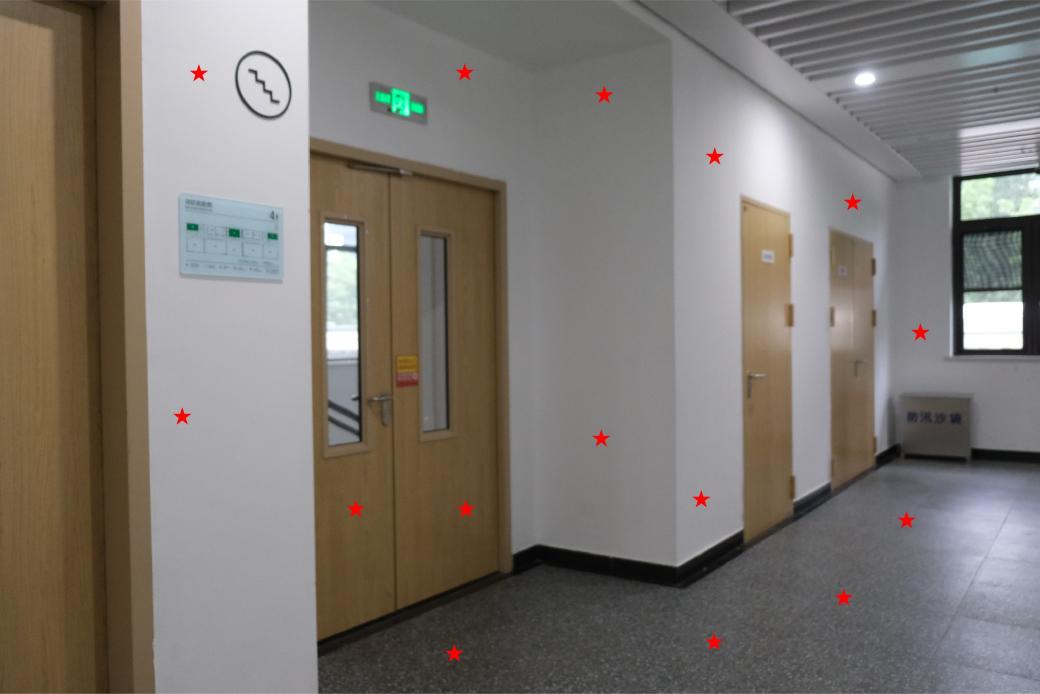}\\ 
    \includegraphics[width=0.45\textwidth]{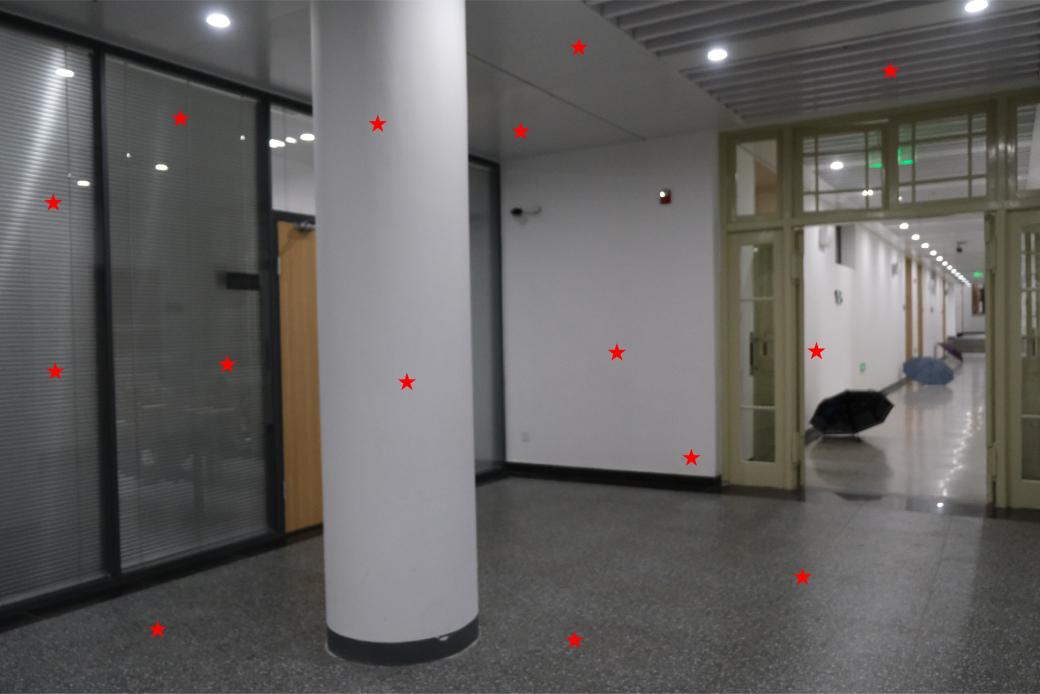}
    \includegraphics[width=0.45\textwidth]{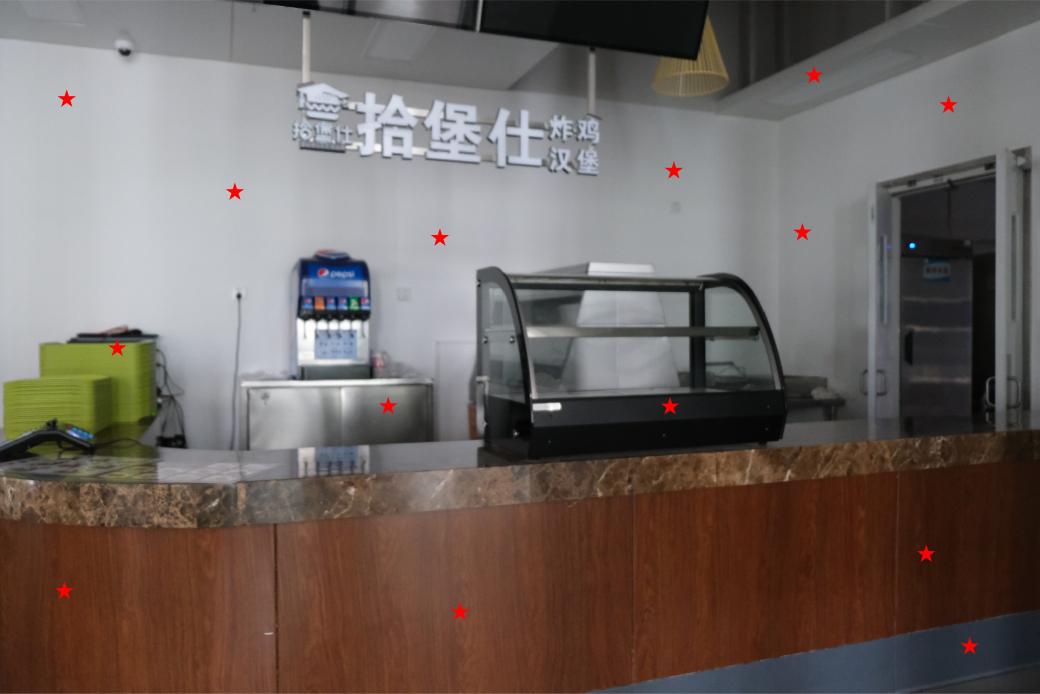}\\ 
    \includegraphics[width=0.45\textwidth]{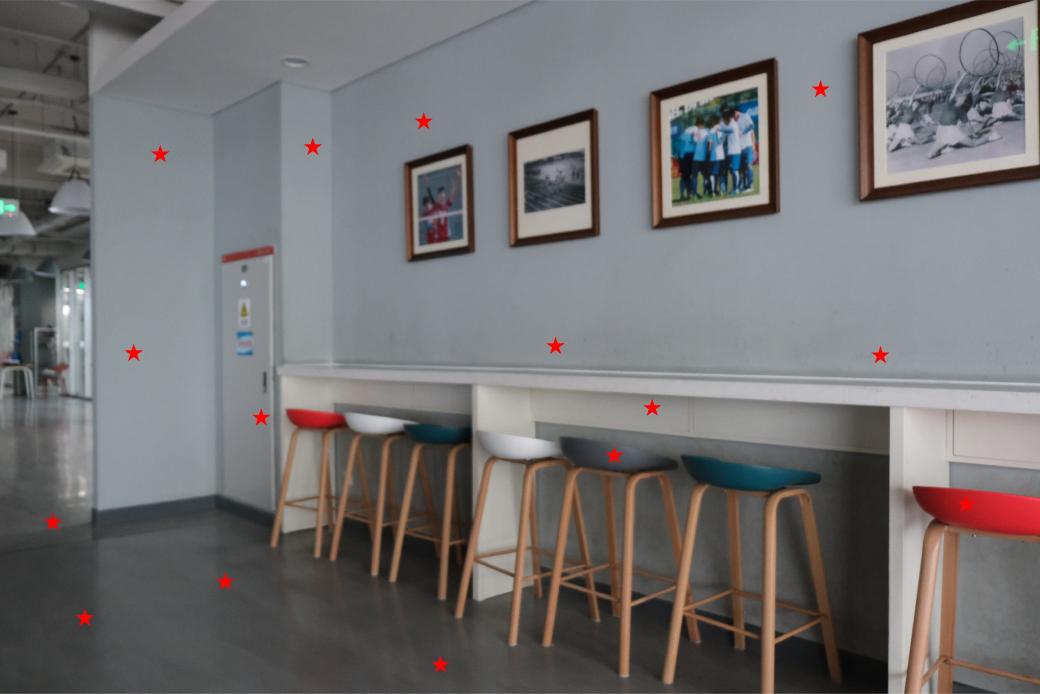}
    \includegraphics[width=0.45\textwidth]{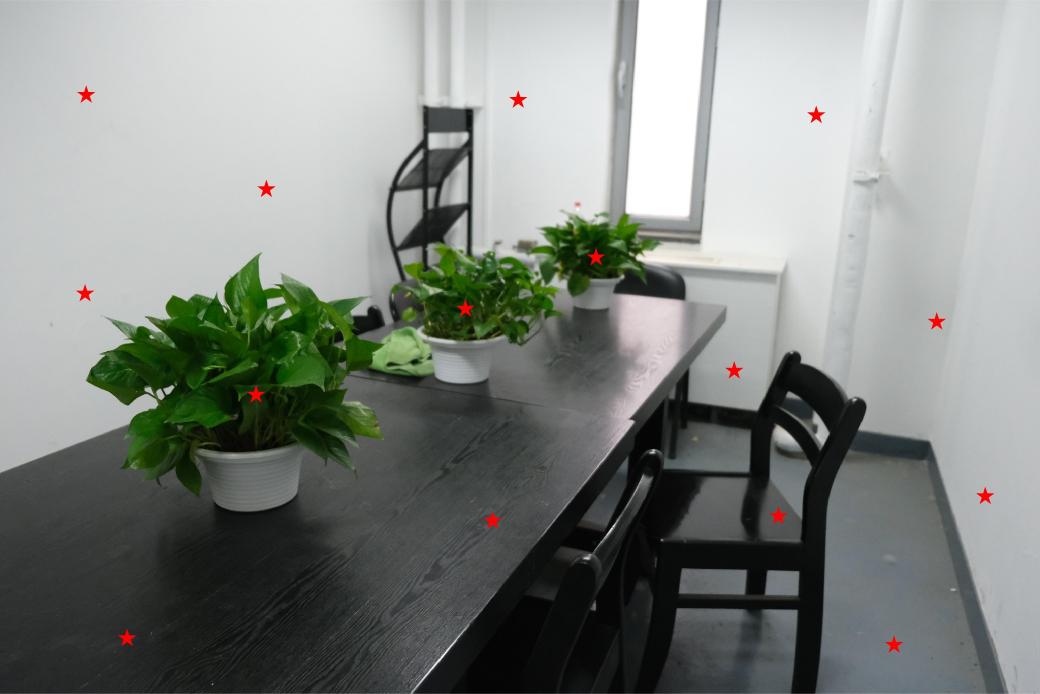}\\
    \caption{Some scenes and selected points in the Campus Indoor dataset}
    \label{fig:fig_4}
\end{figure}

The Campus Indoor dataset is made up of 99 monocular images captured with a camera FUJIFILM X-T30 in 18 indoor scenes in a campus. We select 13-16 points in large depth ranges and uniform distribution in each image as shown in \autoref{fig:fig_4}, approximately 1500 points in total. The red stars show the positions of selected points. \textcolor{red}{The points are evenly distributed in the image and cover small to large depths. We select points on as many objects as possible.} We measure depths of selected points from the camera with a laser rangefinder as ground truth.

\textbf{Experimental setup:} The depth estimation network architecture is based on $\mathrm{P^2Net}$ \citep{Yu2020}. The optical flow estimation network architecture is based on ARFlow \citep{Liu2020} because ARFlow is lightweight and has been pretrained on the animated film MPI Sintel dataset. We train both the optical flow estimation network and the depth estimation network with the same training set. The training dataset is randomly color augmented and flipped. Images are resized to 288 × 384 pixels while training. 

We train $\mathrm{F^{2}Depth}$ on an NVIDIA GeForce RTX 3090 GPU. The flow estimation network has been pretrained with original loss function on the MPI Sintel dataset made up of animated film frames. We finetune the optical flow estimation network on NYU Depth V2 dataset for 37 epochs with the batch size of 20. The target frame and its next source frame are used for finetuning. The learning rate is set to 5e-5 without adjustment. We train the depth estimation network for 44 epochs with the batch size of 12 using one target frame and 2 source frames. For the depth estimation network, the learning rate is set to 1e-4 for the first 27 epochs, decayed by 0.1 for the next 15 epochs and reduced to 1e-6 in the last 2 epochs.

\textbf{Evaluation metrics:} We adopt the same median scaling strategy and metrics as previous methods for depth evaluation \citep{Yu2020, Zhou2019}. The evaluation metrics include the mean absolute relative error ($Abs Rel$), the root mean squared error ($RMS$), the mean log10 error and accuracies under three thresholds ($\delta<1.25$, $\delta<{1.25^2}$, $\delta <{1.25^3}$). The metrics can be formulated as 
\begin{equation}
Abs\;Rel = \frac{1}{N}\sum\limits_{i \in N} {\frac{{|{D_i} - {{\hat D}_i}|}}{{{{\hat D}_i}}}}
\end{equation}
\begin{equation}
RMS = \sqrt {\frac{1}{N}\sum\limits_{i \in N} {||{D_i} - {{\hat D}_i}|{|^2}} }
\end{equation}
\begin{equation}
    Mean\;\log 10 = \frac{1}{N}\sum\limits_{i \in N} {|{{\log }_{10}}{D_i} - {{\log }_{10}}{{\hat D}_i}|}
\end{equation}
\begin{equation}
Accuracies\;\%\;of\;{D_i}\;s.t.\;\max \left( {\frac{{{D_i}}}{{{{\hat D}_i}}},\frac{{{{\hat D}_i}}}{{{D_i}}}} \right) = \delta \; < \;thr
\end{equation}
where $\hat{D}$ denotes the ground truths of depth, $D$ denotes the prediction results of depth, $N$ denotes the number of pixels, $thr$  is the threshold.

\subsection{Results}
We introduce results of depth estimation on NYU Depth V2, \textcolor{red}{zero-shot generalization results on the 7-Scenes and the Campus Indoor dataset respectively.}

\subsubsection{Evaluation of depth estimation on NYU Depth V2}

We evaluate on the official test split provided by NYU Depth V2 dataset for comparison with previous methods including some supervised ones. The results are shown in \autoref{tab:tab1}. \textcolor{red}{Post-processing is proposed in \citep{Godard2017}, referring to left-right flipping augmentation at test time. Inference with post-processing is efficient, taking only 13.5 ms per image. Inference without post-processing takes 9.4 ms.} Compared with $\mathrm{P^2Net}$ \citep{Yu2020}, our method $\mathrm{F^2Depth}$ improves on all metrics. \textcolor{red}{The RMS error decreases by 2.671\% for the results without post-processing, and 3.885\% with post-processing.} The improvement indicates the effectiveness of adding optical flow learning.

\begin{footnotesize}

\setlength\tabcolsep{3pt}
\begin{longtable}[c]{>{\raggedright\arraybackslash}m{4.9cm} m{0.2cm} m{0.2cm} c c c c c c}

\hline
Method & S & \textcolor{red}{T} & \makecell[c]{REL\\ $\downarrow$} & \makecell[c]{RMS\\ $\downarrow$} & \makecell[c]{log10\\ $\downarrow$} & \makecell[c]{$\delta\!<\!1.25$\\$\uparrow$} & \makecell[c]{$\delta\!<\!1.25^2$\\$\uparrow$} &\makecell[c]{$\delta\!<\!1.25^3$\\$\uparrow$}  \\
\hline
Make3D~\citep{Saxena2005} & \checkmark & \textcolor{red}{$\times$} & 0.349 & 1.214 & - & 0.447 & 0.745 & 0.897 \\
Liu~\citep{Liu2014}& \checkmark & \textcolor{red}{$\times$} & 0.335 & 1.06 & 0.127 & - & - & - \\
Li~\citep{Li2015}& \checkmark & \textcolor{red}{$\times$} & 0.232 & 0.821 & 0.094 & 0.621 & 0.886 & 0.968 \\
Liu~\citep{Liu2015}& \checkmark & \textcolor{red}{$\times$} & 0.213 & 0.759 & 0.087 & 0.650 & 0.906 & 0.976 \\
Eigen~\citep{Eigen2015}& \checkmark & \textcolor{red}{$\times$} & 0.158 & 0.641 & - & 0.769 & 0.950 & 0.988 \\
Li~\citep{Li2017}& \checkmark & \textcolor{red}{$\times$} & 0.143 & 0.635 & 0.063 & 0.788 & 0.958 & 0.991 \\
PlaneNet~\citep{Liu2018} & \checkmark & \textcolor{red}{$\times$} & 0.142 & 0.514 & 0.060 & 0.827 & 0.963 & 0.990 \\
PlaneReg~\citep{Yu2019} & \checkmark & \textcolor{red}{$\times$} & 0.134 & 0.503 & 0.057 & 0.827 & 0.963 & 0.990 \\
Laina~\citep{Laina2016}& \checkmark & \textcolor{red}{$\times$} & 0.127 & 0.573 & 0.055 & 0.811 & 0.953 & 0.988 \\
DORN~\citep{Fu2018}& \checkmark & \textcolor{red}{$\times$} & 0.115 & 0.509 & 0.051 & 0.828 & 0.965 & 0.992 \\
VNL~\citep{Yin2019}& \checkmark & \textcolor{red}{$\times$} & 0.108 & 0.416 & 0.048 & 0.875 & 0.976 & 0.994 \\
P3Depth~\citep{Patil2022}& \checkmark & \textcolor{red}{$\times$} & 0.104 & \textbf{0.356} & \textbf{0.043} & 0.898 & 0.981 & 0.996 \\
Jun~\citep{Jun2022} & \checkmark & \textcolor{red}{\checkmark} & \textbf{0.100} & 0.362 & \textbf{0.043} & \textbf{0.907} & \textbf{0.986} & \textbf{0.997} \\
\hline
\textcolor{red}{MonoIndoor++}~\citep{Li2022}& \textcolor{red}{$\times$} & \textcolor{red}{\checkmark} & \textcolor{red}{0.132} & \textcolor{red}{0.517} & \textcolor{red}{-} & \textcolor{red}{0.834} & \textcolor{red}{0.961} & \textcolor{red}{0.990} \\
\textcolor{red}{GasMono}~\citep{Zhao2023}& \textcolor{red}{$\times$} & \textcolor{red}{\checkmark} & \textcolor{red}{\textbf{0.113}} & \textcolor{red}{\textbf{0.459}} & \textcolor{red}{-} & \textcolor{red}{\textbf{0.871}} & \textcolor{red}{\textbf{0.973}} & \textcolor{red}{\textbf{0.992}} \\
\hline
Moving Indoor~\citep{Zhou2019}& × & \textcolor{red}{$\times$} & 0.208 & 0.712 & 0.086 & 0.674 & 0.900 & 0.968 \\
TrianFlow~\citep{Zhao2020} & × & \textcolor{red}{$\times$} & 0.189 & 0.686 & 0.079 & 0.701 & 0.912 & 0.978 \\
Zhang~\citep{Zhang2022}& × &\textcolor{red}{ $\times$} & 0.177 & 0.634 & - & 0.733 & 0.936 & - \\
Monodepth2~\citep{Godard2019}& × & \textcolor{red}{$\times$} & 0.170 & 0.617 & 0.072 & 0.748 & 0.942 & 0.986 \\
\textcolor{red}{ADPDepth}~\citep{Song2023}& \textcolor{red}{$\times$} &\textcolor{red}{ $\times$} & \textcolor{red}{0.165} & \textcolor{red}{0.592} & \textcolor{red}{0.071} & \textcolor{red}{0.753} & \textcolor{red}{0.934} & \textcolor{red}{0.981} \\
SC-Depth~\citep{Bian2021_RN52}& × & \textcolor{red}{$\times$} & 0.159 & 0.608 & 0.068 & 0.772 & 0.939 & 0.982 \\
$\mathrm{P^2Net}$ \citep{Yu2020}& × & \textcolor{red}{$\times$} & 0.159 & 0.599 & 0.068 & 0.772 & 0.942 & 0.984 \\
\textcolor{red}{$\mathrm{F^2Depth}$} & × & \textcolor{red}{$\times$} & {0.158} & {0.583} & {0.067} & {0.779} & {0.947} & \textbf{0.987} \\
\textcolor{red}{$\mathrm{P^2Net}$+PP} \citep{Yu2020}& \textcolor{red}{$\times$} & \textcolor{red}{$\times$} & \textcolor{red}{0.157} & \textcolor{red}{0.592} & \textcolor{red}{0.067} & \textcolor{red}{0.777} & \textcolor{red}{0.944} & \textcolor{red}{0.985} \\
\textcolor{red}{$\mathrm{F^2Depth}$+PP} & \textcolor{red}{$\times$} & \textcolor{red}{$\times$} & \textcolor{red}{\textbf{0.153}} & \textcolor{red}{\textbf{0.569}} & \textcolor{red}{\textbf{0.065}} & \textcolor{red}{\textbf{0.787}} & \textcolor{red}{\textbf{0.950}} & \textcolor{red}{\textbf{0.987}} \\
\hline
\caption{Comparison of evaluation results between our method and other methods on the NYU Depth V2 dataset. S represents supervision. \textcolor{red}{T indicates the usage of Transformer. PP denotes the results with post-processing.} $\downarrow$ means the lower the better, $\uparrow$  means the higher the better.}\\
\label{tab:tab1}\\

\end{longtable}

\end{footnotesize}

\begin{figure}[H]
    \centering
    \includegraphics[width=0.23\textwidth]{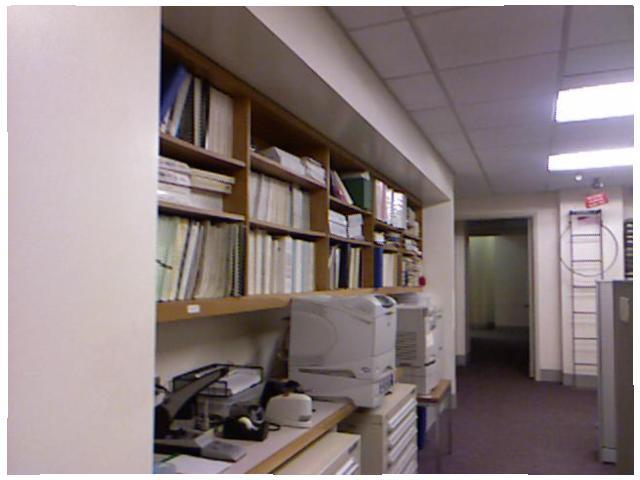}
    \includegraphics[width=0.23\textwidth]{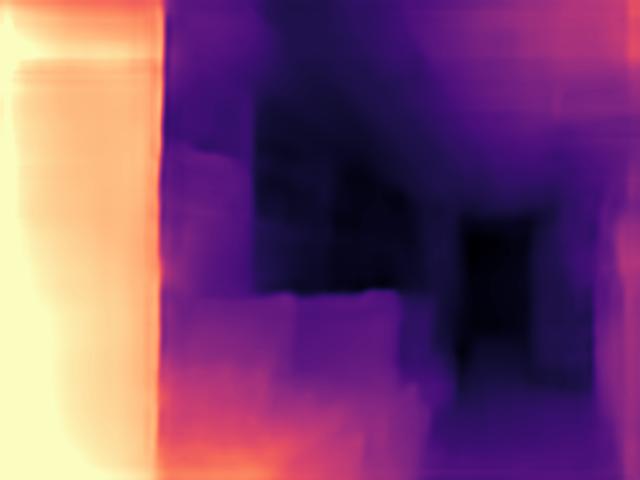}
    \includegraphics[width=0.23\textwidth]{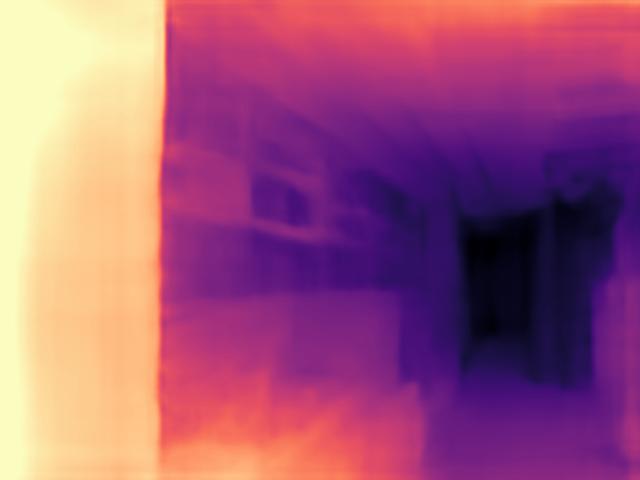}
    \includegraphics[width=0.23\textwidth]{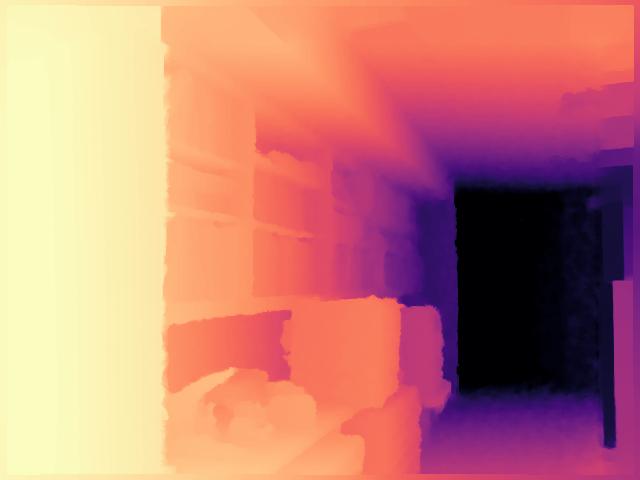}
    \includegraphics[width=0.23\textwidth]{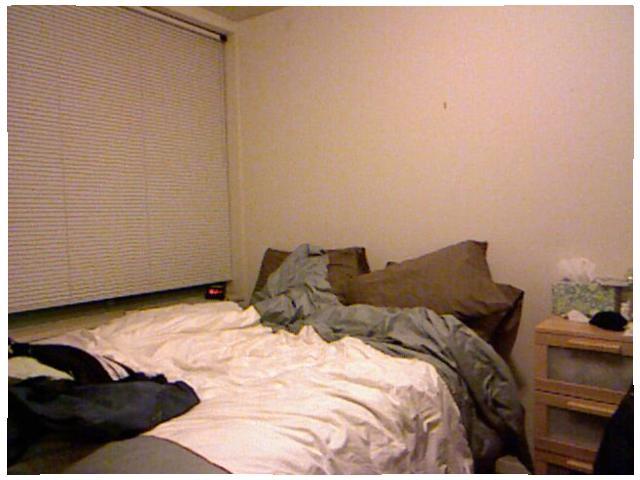}
    \includegraphics[width=0.23\textwidth]{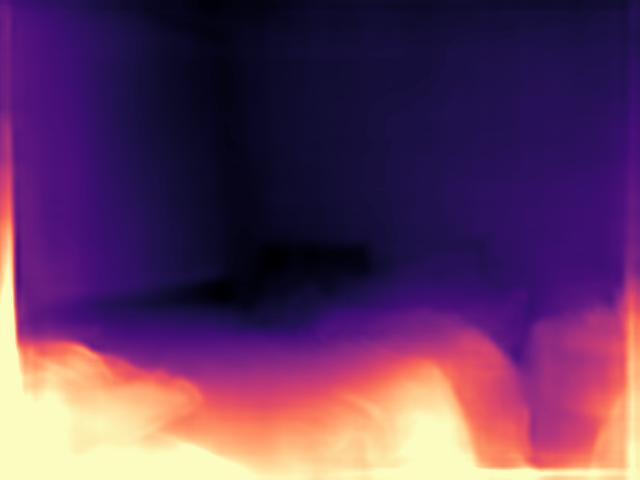}
    \includegraphics[width=0.23\textwidth]{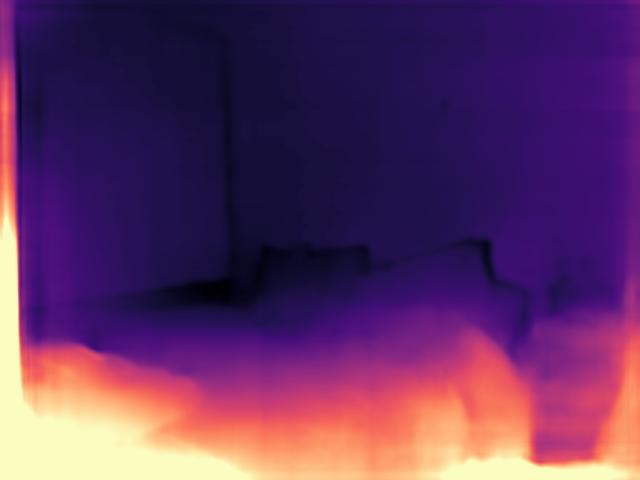}
    \includegraphics[width=0.23\textwidth]{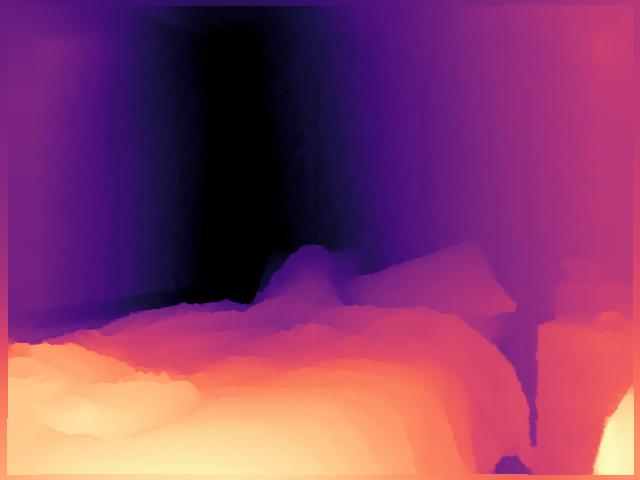}
    \includegraphics[width=0.23\textwidth]{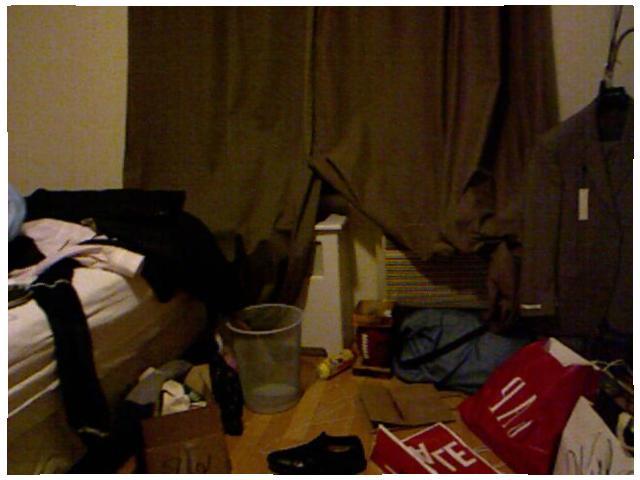}
    \includegraphics[width=0.23\textwidth]{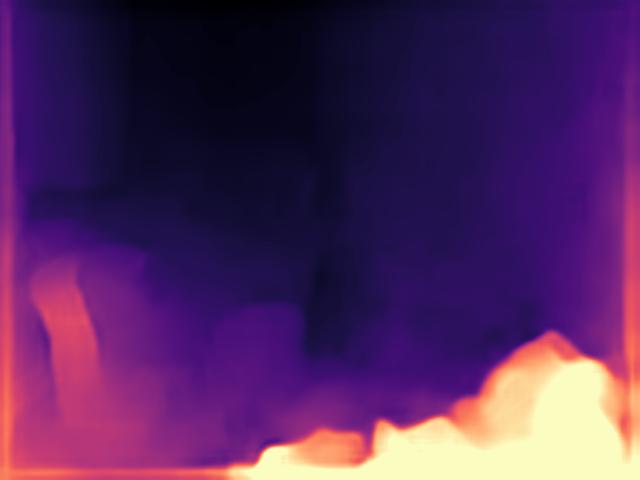}
    \includegraphics[width=0.23\textwidth]{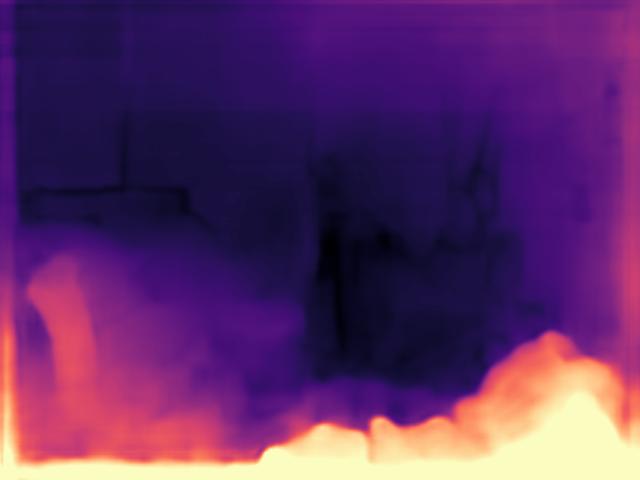}
    \includegraphics[width=0.23\textwidth]{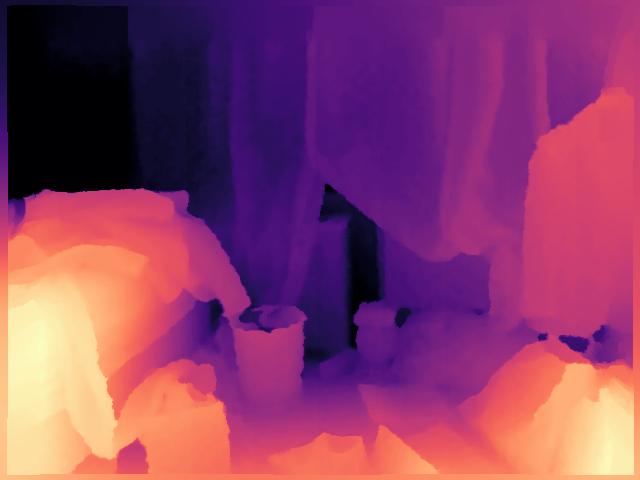}
    \includegraphics[width=0.23\textwidth]{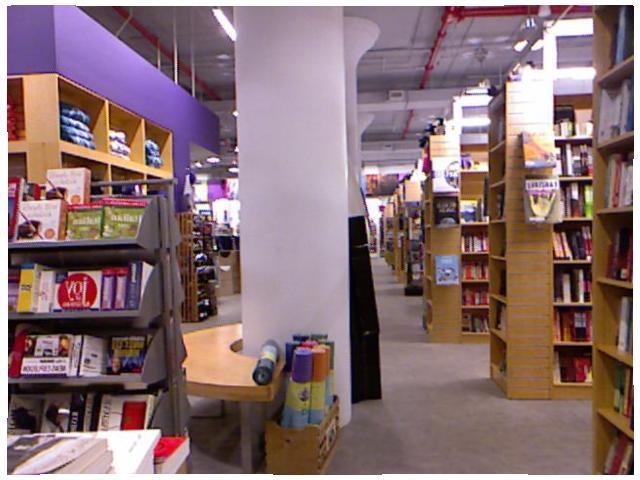}
    \includegraphics[width=0.23\textwidth]{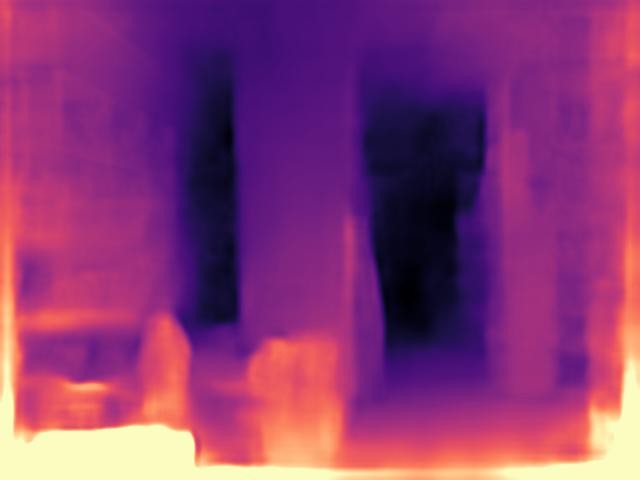}
    \includegraphics[width=0.23\textwidth]{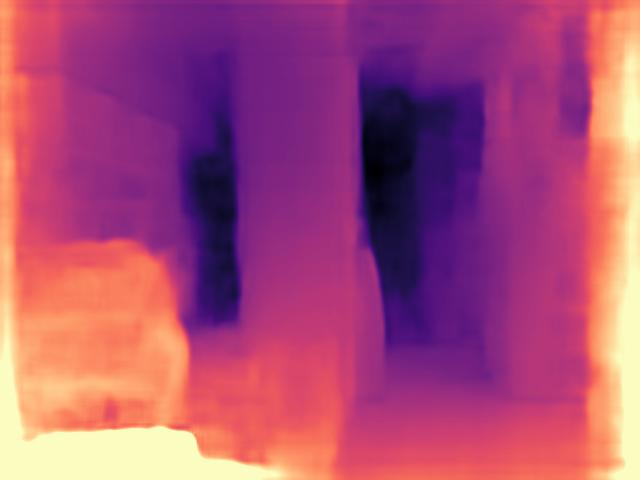}
    \includegraphics[width=0.23\textwidth]{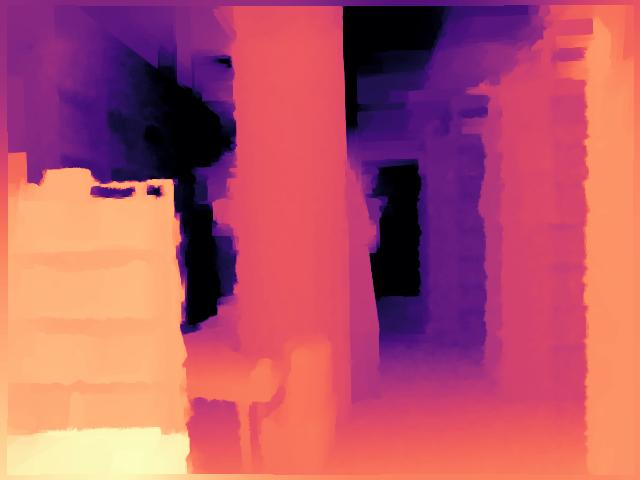}
    \includegraphics[width=0.23\textwidth]{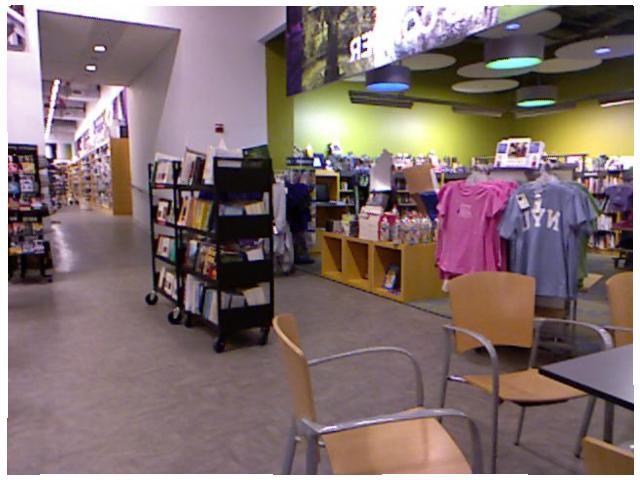}
    \includegraphics[width=0.23\textwidth]{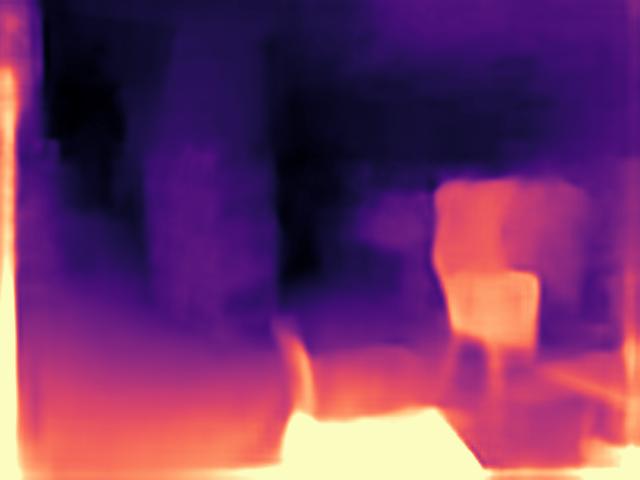}
    \includegraphics[width=0.23\textwidth]{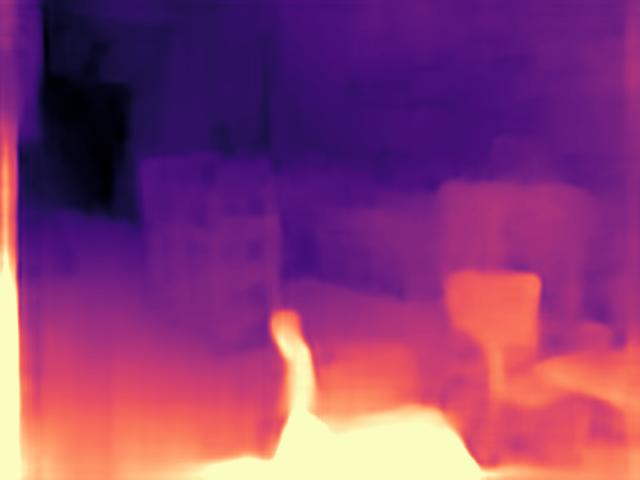}
    \includegraphics[width=0.23\textwidth]{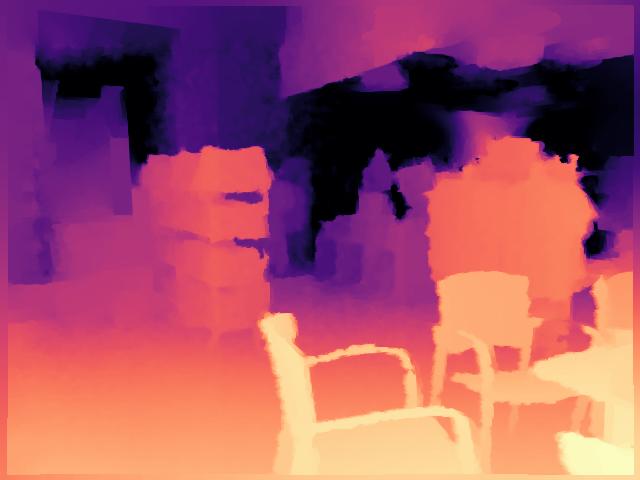}
    \includegraphics[width=0.23\textwidth]{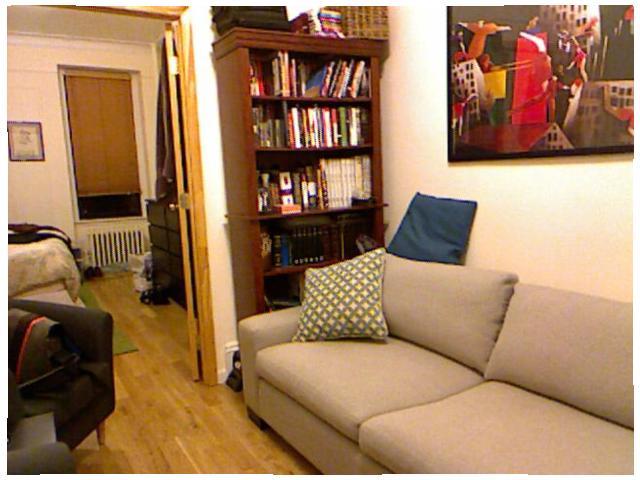}
    \includegraphics[width=0.23\textwidth]{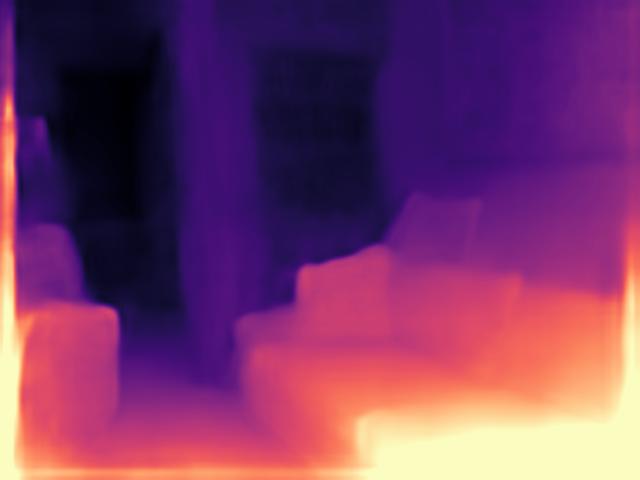}
    \includegraphics[width=0.23\textwidth]{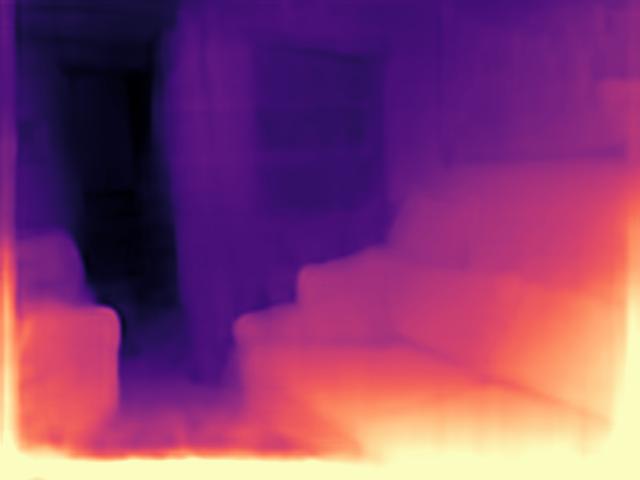}
    \includegraphics[width=0.23\textwidth]{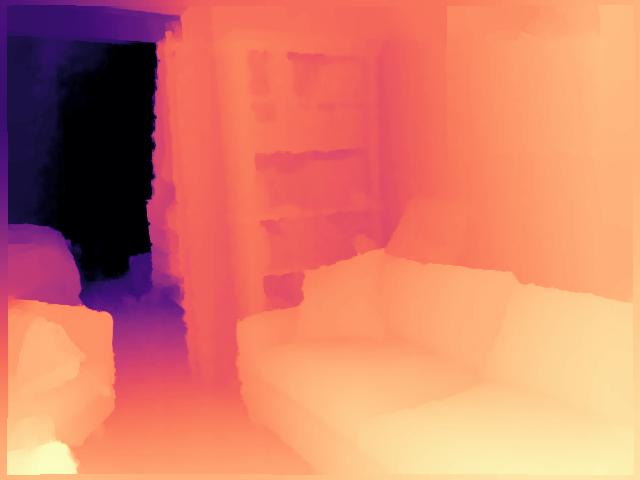}\\
    \vspace{-0.25cm}
    \begin{subfigure}[b]{0.23\textwidth}
        \caption{\textcolor{red}{RGB image}}
    \end{subfigure}
    \begin{subfigure}[b]{0.23\textwidth}
        \caption{\textcolor{red}{$\mathrm{P^2Net}$}}
    \end{subfigure}
    \begin{subfigure}[b]{0.23\textwidth}
        \caption{\textcolor{red}{Ours}}
    \end{subfigure}
    \begin{subfigure}[b]{0.23\textwidth}
        \caption{\textcolor{red}{Ground Truth}}
    \end{subfigure}\\
    \vspace{-0.3cm}
    \caption{Depth estimation results on NYU Depth V2. Left to right: RGB image, $\mathrm{P^2Net}$ \citep{Yu2020}, ours and ground truth.}
    \label{fig:fig_5}
\end{figure}

\autoref{fig:fig_5} shows depth estimation results of $\mathrm{P^2Net}$ \citep{Yu2020} and our method. Our method outperforms $\mathrm{P^2Net}$ for regions with protruding structures such as bookshelves and low-textured areas such as walls and curtains. 

\begin{figure}
    \centering
    \includegraphics[scale=0.84]{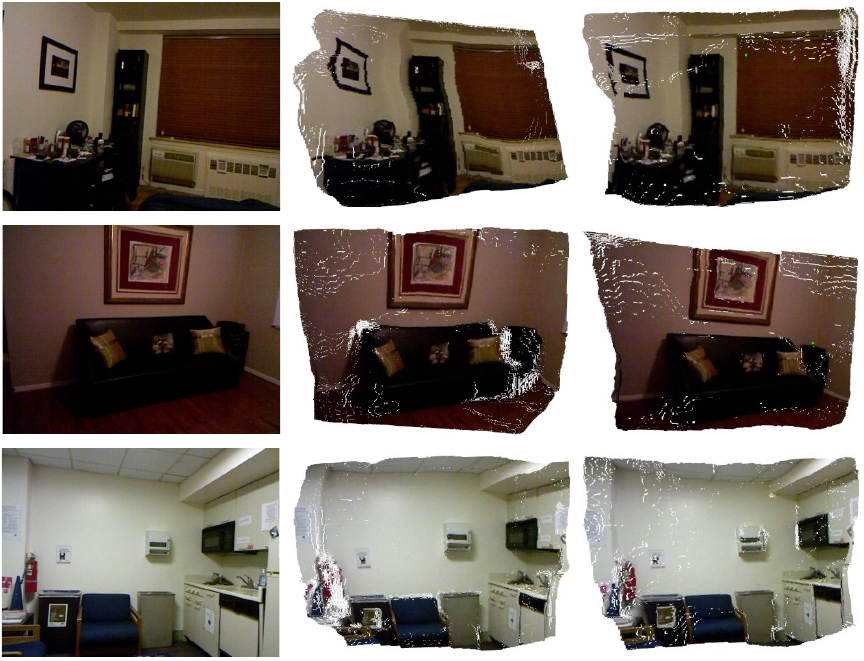}\\
    \vspace{-0.25cm}
    \begin{subfigure}[b]{0.3\textwidth}
        \caption{\textcolor{red}{RGB Image}}
    \end{subfigure}
    \begin{subfigure}[b]{0.3\textwidth}
        \caption{\textcolor{red}{$\mathrm{P^2Net}$}}
    \end{subfigure}
    \begin{subfigure}[b]{0.3\textwidth}
        \caption{\textcolor{red}{Ours}}
    \end{subfigure}\\
    \vspace{-0.25cm}
    \caption{3D reconstruction results on NYU Depth V2. Left to right: RGB image, $\mathrm{P^2Net}$ \citep{Yu2020} and ours.}
    \label{fig:fig_6}
\end{figure}

Visualization of 3D reconstruction is shown in \autoref{fig:fig_6}. Our reconstruction is better than $\mathrm{P^2Net}$ \citep{Yu2020} in terms of completeness and accuracy. \textcolor{red}{The 3D reconstruction is based on the TSDF (Truncated Signed Distance Field) method, which combines multiple depth maps of the same scene into a single volumetric reconstruction. The reconstruction is further converted to textured 3D meshes for visualization. Both the depth map and the volumetric reconstruction represent the 3D structure of the scene, but in different forms. Benefiting from better depth estimation performance, $\mathrm{F^2Depth}$ can obtain better 3D reconstruction.}

\subsubsection{\texorpdfstring{\textcolor{red}{Zero-shot generalization results on 7-Scenes}}{Zero-shot generalization results on 7-Scenes}}
\begin{figure}[H]
    \centering
    \includegraphics[width=0.242\textwidth]{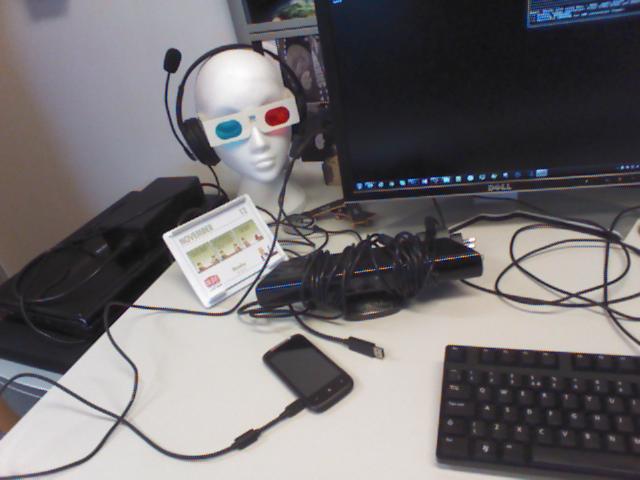}
    \includegraphics[width=0.242\textwidth]{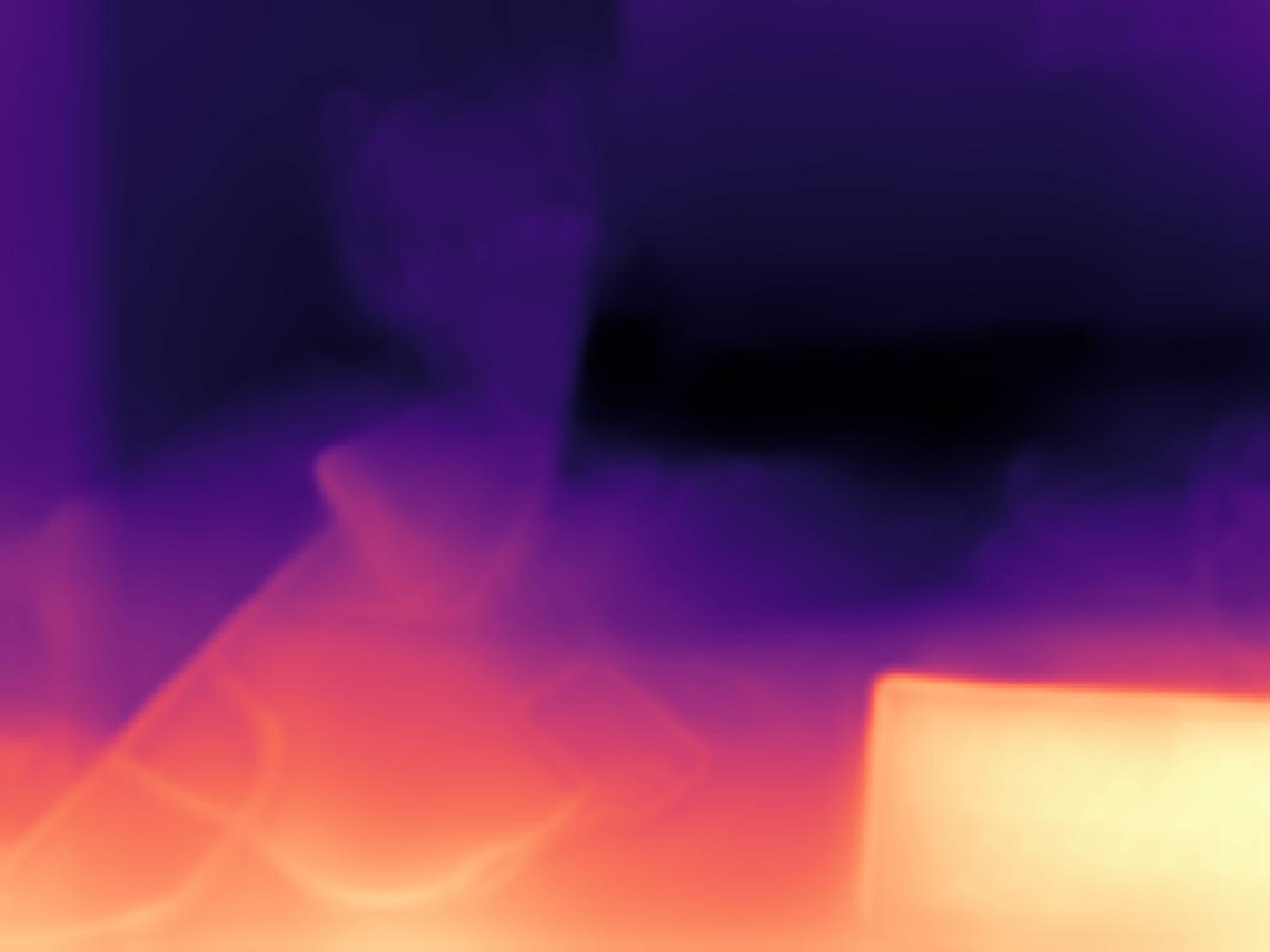}
    \includegraphics[width=0.242\textwidth]{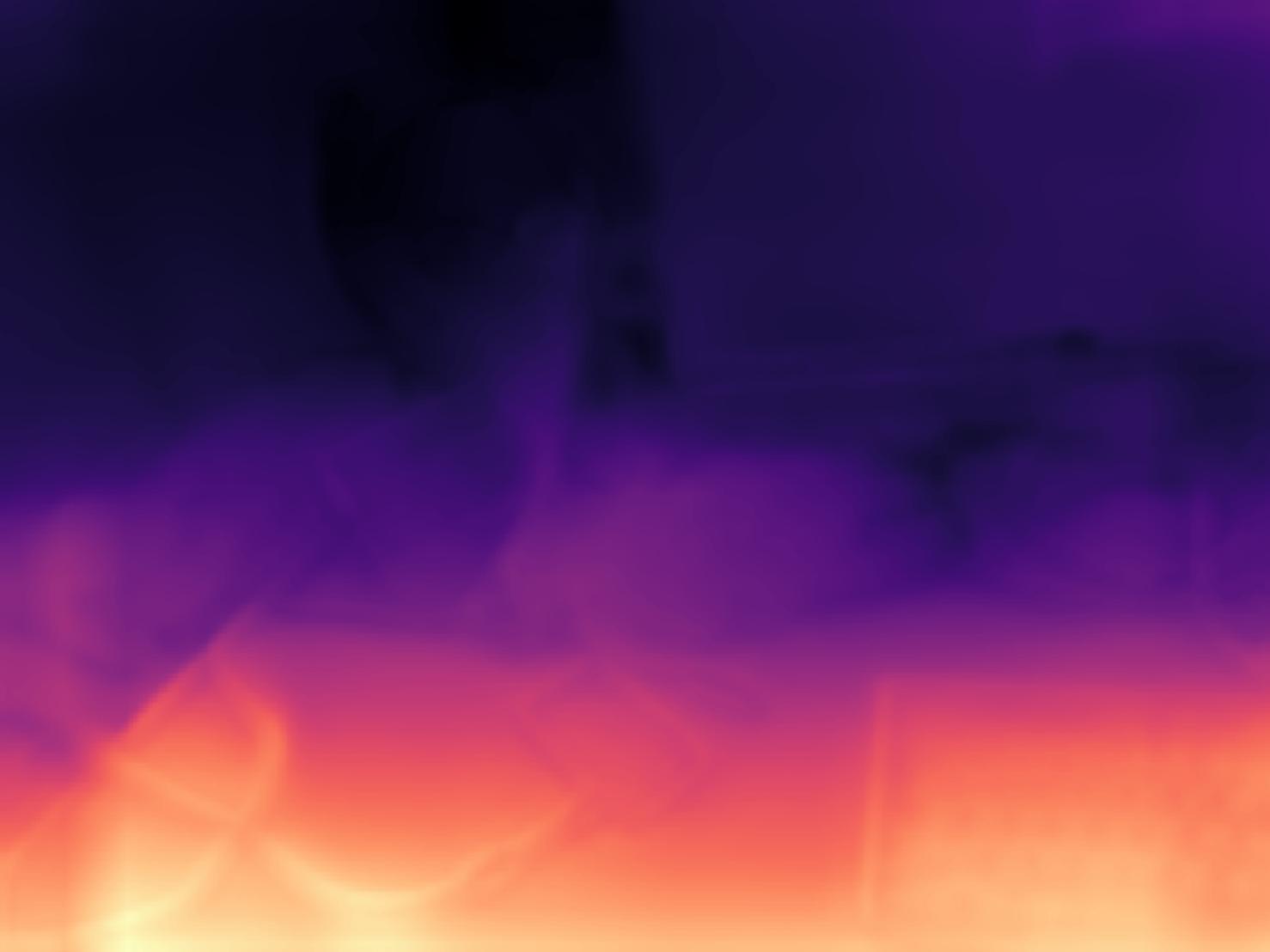}
    \includegraphics[width=0.242\textwidth]{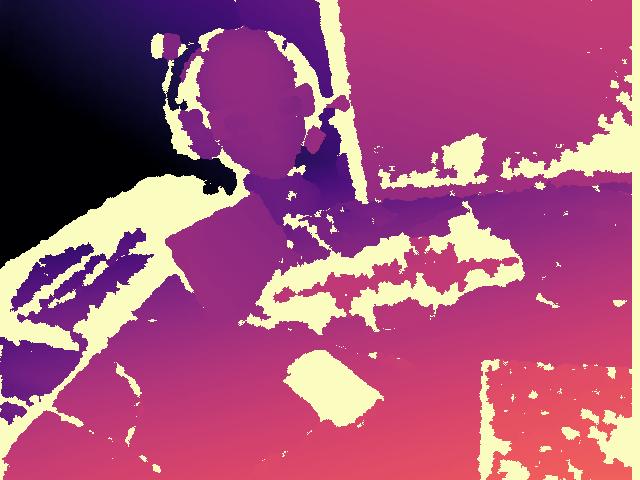}
    \includegraphics[width=0.242\textwidth]{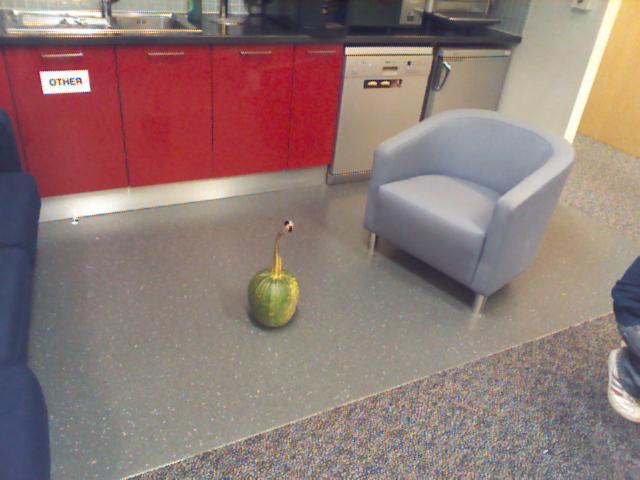}
    \includegraphics[width=0.242\textwidth]{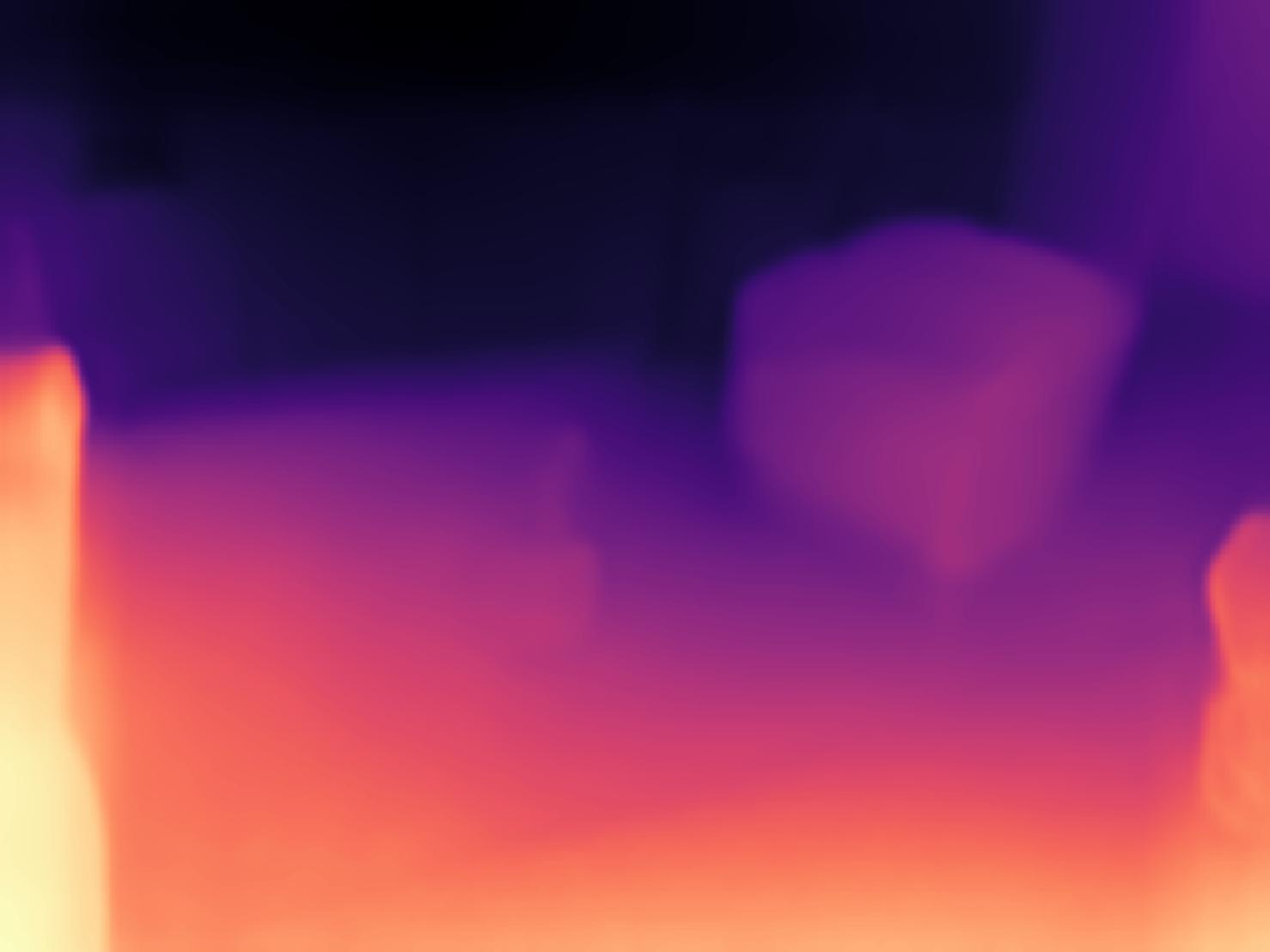}
    \includegraphics[width=0.242\textwidth]{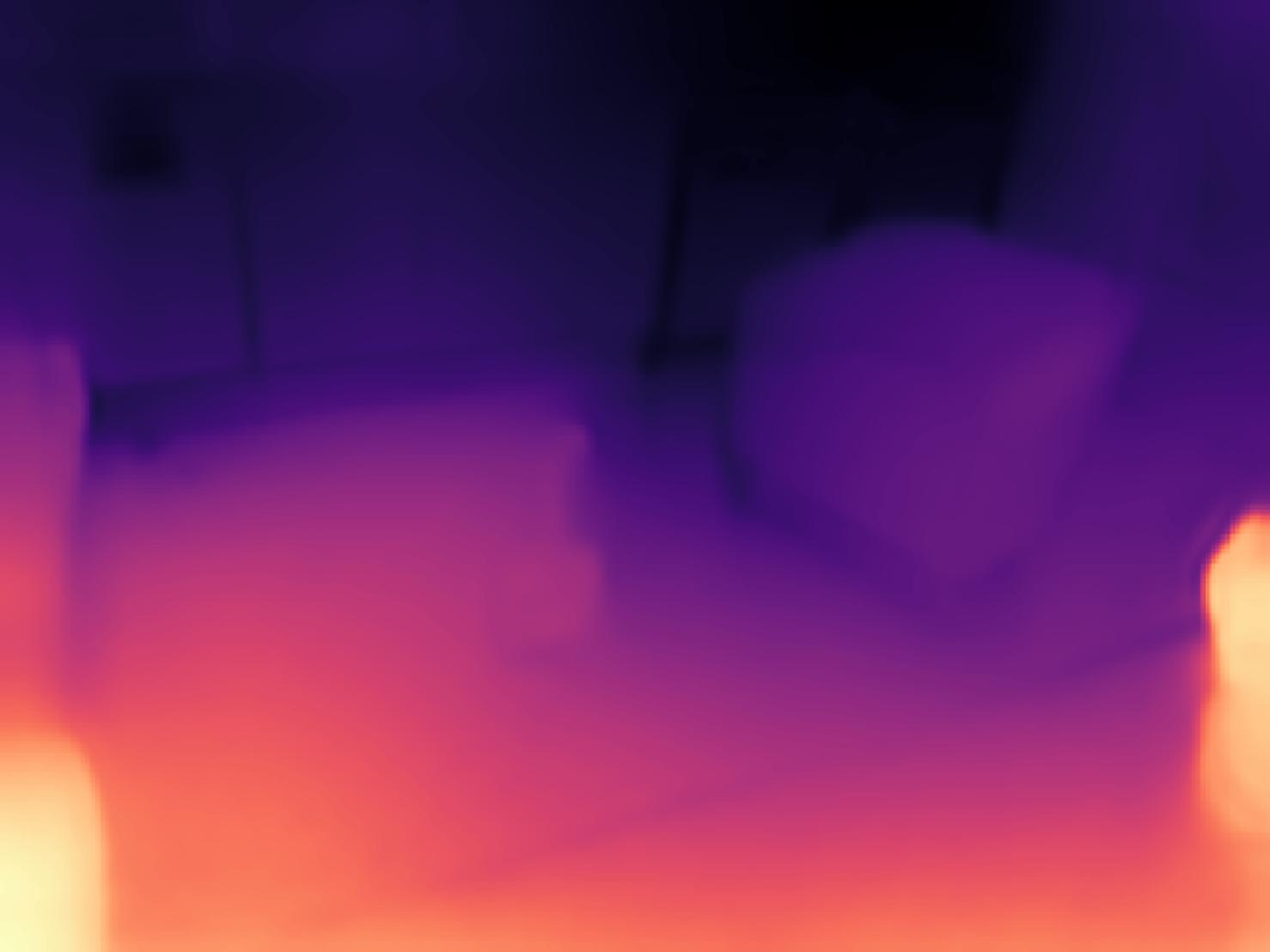}
    \includegraphics[width=0.242\textwidth]{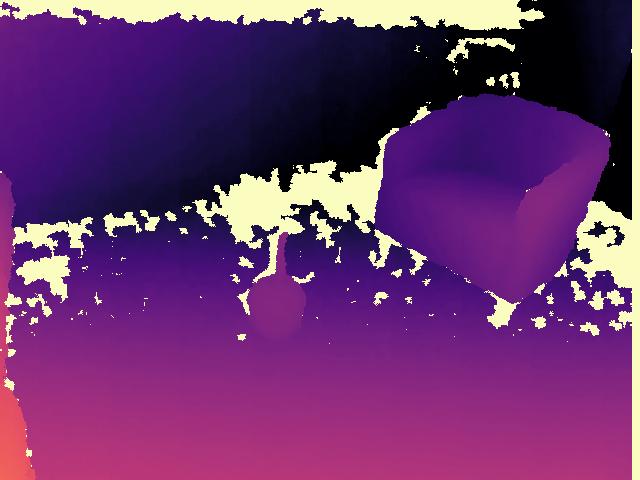}
    \includegraphics[width=0.242\textwidth]{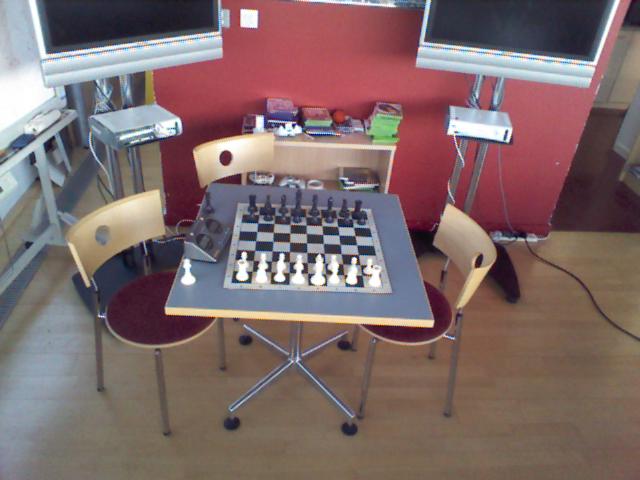}
    \includegraphics[width=0.242\textwidth]{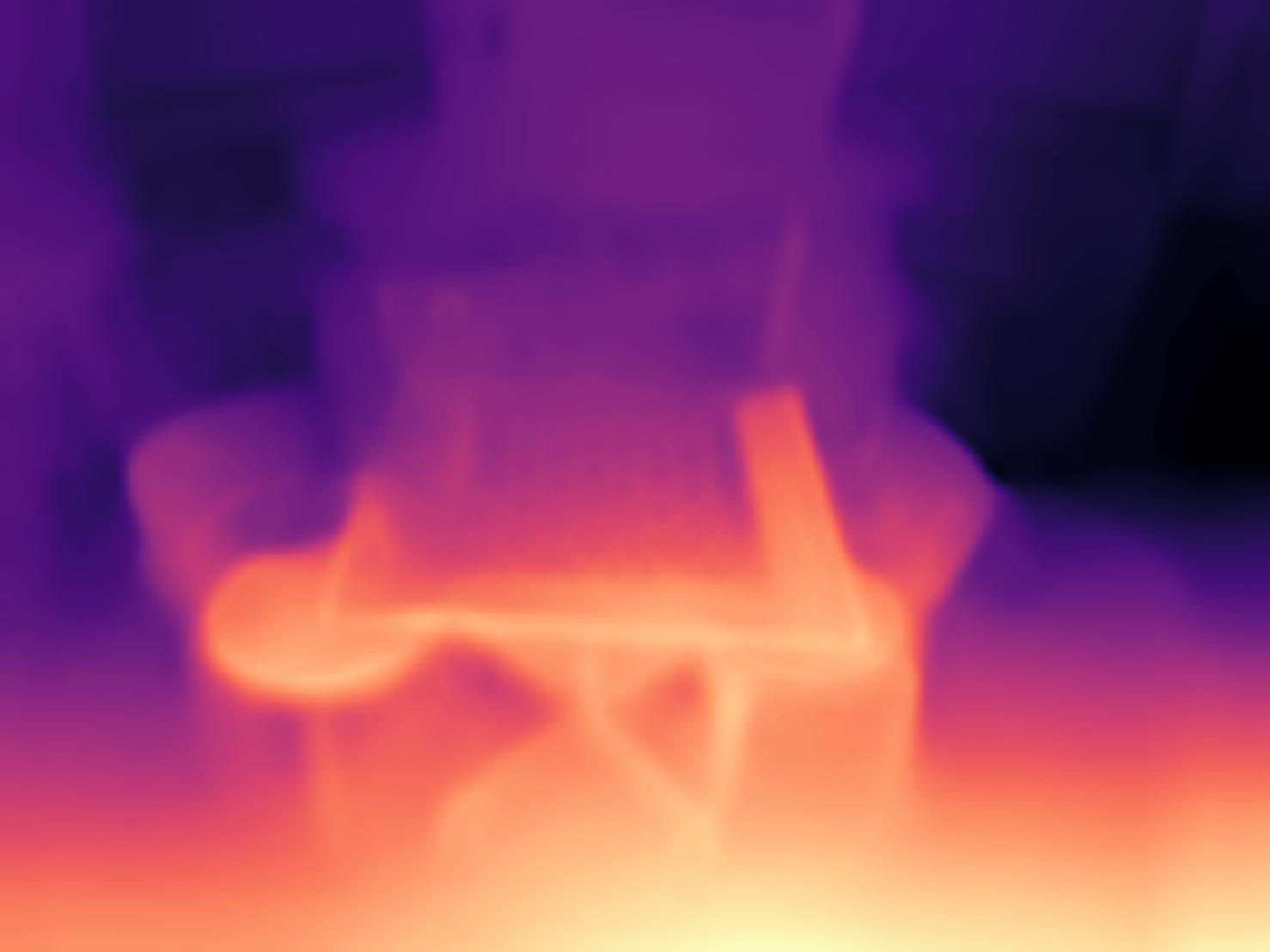}
    \includegraphics[width=0.242\textwidth]{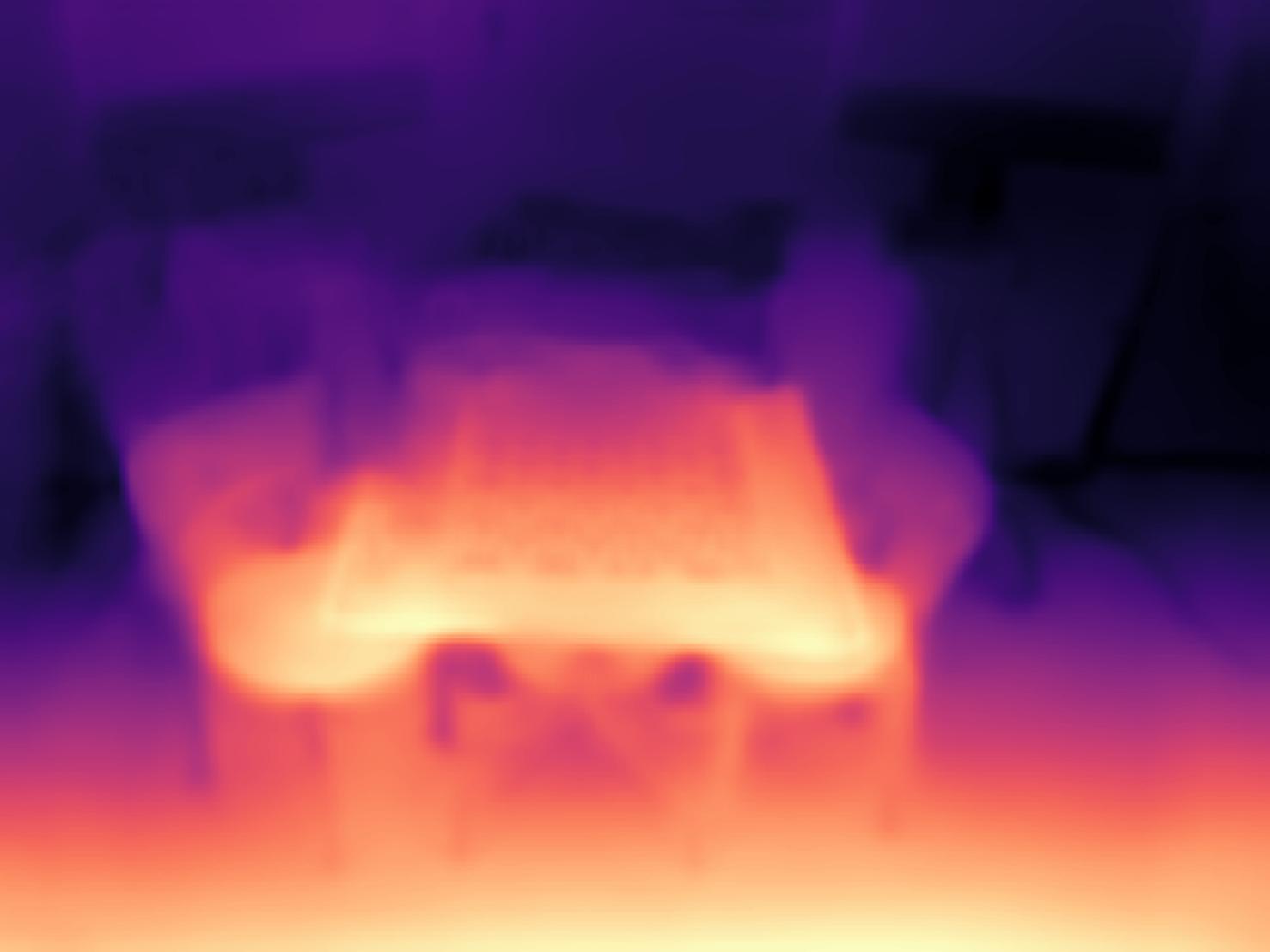}
    \includegraphics[width=0.242\textwidth]{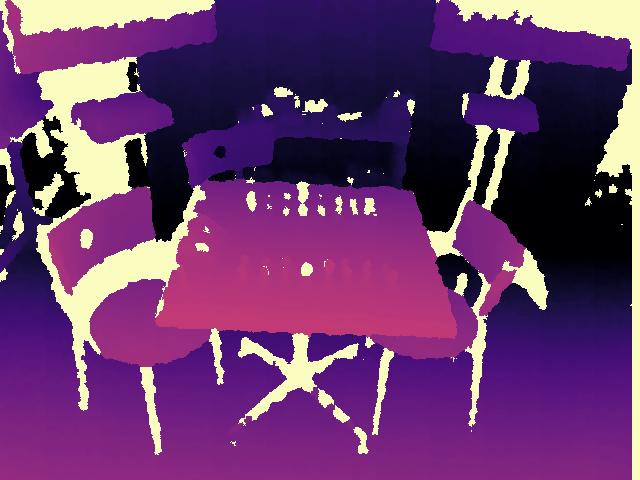}\\
    \vspace{-0.25cm}
    \begin{subfigure}[b]{0.242\textwidth}
        \caption{\textcolor{red}{RGB image}}
    \end{subfigure}
    \begin{subfigure}[b]{0.242\textwidth}
        \caption{\textcolor{red}{$\mathrm{P^2Net}$}}
    \end{subfigure}
    \begin{subfigure}[b]{0.242\textwidth}
        \caption{\textcolor{red}{Ours}}
    \end{subfigure}
    \begin{subfigure}[b]{0.242\textwidth}
        \caption{\textcolor{red}{Ground Truth}}
    \end{subfigure}\\
    \vspace{-0.25cm}
    \caption{\textcolor{red}{Qualitative zero-shot generalization results on 7-Scenes dataset. Left to right: RGB image, $\mathrm{P^2Net}$} \citep{Yu2020}\textcolor{red}{, ours and ground truth.}}
    \label{fig:fig_7scenes}
\end{figure}

\textcolor{red}{Zero-shot generalization experiments are performed on the public 7-Scenes dataset. The results are listed in} \autoref{tab:tab7scenes}. \textcolor{red}{From the table, we can see that $\mathrm{F^2Depth}$ outperforms $\mathrm{P^2Net}$ in most scenes and achieves better overall performance.}
\begin{table}[htp]
\color{red}

\begin{adjustbox}{center}
\footnotesize
\begin{tabular}[c]{>{\raggedright\arraybackslash}m{1.5cm} >{\raggedright\arraybackslash}m{0.7cm} >{\raggedright\arraybackslash}m{0.7cm} >{\raggedright\arraybackslash}M >{\raggedright\arraybackslash}M >{\raggedright\arraybackslash}M >{\raggedright\arraybackslash}m{0.7cm} >{\raggedright\arraybackslash}m{0.7cm} >{\raggedright\arraybackslash}M >{\raggedright\arraybackslash}M >{\raggedright\arraybackslash}M}
\hline

Method  &\multicolumn{5}{c}{$\mathrm{F^2Depth}$ + PP} & \multicolumn{5}{c}{$\mathrm{P^2Net}$ + PP\citep{Yu2020}}  \\
\hline
\makecell[l]{Scene} & \makecell[c]{REL\\↓} & \makecell[c]{RMS\\↓} & \makecell[c]{$\delta\!<\!1.25$\\↑} & \makecell[c]{$\delta\!<\!1.25^2$\\↑} & \makecell[c]{$\delta\!<\!1.25^3$\\↑} & \makecell[c]{REL\\↓} & \makecell[c]{RMS\\↓} & \makecell[c]{$\delta\!<\!1.25$\\↑} & \makecell[c]{$\delta\!<\!1.25^2$\\↑} & \makecell[c]{$\delta\!<\!1.25^3$\\↑}  \\
\hline
Chess&\textbf{0.18}&\textbf{0.4}&\textbf{0.689}&\textbf{0.942}&\textbf{0.994}&0.184&0.41&0.67&0.94&0.993\\
Fire&0.172&0.315&0.713&0.955&0.992&\textbf{0.159}&\textbf{0.294}&\textbf{0.763}&\textbf{0.964}&\textbf{0.994}\\
Heads&\textbf{0.18}&\textbf{0.19}&\textbf{0.725}&\textbf{0.93}&\textbf{0.987}&0.188&0.198&0.701&0.924&0.982\\
Office&\textbf{0.154}&\textbf{0.35}&\textbf{0.781}&\textbf{0.97}&\textbf{0.997}&0.156&\textbf{0.35}&0.774&0.97&\textbf{0.997}\\
Pumpkin&\textbf{0.12}&\textbf{0.335}&\textbf{0.86}&\textbf{0.981}&\textbf{0.99}6&0.141&0.378&0.798&0.977&0.995\\
RedKitchen&\textbf{0.162}&\textbf{0.396}&\textbf{0.751}&\textbf{0.954}&\textbf{0.995}&0.165&0.403&0.734&0.952&0.994\\
Stairs&0.162&0.468&0.754&\textbf{0.911}&0.967&\textbf{0.157}&\textbf{0.455}&\textbf{0.769}&\textbf{0.911}&\textbf{0.971}\\
Average&\textbf{0.159}&\textbf{0.361}&\textbf{0.758}&\textbf{0.956}&\textbf{0.993}&0.163&0.367&0.747&0.955&\textbf{0.993}\\

\hline
\end{tabular}

\end{adjustbox}
\caption{\textcolor{red}{Separate generalization results of 7 scenes. All models are trained on NYU Depth V2. We report average results of all scenes in the last row. PP denotes the results with post-processing. $\downarrow$ means the lower the better, $\uparrow$  means the higher the better.}}
\label{tab:tab7scenes}
\end{table}
\autoref{fig:fig_7scenes} \textcolor{red}{shows several qualitative generalization results on 7-Scenes. The visualization maps indicate that our $\mathrm{F^2Depth}$ generalizes better than $\mathrm{P^2Net}$. Especially in low-textured regions such as walls and cupboard doors, our method is significantly better.}

\subsubsection{Zero-shot generalization results on Campus Indoor}

\textcolor{red}{Besides 7-Scenes,} we also run generalization experiments on our Campus Indoor dataset. Results are shown in \autoref{tab:tab3}. Through comparing with evaluation results on NYU Depth V2 in the last row, we can see that our method can generalize well to monocular datasets captured in unknown indoor scenes.

As shown in \autoref{tab:tab3}, our method outperforms other self-supervised methods in generalization experiments. In \autoref{tab:tab4}, we list generalization results of $\mathrm{P^2Net}$ \citep{Yu2020} and our method in all 18 scenes. It is shown that in 13 scenes our method outperforms $\mathrm{P^2Net}$, while in the remaining 5 scenes $\mathrm{P^2Net}$ achieves better performance than us. Generally, it can be considered that our method has better generalization ability than $\mathrm{P^2Net}$.

\begin{table}[h]

\footnotesize
\setlength\tabcolsep{4pt}
\begin{tabular}[c]{>{\raggedright\arraybackslash}m{3.9cm} c c c c c c c}
\hline
Method & S & REL$\downarrow$ & RMS$\downarrow$ & \makecell{log10$\downarrow$} & \makecell{$\delta\!<\!1.25$$\uparrow$} & \makecell{$\delta\!<\!1.25^2$$\uparrow$} &\makecell{$\delta\!<\!1.25^3$$\uparrow$}  \\
\hline
VNL \citep{Yin2019}& \checkmark & 0.132 & 0.586 & 0.054 & 0.852 & \textbf{0.978} & 0.995 \\
Jun \citep{Jun2022}& \checkmark & \textbf{0.116} & \textbf{0.525} & \textbf{0.048} & \textbf{0.867} & 0.977 & \textbf{0.999} \\
\hline
TrianFlow \citep{Zhao2020}& × & 0.204 & 0.803 & 0.080 & 0.695 & 0.910 & 0.975 \\
$\mathrm{P^2Net}$ \citep{Yu2020}& × & 0.174 & 0.783 & 0.072 & 0.733 & 0.939 & \textbf{0.991} \\
\textcolor{red}{$\mathrm{F^2Depth}$} & × & \textbf{0.164} & \textbf{0.753} & \textbf{0.069} & \textbf{0.760} & \textbf{0.950} & \textbf{0.991} \\
\textcolor{red}{$\mathrm{F^2Depth}$} on NYU & × & 0.158 & 0.583 & 0.067 & 0.779 & 0.947 & 0.987 \\
\hline
\end{tabular}
\caption{Comparison of zero-shot generalization results between our method and other methods. All models are trained on NYU Depth V2. We report average results of 18 scenes. S represents supervision. $\downarrow$ means the lower the better, $\uparrow$  means the higher the better.}
\label{tab:tab3}
\end{table}

\begin{table}[htp]

\begin{adjustbox}{center}
\footnotesize
\begin{tabular}{>{\raggedright\arraybackslash}M >{\raggedright\arraybackslash}m{0.7cm} >{\raggedright\arraybackslash}m{0.7cm} >{\raggedright\arraybackslash}m{0.7cm} >{\raggedright\arraybackslash}M >{\raggedright\arraybackslash}M >{\raggedright\arraybackslash}M >{\raggedright\arraybackslash}m{0.7cm} >{\raggedright\arraybackslash}m{0.7cm} >{\raggedright\arraybackslash}m{0.7cm} >{\raggedright\arraybackslash}M >{\raggedright\arraybackslash}M >{\raggedright\arraybackslash}M}
\hline
Method  &\multicolumn{6}{c}{\textcolor{red}{$\mathrm{F^2Depth}$}} & \multicolumn{6}{c}{$\mathrm{P^2Net}$\citep{Yu2020}}  \\

\hline
\makecell[l]{Scene\\No.} & \makecell[c]{REL\\↓} & \makecell[c]{RMS\\↓} & \makecell[c]{log10\\↓} & \makecell[c]{$\delta\!<\!1.25$\\↑} & \makecell[c]{$\delta\!<\!1.25^2$\\↑} & \makecell[c]{$\delta\!<\!1.25^3$\\↑} & \makecell[c]{REL\\↓} & \makecell[c]{RMS\\↓} & \makecell[c]{log10\\↓} & \makecell[c]{$\delta\!<\!1.25$\\↑} & \makecell[c]{$\delta\!<\!1.25^2$\\↑} & \makecell[c]{$\delta\!<\!1.25^3$\\↑}  \\
\hline
1 & 0.164 & \textbf{0.592} & \textbf{0.064} & \textbf{0.766} & 0.947 & 1 & \textbf{0.161} & 0.663 & 0.067 & 0.676 & \textbf{1} & \textbf{1} \\
2 & \textbf{0.135} & \textbf{0.949} & \textbf{0.060} & \textbf{0.798} & \textbf{0.981} & \textbf{1} & 0.171 & 1.036 & 0.074 & 0.726 & 0.968 & \textbf{1} \\
3 & \textbf{0.215} & \textbf{0.889} & \textbf{0.084} & 0.644 & \textbf{0.973} & \textbf{1} & 0.254 & 0.956 & 0.093 & \textbf{0.684} & 0.822 & \textbf{1} \\
4 & \textbf{0.134} & \textbf{0.588} & \textbf{0.058} & \textbf{0.777} & \textbf{1} & \textbf{1} & 0.169 & 0.645 & 0.068 & 0.709 & 0.986 & \textbf{1} \\
5 & \textbf{0.159} & 1.091 & \textbf{0.067} & \textbf{0.746} & \textbf{0.946} & \textbf{1} & 0.172 & \textbf{1.051} & 0.070 & 0.733 & 0.893 & \textbf{1} \\
6 & 0.154 & 1.625 & 0.073 & 0.757 & 0.919 & 0.932 & \textbf{0.140} & \textbf{1.544} & \textbf{0.068} & \textbf{0.812} & \textbf{0.932} & \textbf{0.959} \\
7 & 0.266 & 2.321 & 0.122 & 0.436 & 0.802 & 0.96 & \textbf{0.237} & \textbf{2.104} & \textbf{0.108} & \textbf{0.528} & \textbf{0.817} & \textbf{0.961} \\
8 & \textbf{0.149} & \textbf{0.699} & 0.063 & \textbf{0.772} & 0.929 & \textbf{1} & 0.154 & 0.778 & \textbf{0.062} & 0.77 & \textbf{0.943} & \textbf{1} \\
9 & \textbf{0.088} & \textbf{0.372} & \textbf{0.037} & \textbf{0.946} & \textbf{1} & \textbf{1} & 0.122 & 0.514 & 0.050 & 0.88 & \textbf{1} & \textbf{1} \\
10 & 0.193 & 0.686 & 0.076 & 0.717 & 0.917 & \textbf{1} & \textbf{0.183} & \textbf{0.637} & \textbf{0.069} & \textbf{0.733} & \textbf{0.933} & \textbf{1} \\
11 & \textbf{0.168} & \textbf{0.667} & \textbf{0.072} & \textbf{0.787} & \textbf{0.973} & \textbf{1} & 0.181 & 0.784 & 0.078 & 0.680 & 0.960 & \textbf{1} \\
12 & 0.143 & \textbf{0.536} & 0.066 & 0.736 & \textbf{1} & \textbf{1} & 0.143 & 0.545 & \textbf{0.065} & \textbf{0.775} & \textbf{1} & \textbf{1} \\
13 & 0.147 & \textbf{0.594} & 0.067 & \textbf{0.803} & \textbf{0.974} & \textbf{1} & \textbf{0.142} & 0.613 & \textbf{0.066} & 0.802 & 0.962 & \textbf{1} \\
14 & \textbf{0.135} & \textbf{0.422} & \textbf{0.060} & \textbf{0.88}0 & \textbf{1} & \textbf{1} & 0.158 & 0.526 & 0.072 & 0.667 & \textbf{1} & \textbf{1} \\
15 & \textbf{0.236} & \textbf{0.409} & \textbf{0.084} & \textbf{0.720} & 0.867 & 0.946 & 0.253 & 0.450 & 0.092 & 0.640 & 0.867 & 0.946\\
16 & \textbf{0.202} & \textbf{0.647} & \textbf{0.079} & \textbf{0.707} & \textbf{0.920} & \textbf{1} & 0.233 & 0.773 & 0.090 & 0.653 & 0.880 & 0.987\\
17 & \textbf{0.114} & \textbf{0.244} & \textbf{0.047} & 0.893 & \textbf{0.960} & \textbf{1} & 0.129 & 0.268 & 0.050 & \textbf{0.933} & 0.947 & 0.987\\
18 & 0.144 & 0.230 & 0.061 & 0.787 & 0.987 & \textbf{1} & \textbf{0.136} & \textbf{0.209} & \textbf{0.058} & 0.787 & \textbf{1} & \textbf{1}\\

\hline
\end{tabular}

\end{adjustbox}
\caption{Separate generalization results of 18 scenes. All models are trained on NYU Depth V2. We report average result of images in every scene. $\downarrow$ means the lower the better, $\uparrow$  means the higher the better.}
\label{tab:tab4}
\end{table}

\begin{figure}[h]
    \centering
    \includegraphics[width=0.23\textwidth]{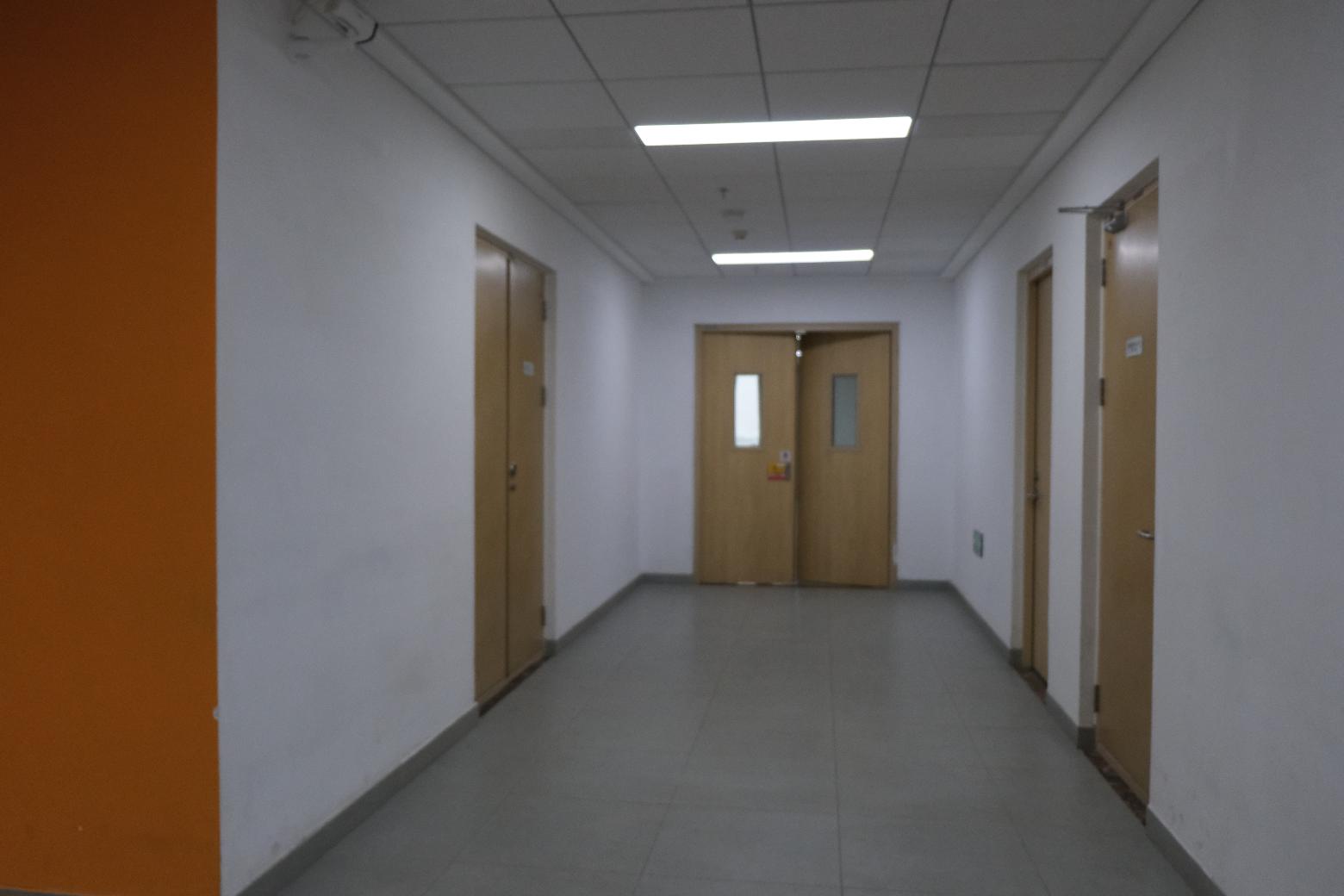}
    \includegraphics[width=0.23\textwidth]{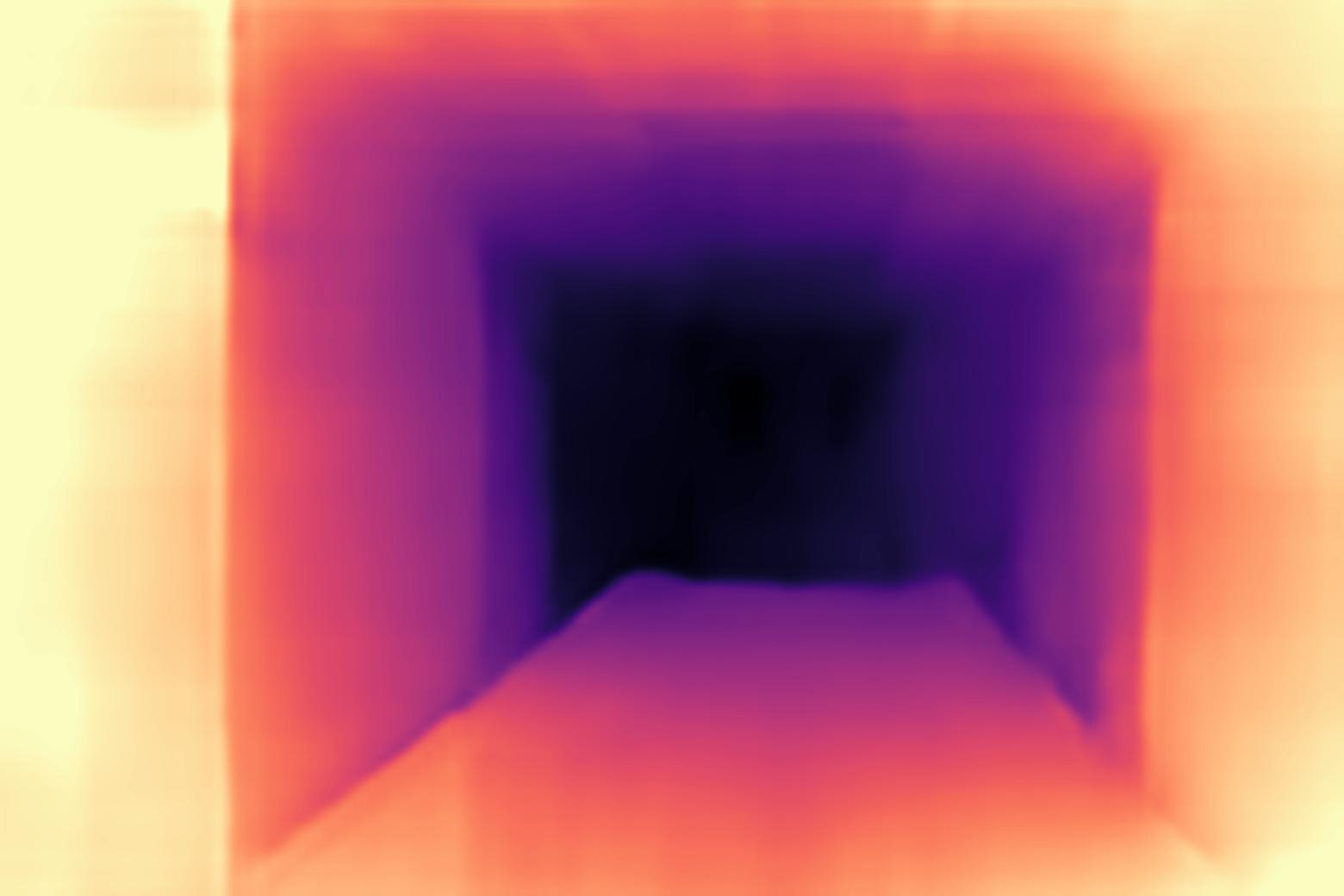}
    \includegraphics[width=0.23\textwidth]{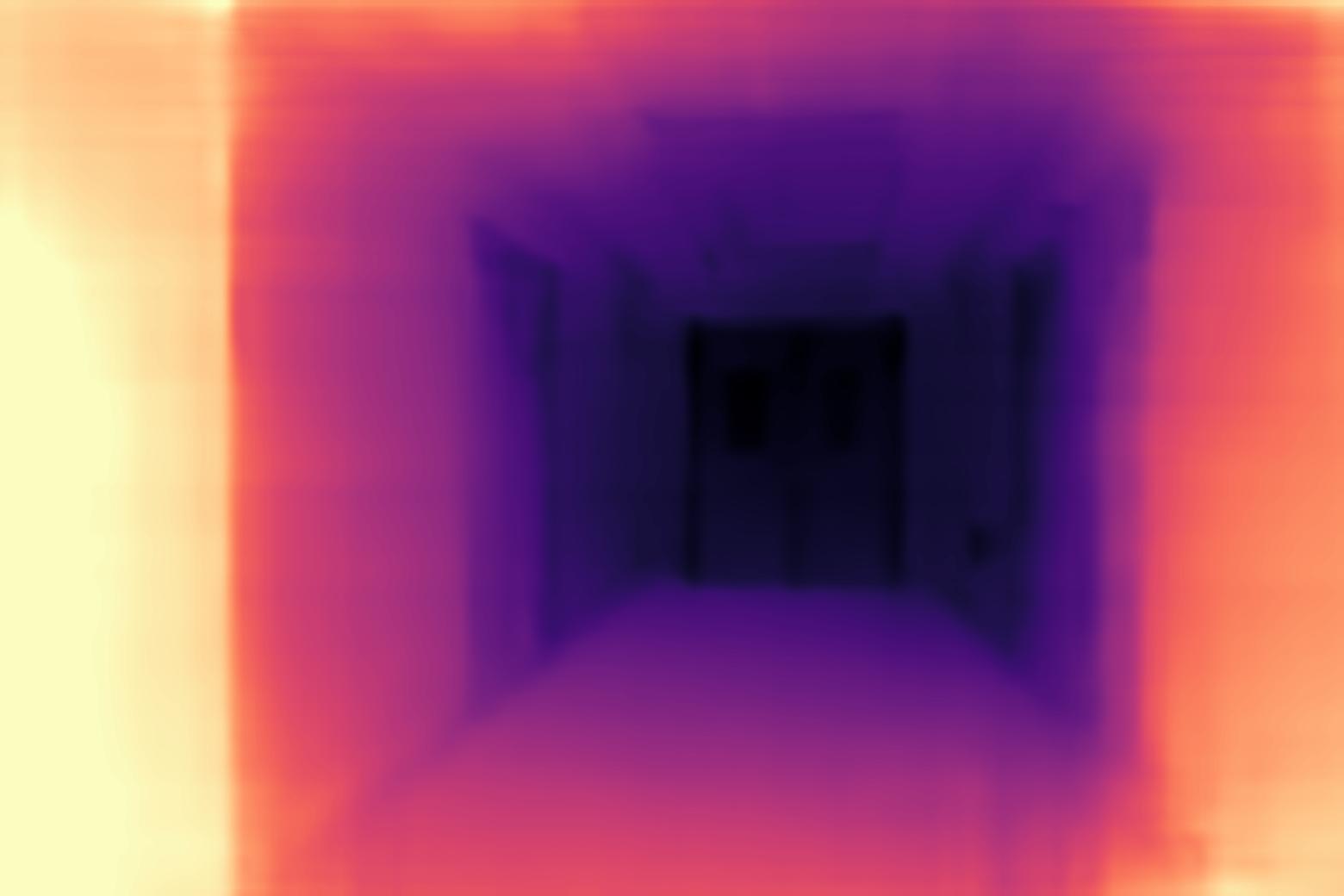}\\    
    \includegraphics[width=0.23\textwidth]{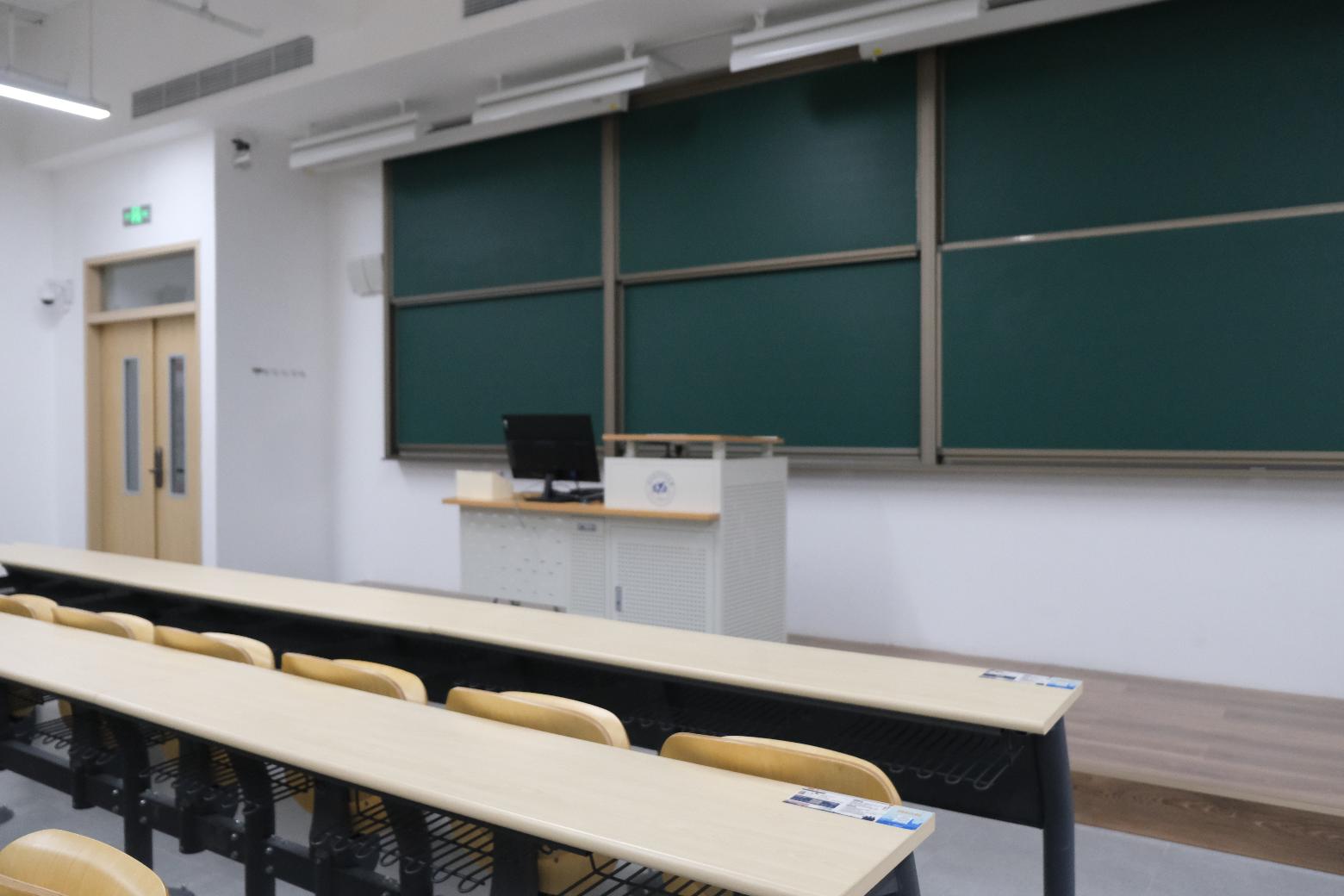}
    \includegraphics[width=0.23\textwidth]{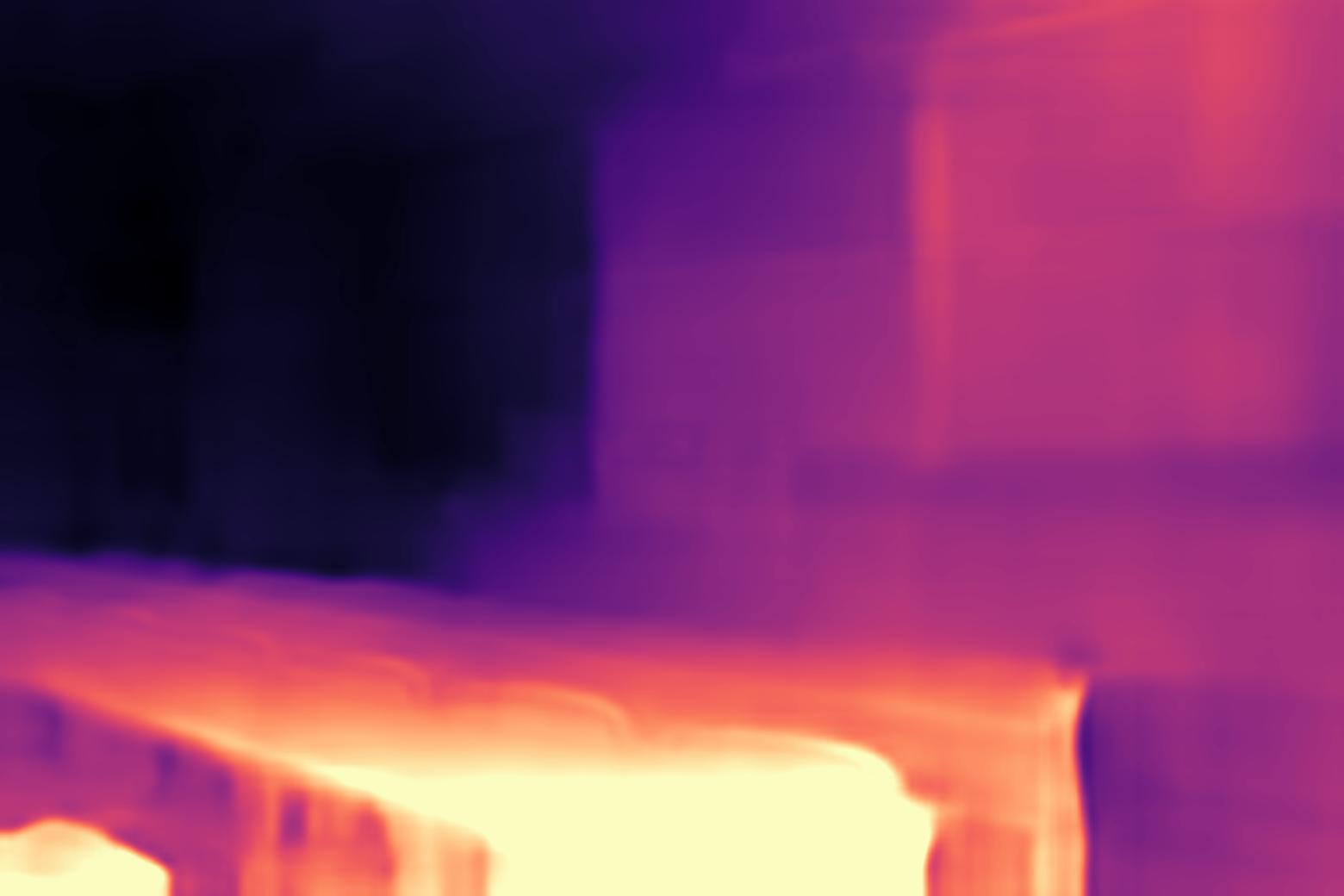}
    \includegraphics[width=0.23\textwidth]{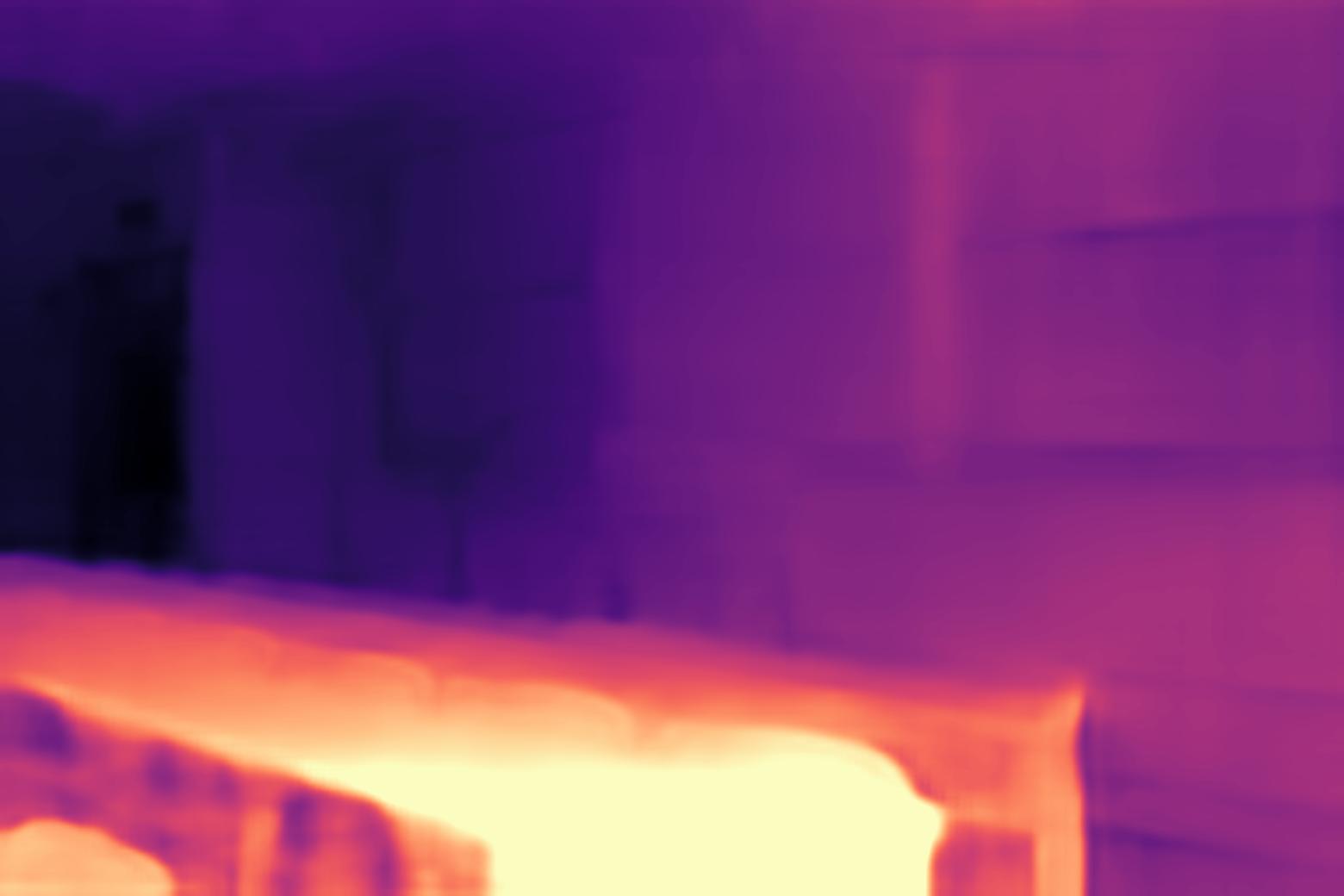}\\
    \includegraphics[width=0.23\textwidth]{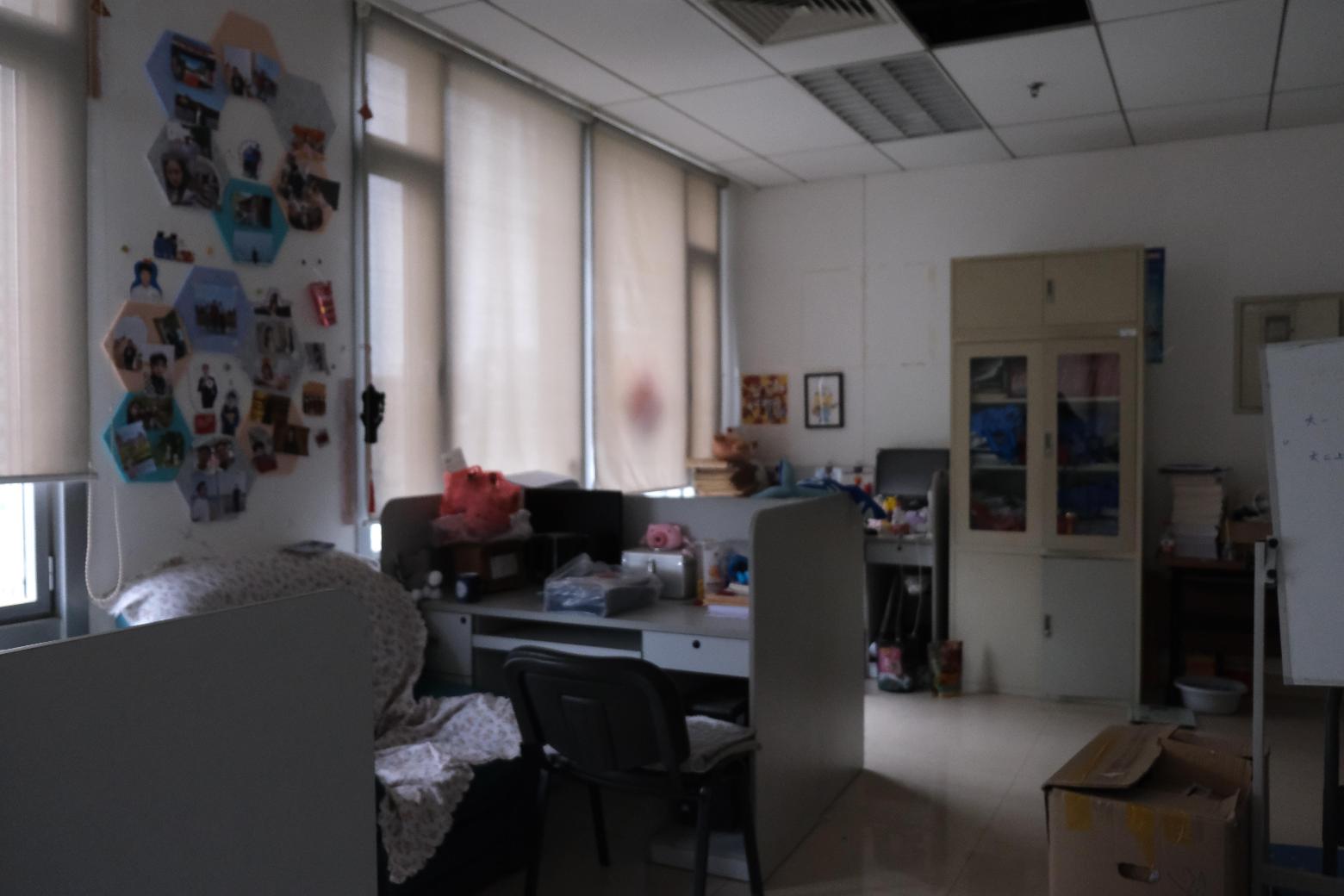}
    \includegraphics[width=0.23\textwidth]{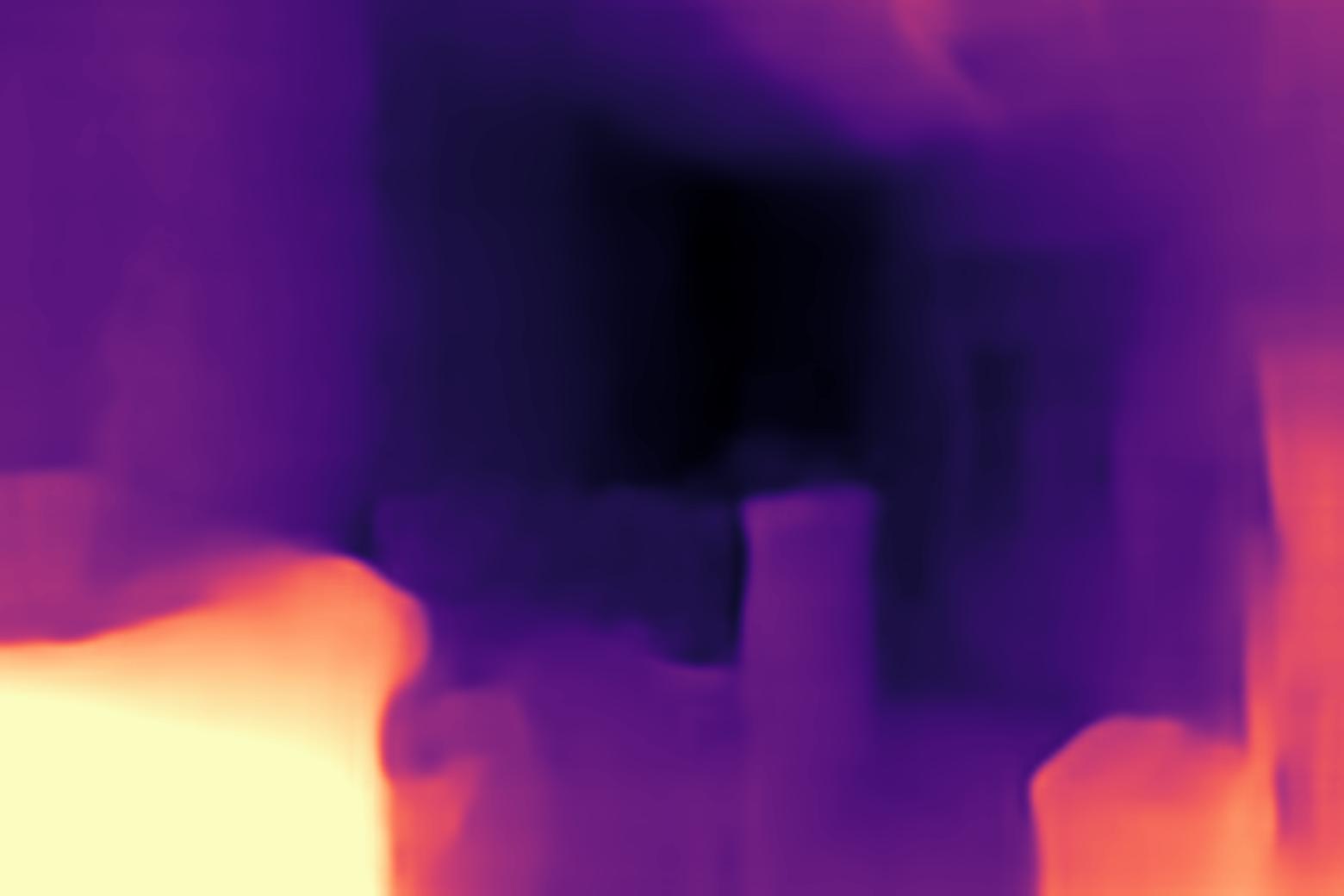}
    \includegraphics[width=0.23\textwidth]{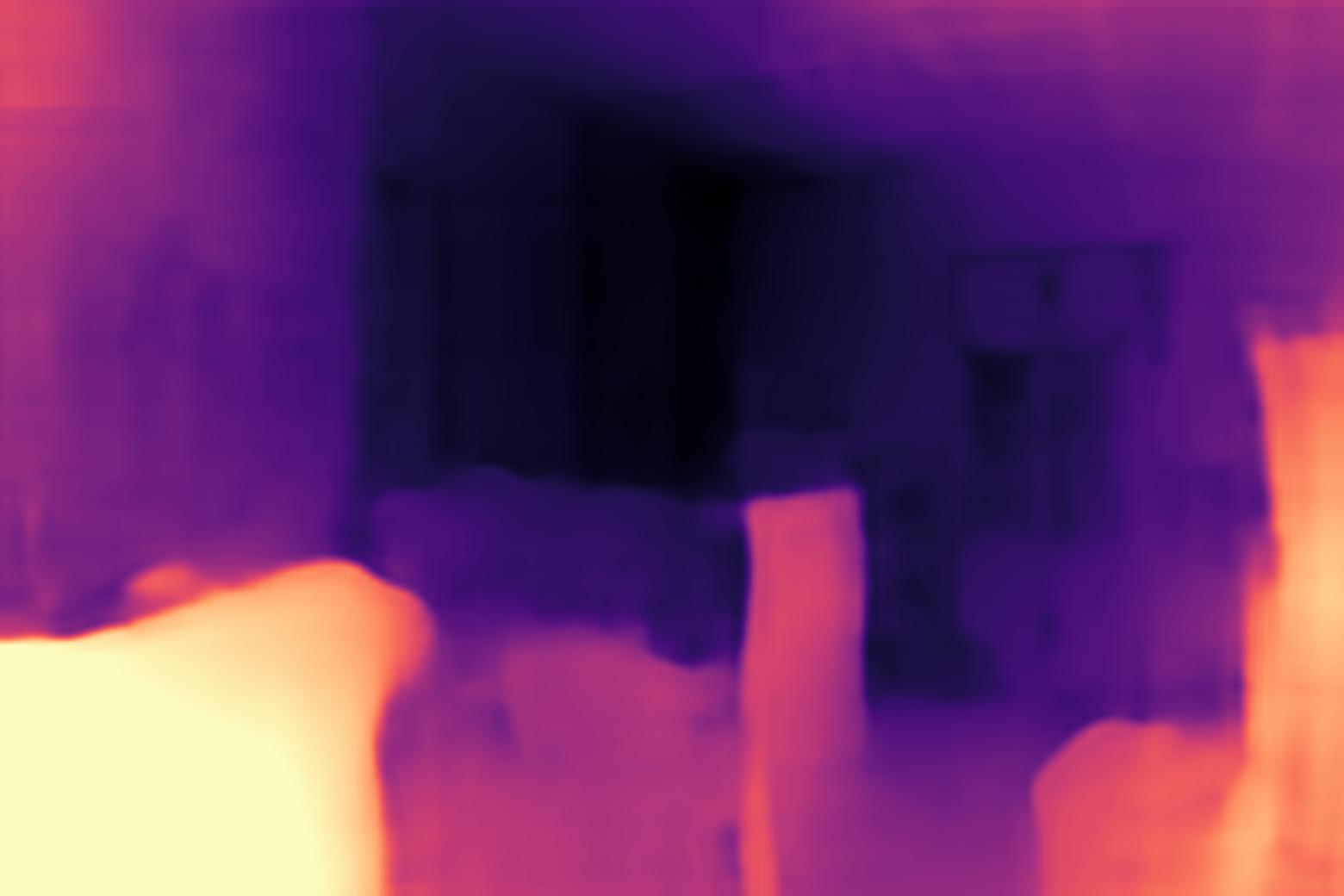}\\
    \includegraphics[width=0.23\textwidth]{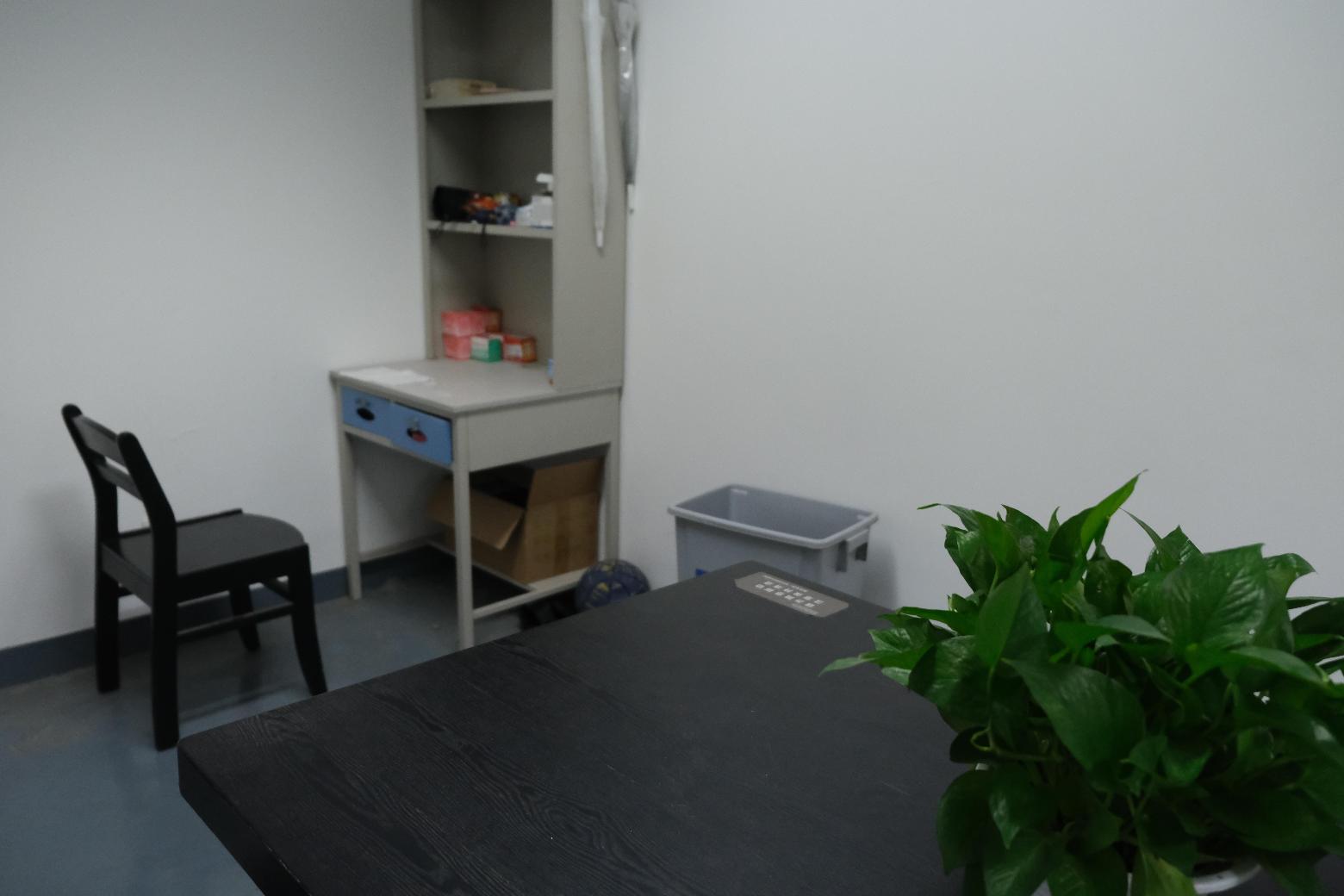}
    \includegraphics[width=0.23\textwidth]{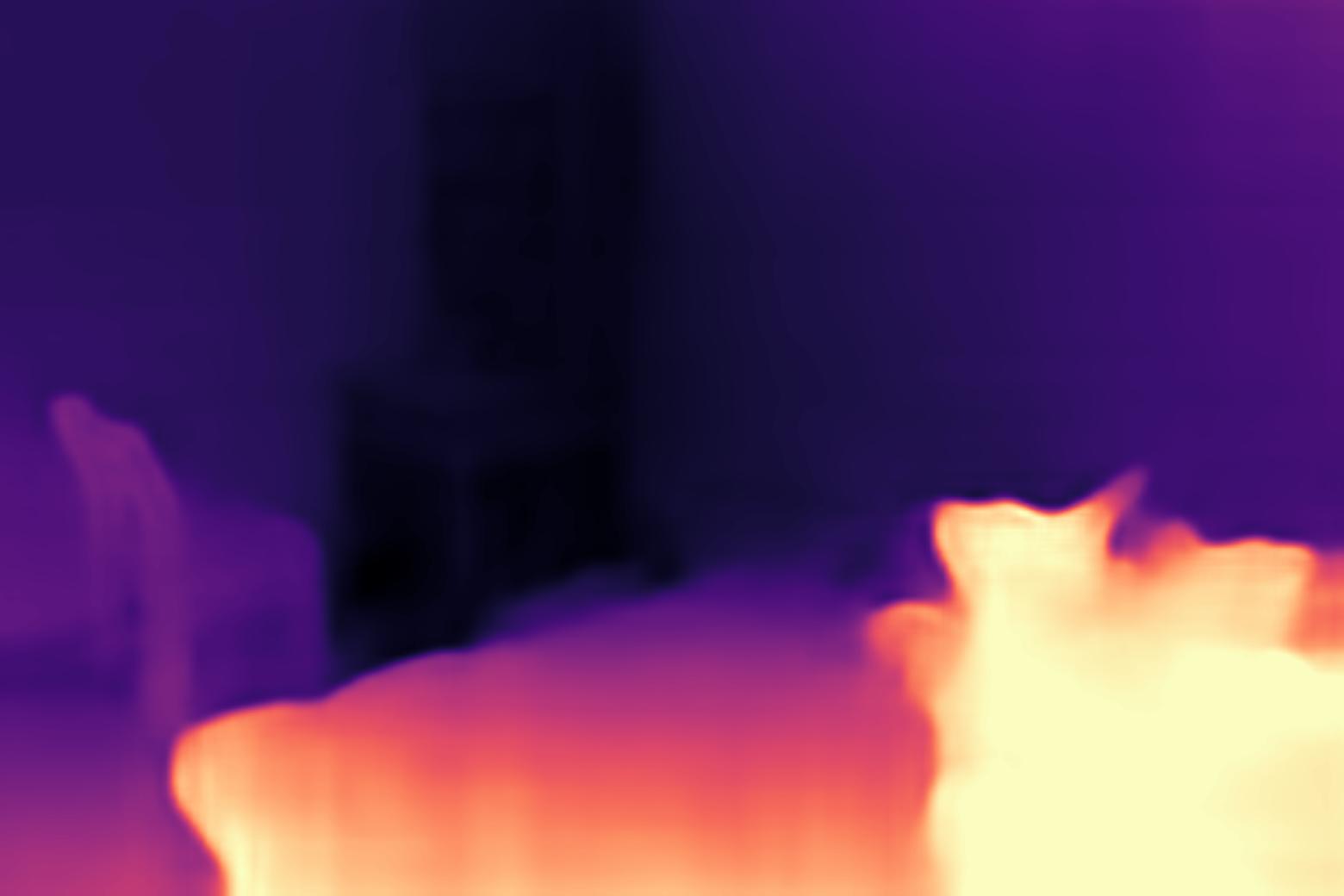}
    \includegraphics[width=0.23\textwidth]{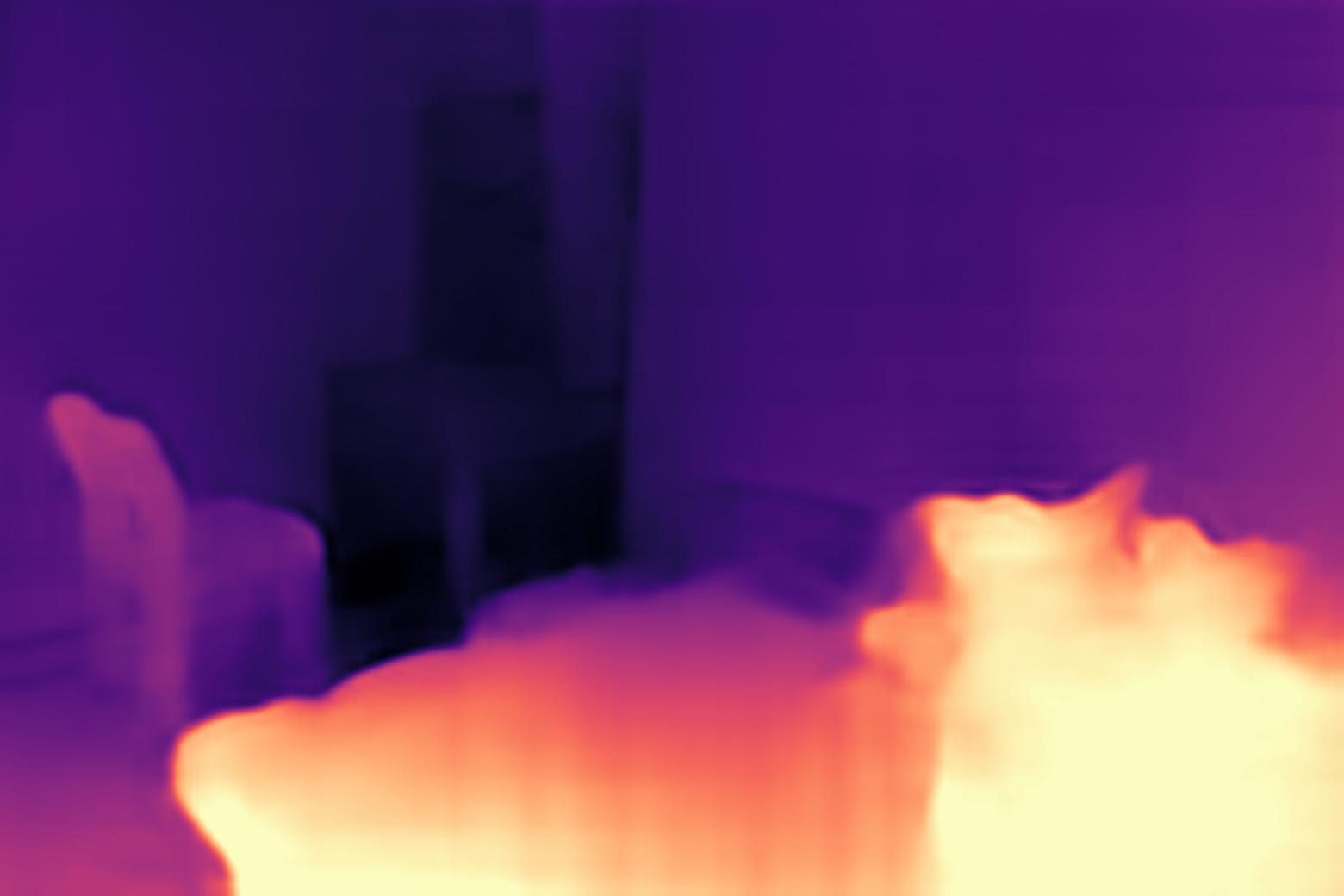}\\
    \vspace{-0.25cm}
    \begin{subfigure}[b]{0.23\textwidth}
        \caption{\textcolor{red}{RGB Image}}
    \end{subfigure}
    \begin{subfigure}[b]{0.23\textwidth}
        \caption{\textcolor{red}{$\mathrm{P^2Net}$}}
    \end{subfigure}
    \begin{subfigure}[b]{0.23\textwidth}
        \caption{\textcolor{red}{Ours}}
    \end{subfigure}\\
    \vspace{-0.25cm}
    \caption{Qualitative zero-shot generalization results. Left to right: RGB image, $\mathrm{P^{2}Net}$ \citep{Yu2020} and ours.}
    \label{fig:fig_7}
\end{figure}

We select several scenes in which our method outperforms $\mathrm{P^{2}Net}$ \citep{Yu2020} significantly including No. 2, 3, 4, 9, 11, 14, 15 to analyze the characteristics of these scenes. We find out that the scenes are mainly made up of low-textured areas such as walls, tables and doors. In \autoref{fig:fig_7}, four scenes are picked out to show generalization results in comparison with $\mathrm{P^2Net}$ in this kind of scenes. 

\textcolor{red}{In the scenes where $\mathrm{P^{2}Net}$ \citep{Yu2020} achieves better generalization performance than ours, such as No. 7, the darker areas lead to large performance gap. The phenomenon that darker areas are usually predicted deeper is more prominent in $\mathrm{F^2Depth}$ than $\mathrm{P^{2}Net}$. As shown in \autoref{fig:fig_7scenes_cmp}, the framed stair area becomes darker from left to right because of being gradually away from windows. Compared with $\mathrm{P^{2}Net}$, $\mathrm{F^2Depth}$ tends to predict more areas as deeper. For the reason, we suppose that the addition of optical flow learning makes $\mathrm{F^2Depth}$ more sensitive to illumination changes.}

\newdimen\ssizew
\newdimen\ssizeh
\ssizew=0.185\textwidth
\ssizeh=0.66667\ssizew

\begin{figure}[H]
    \centering
    \captionsetup[subfigure]{labelformat=empty}
    \begin{adjustbox}{angle=90}
    \begin{subfigure}[b]{\ssizeh}
        \vspace{-10pt}
        \caption{\textcolor{red}{RGB}}
    \end{subfigure}
    \end{adjustbox}
    \includegraphics[width=\ssizew]{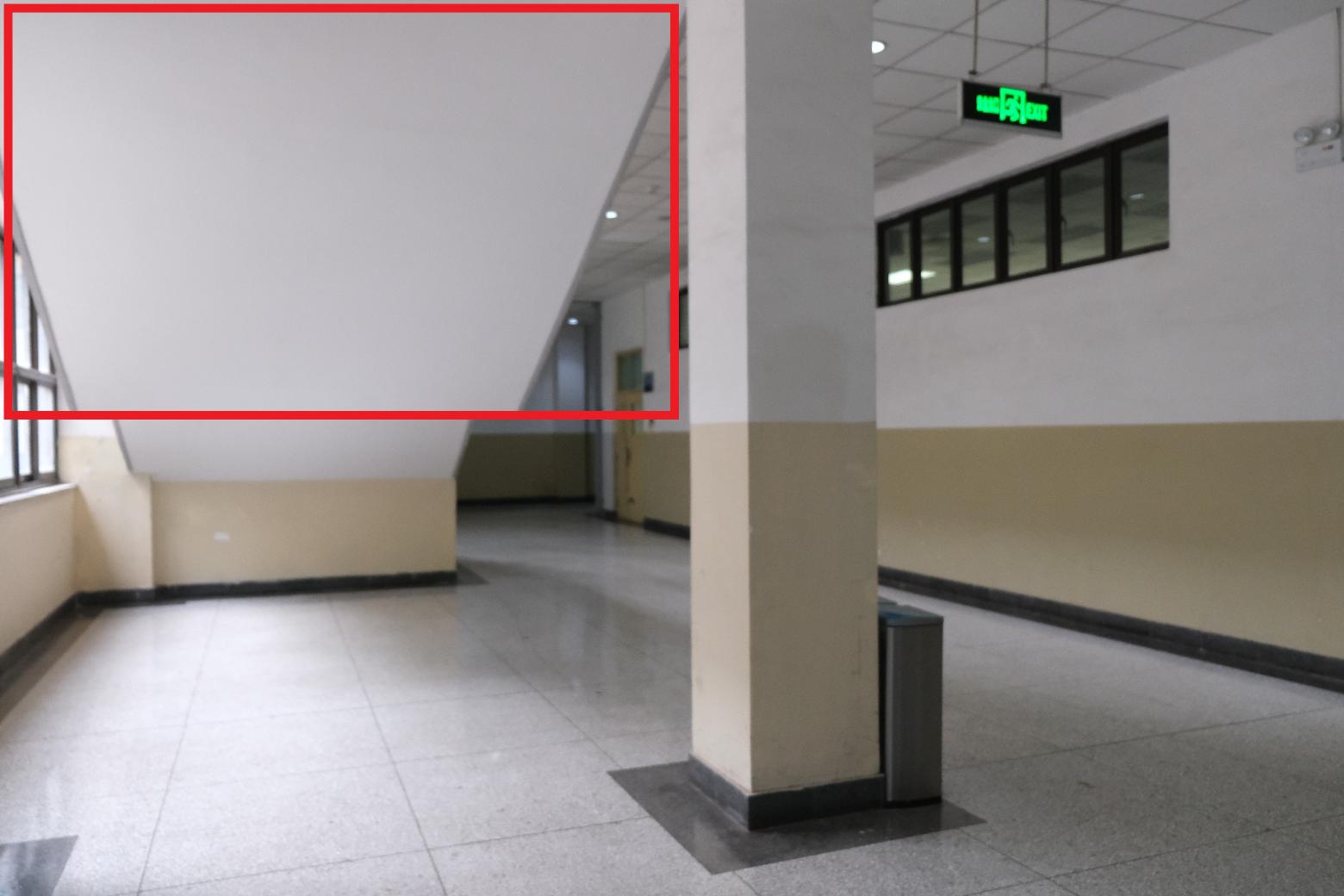}
    \includegraphics[width=\ssizew]{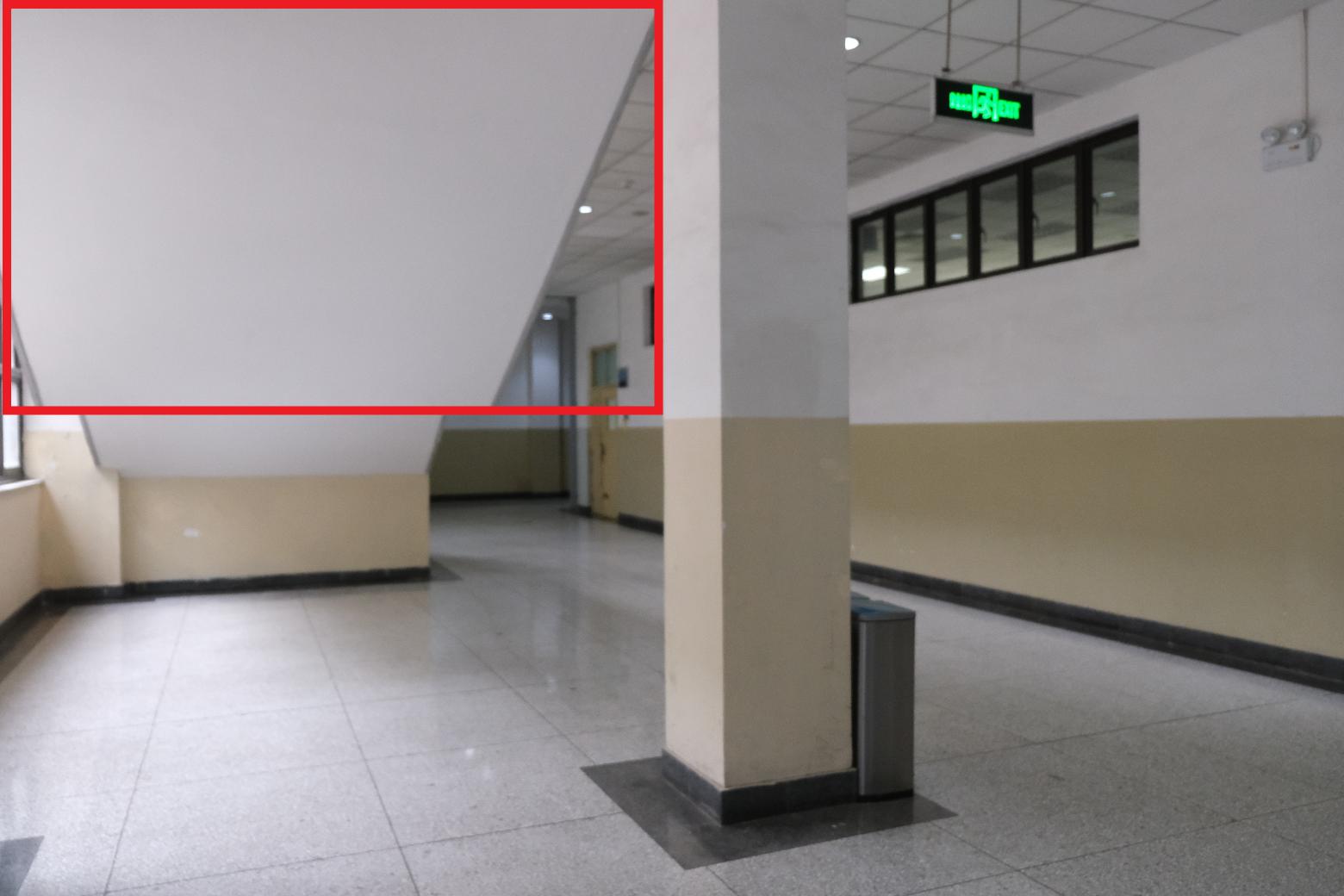}
    \includegraphics[width=\ssizew]{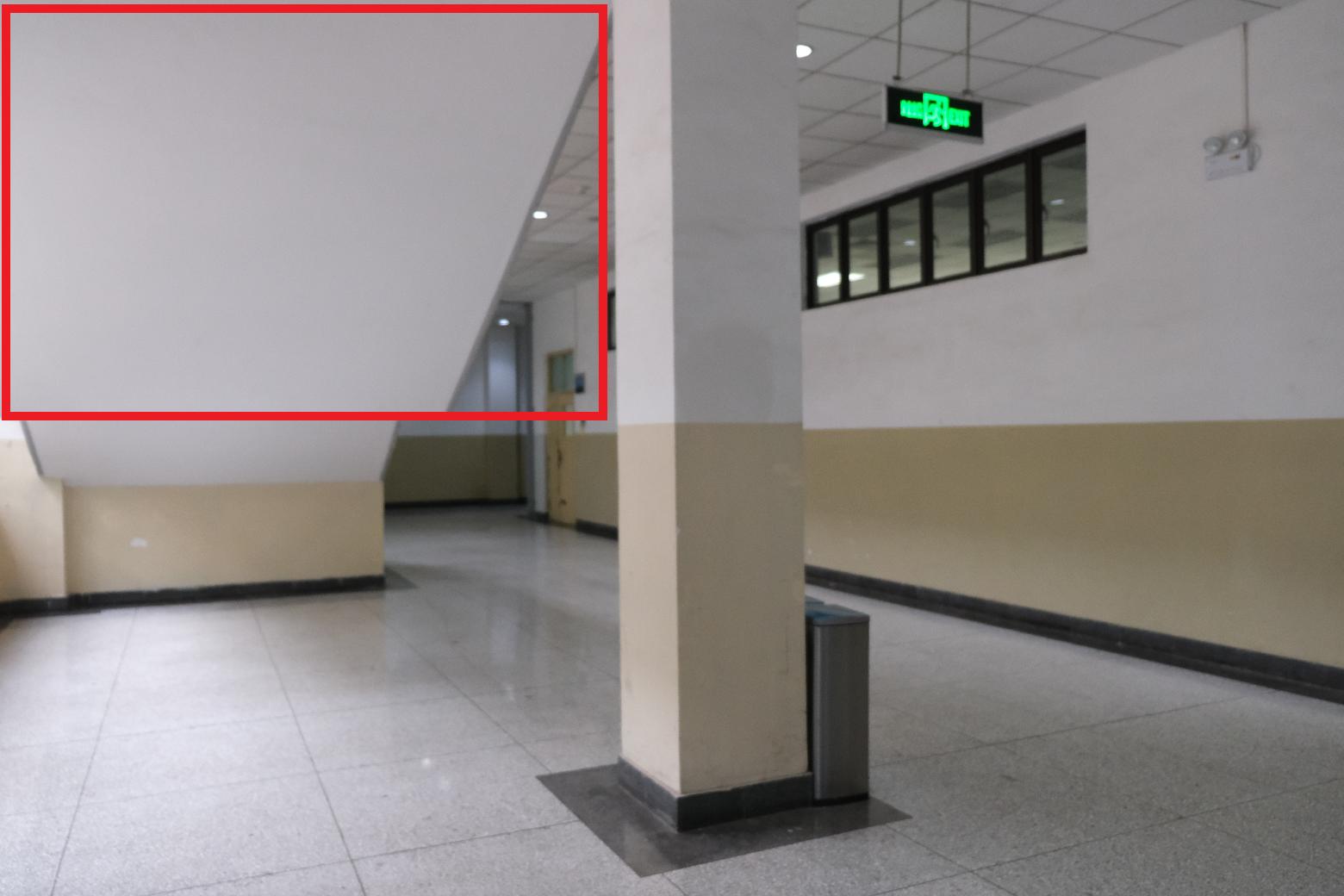}
    \includegraphics[width=\ssizew]{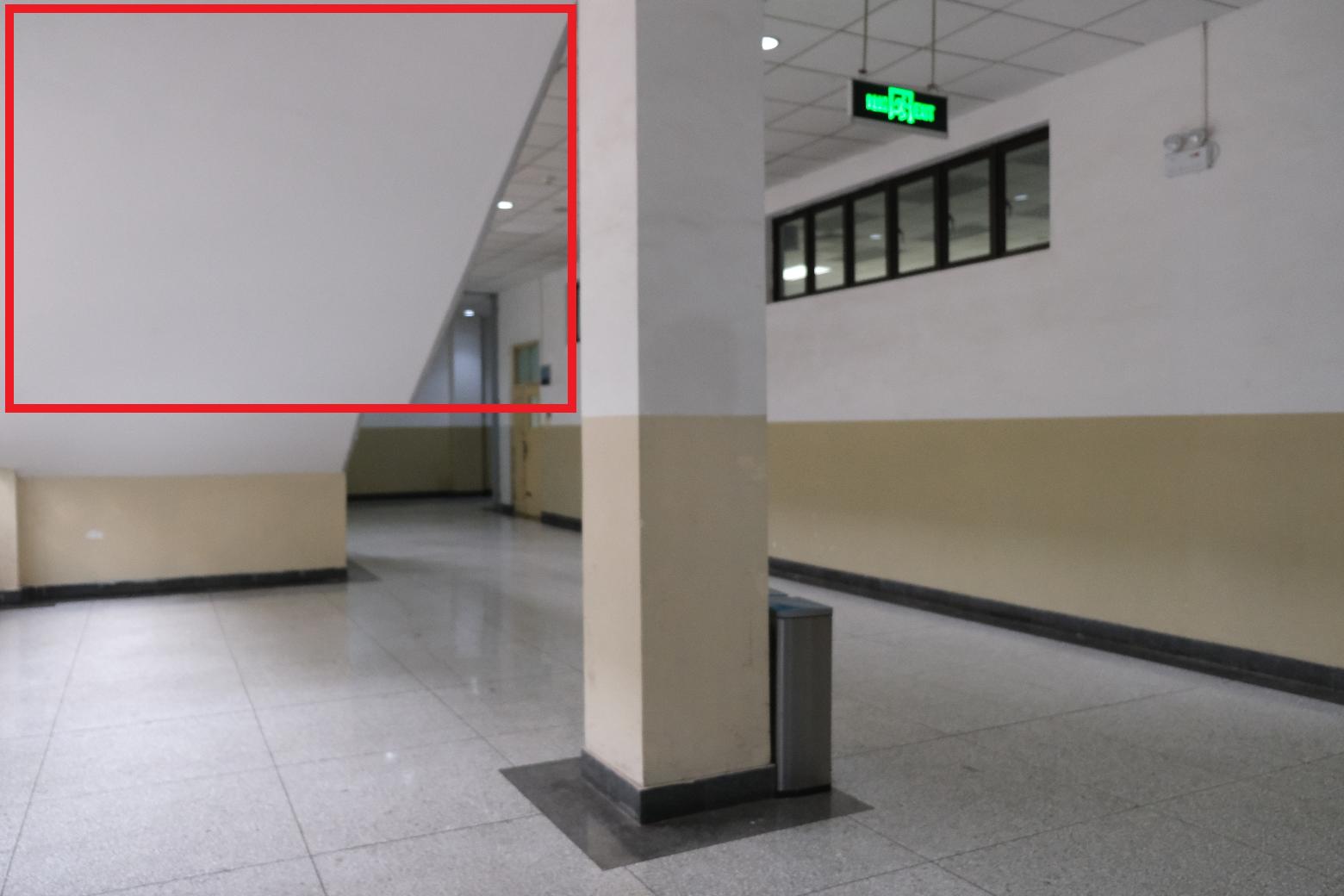}
    \includegraphics[width=\ssizew]{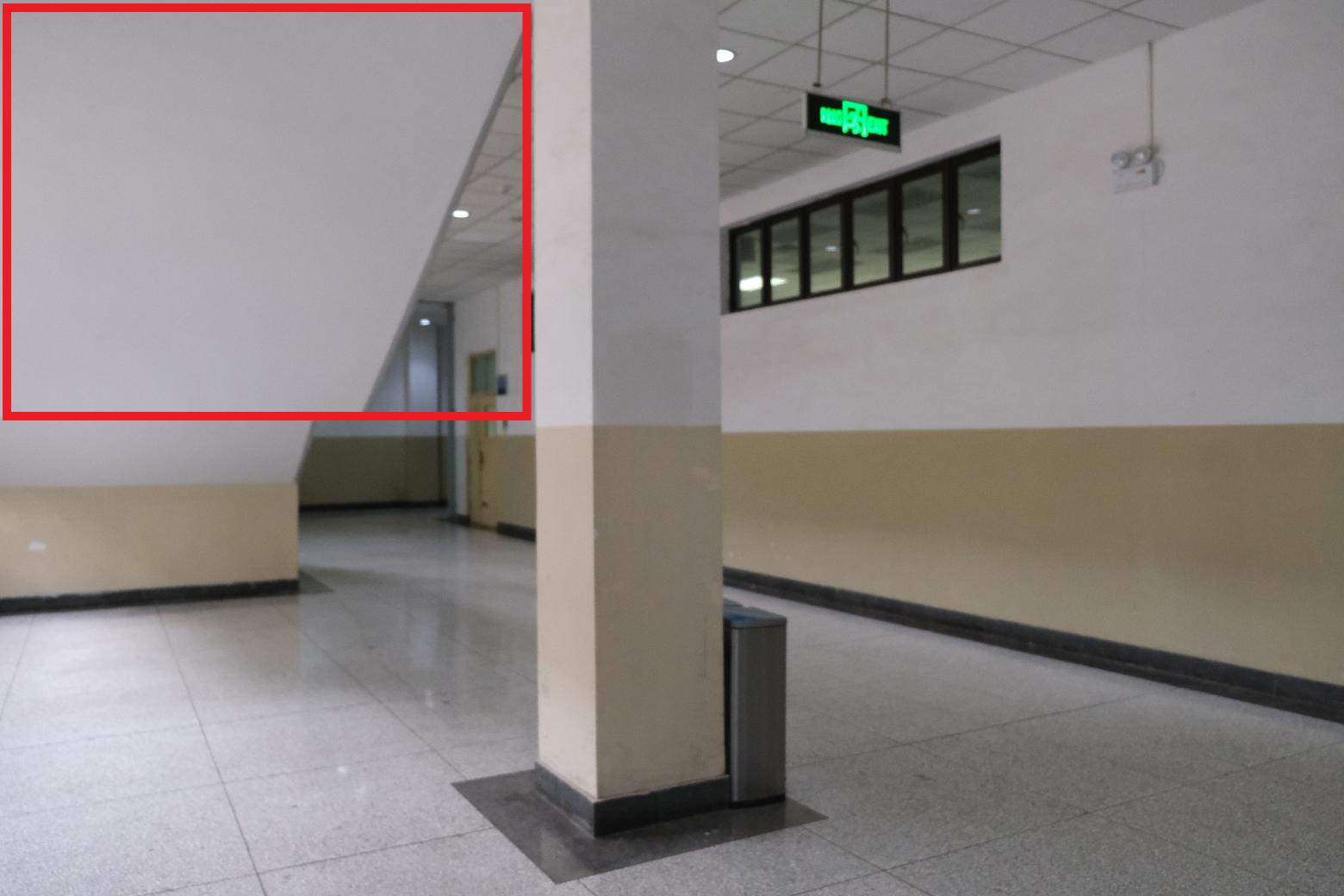}

    \begin{adjustbox}{angle=90}
    \begin{subfigure}[b]{\ssizeh}
        \vspace{-10pt}
        \caption{\textcolor{red}{$\mathrm{P^2Net}$}}
    \end{subfigure}
    \end{adjustbox}
    \includegraphics[width=\ssizew]{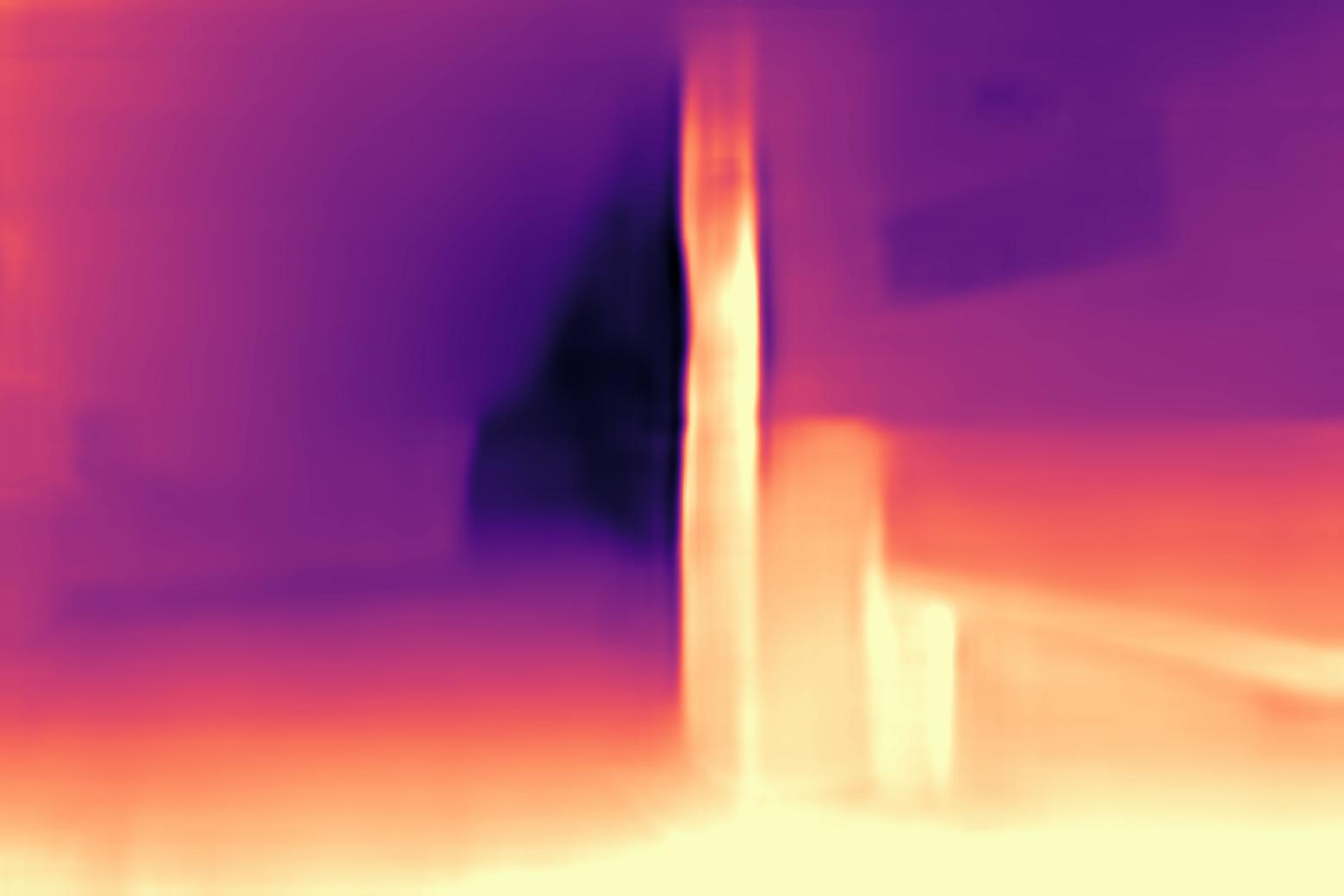}
    \includegraphics[width=\ssizew]{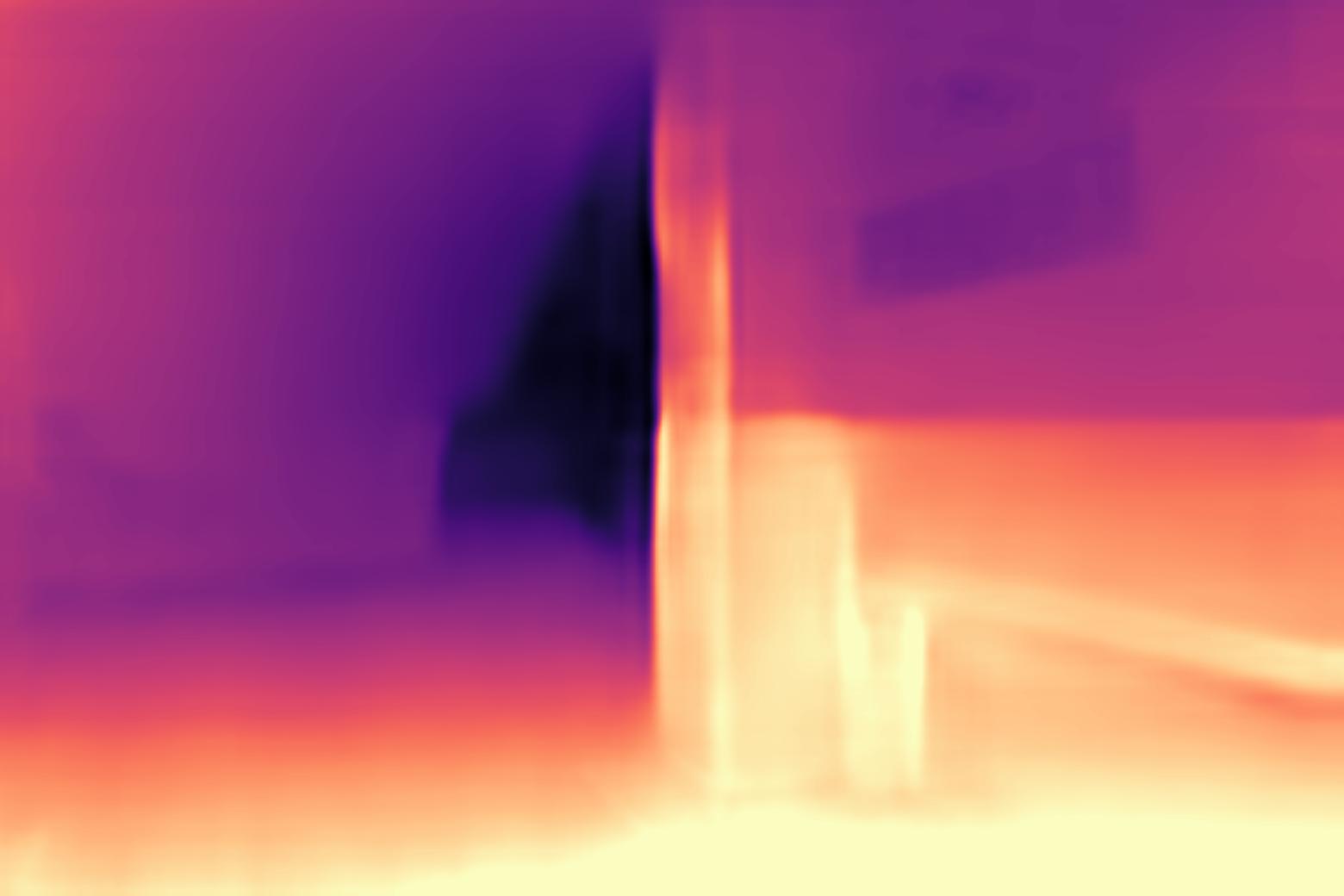}
    \includegraphics[width=\ssizew]{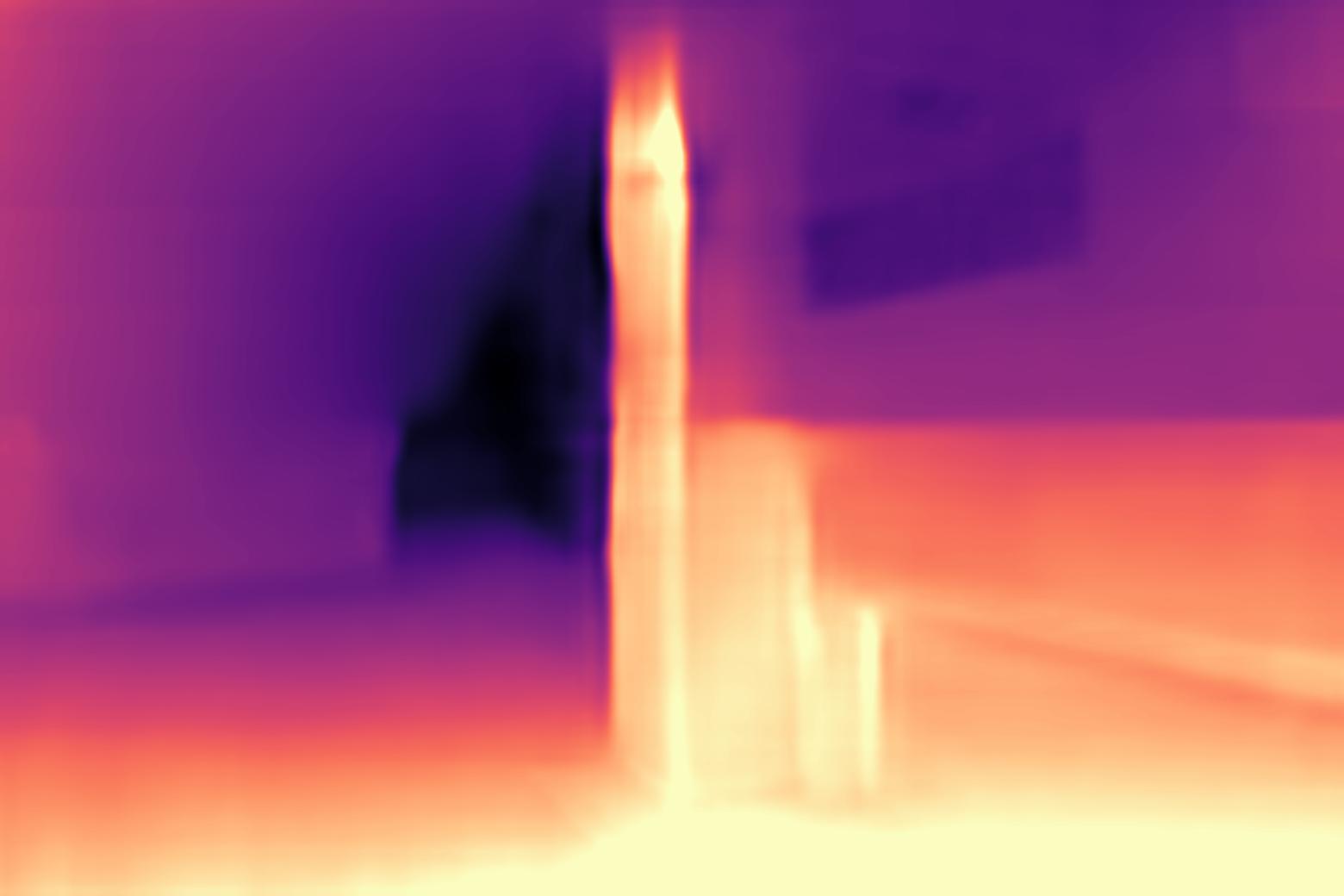}
    \includegraphics[width=\ssizew]{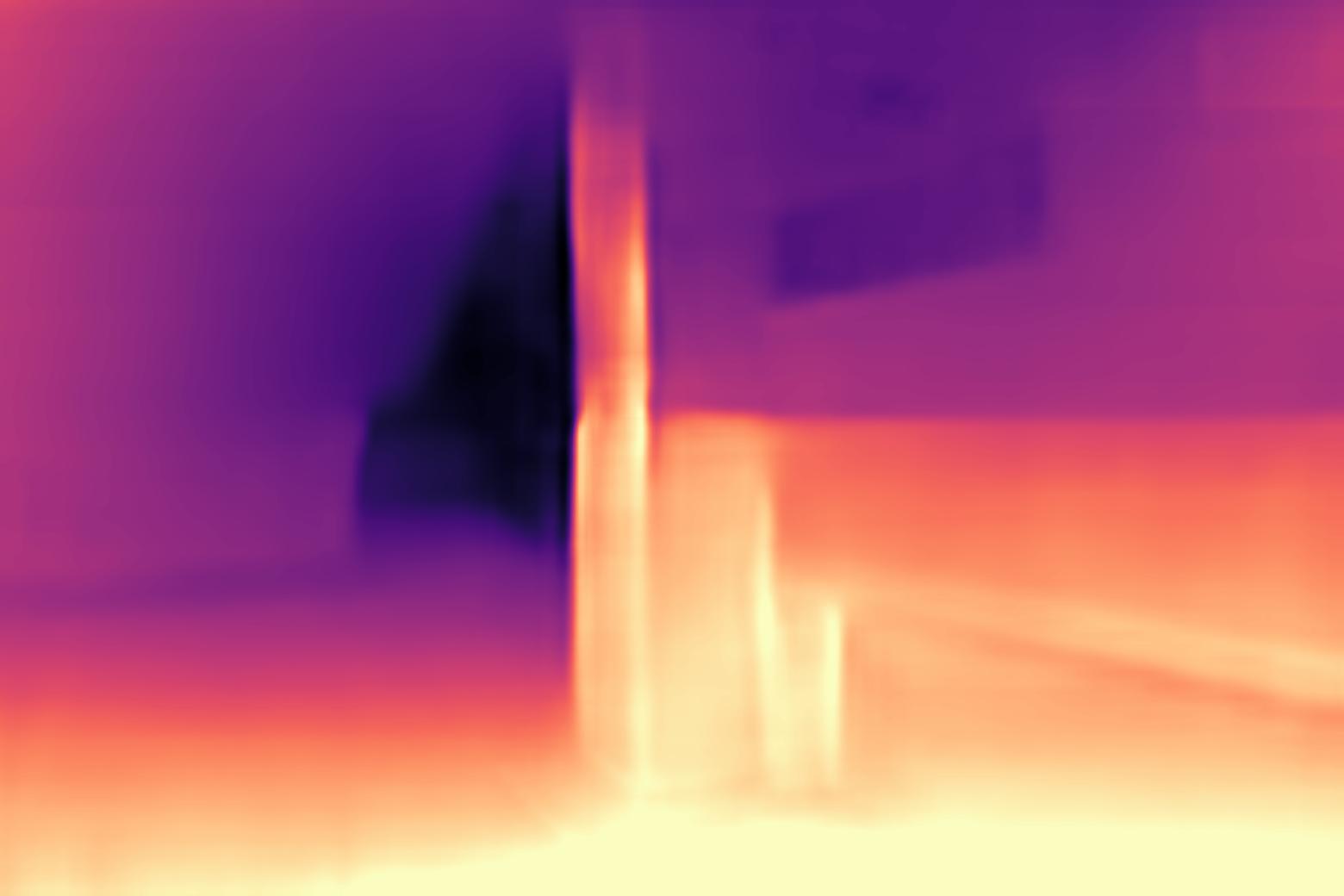}
    \includegraphics[width=\ssizew]{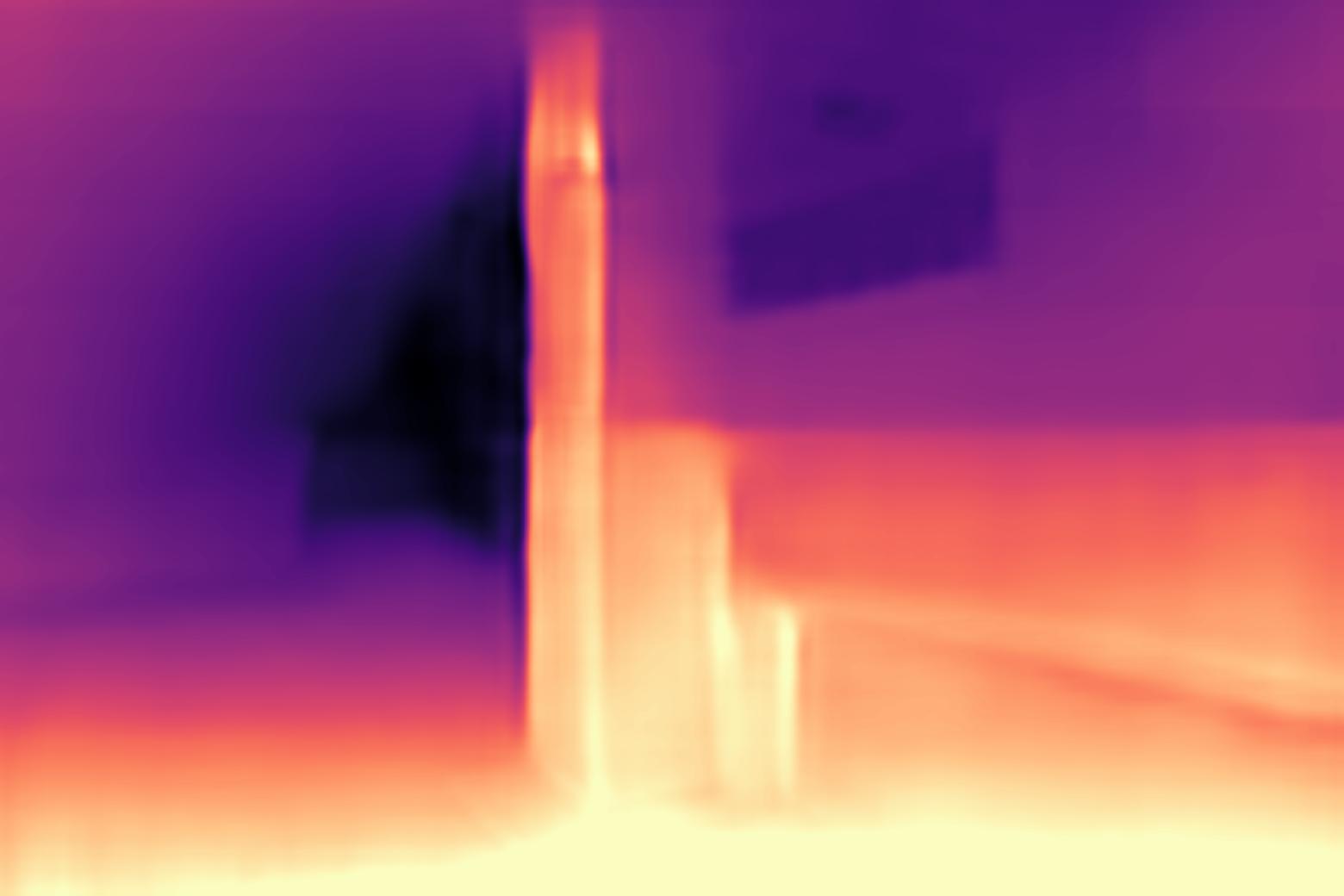}
    
    \begin{adjustbox}{angle=90}
    \begin{subfigure}[b]{\ssizeh}
        \vspace{-10pt}
        \caption{\textcolor{red}{Ours}}
    \end{subfigure}
    \end{adjustbox}
    \includegraphics[width=\ssizew]{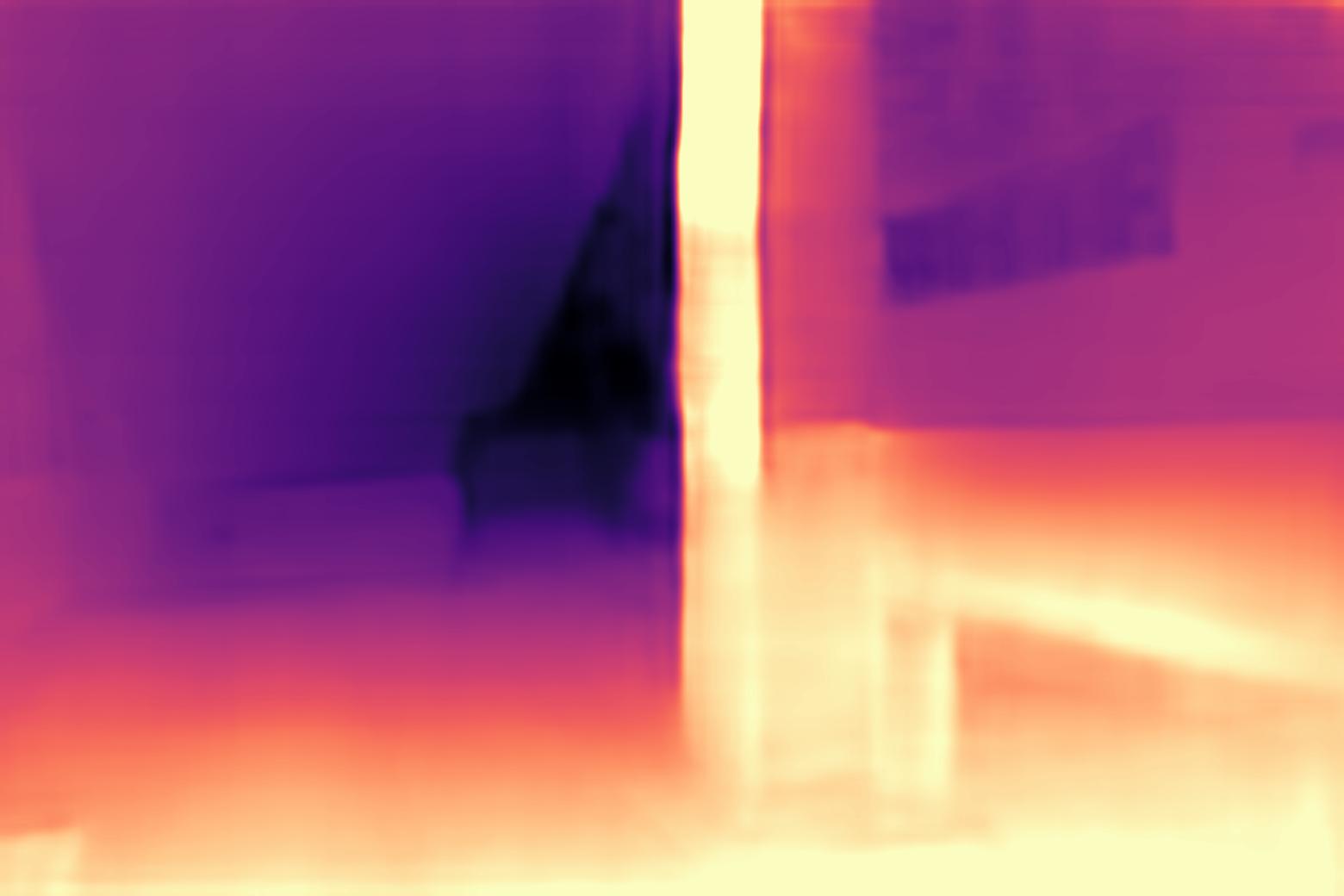}
    \includegraphics[width=\ssizew]{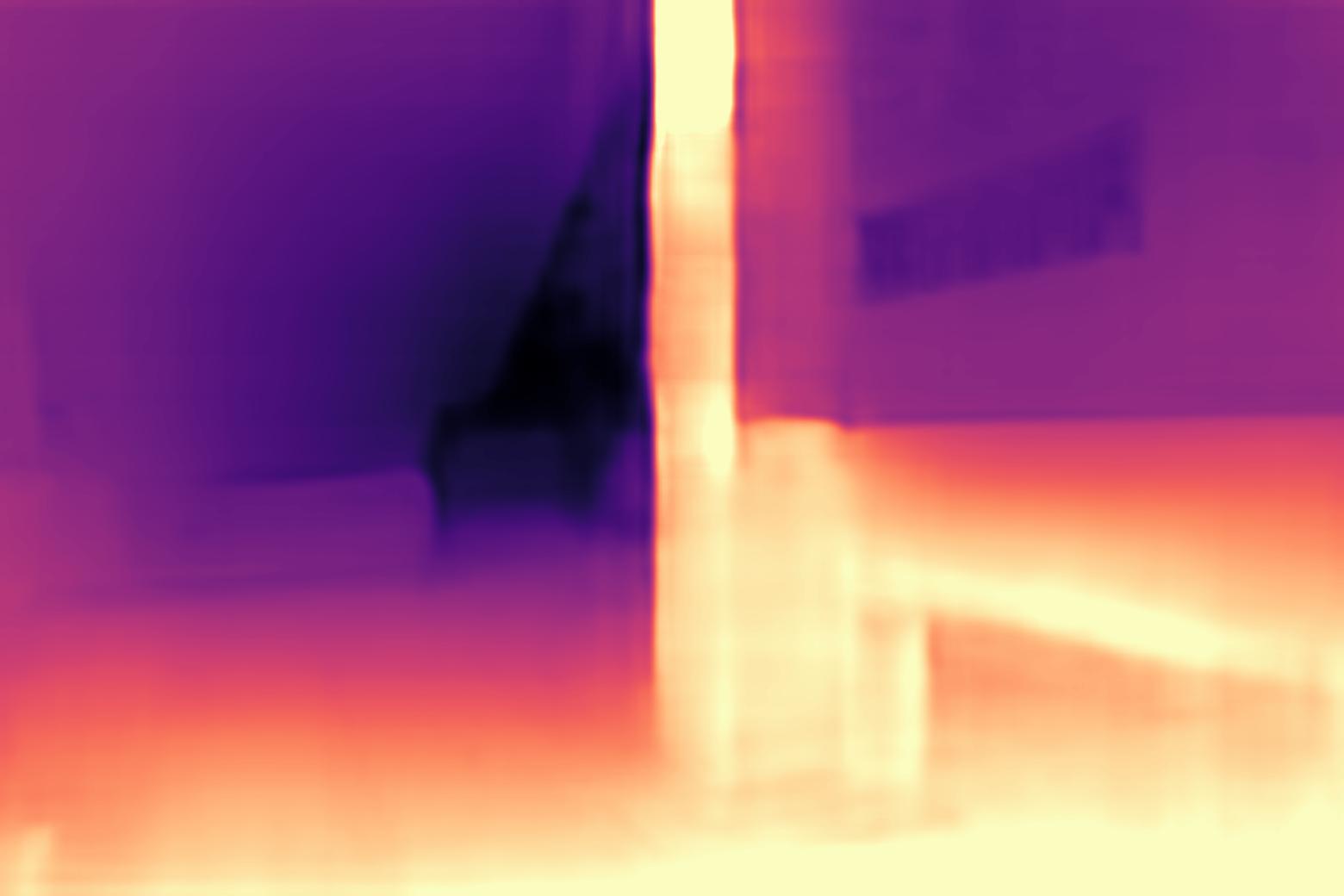}
    \includegraphics[width=\ssizew]{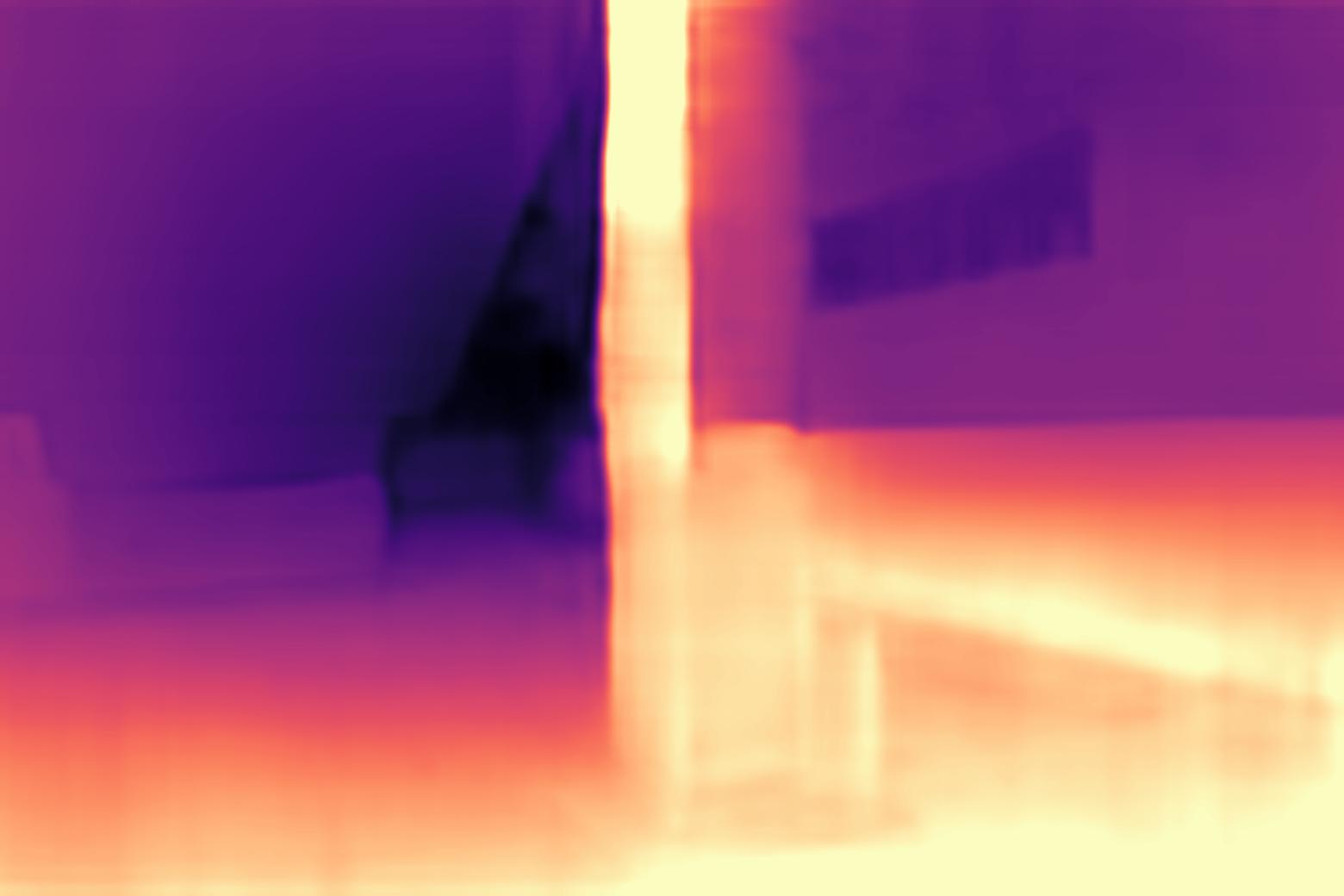}
    \includegraphics[width=\ssizew]{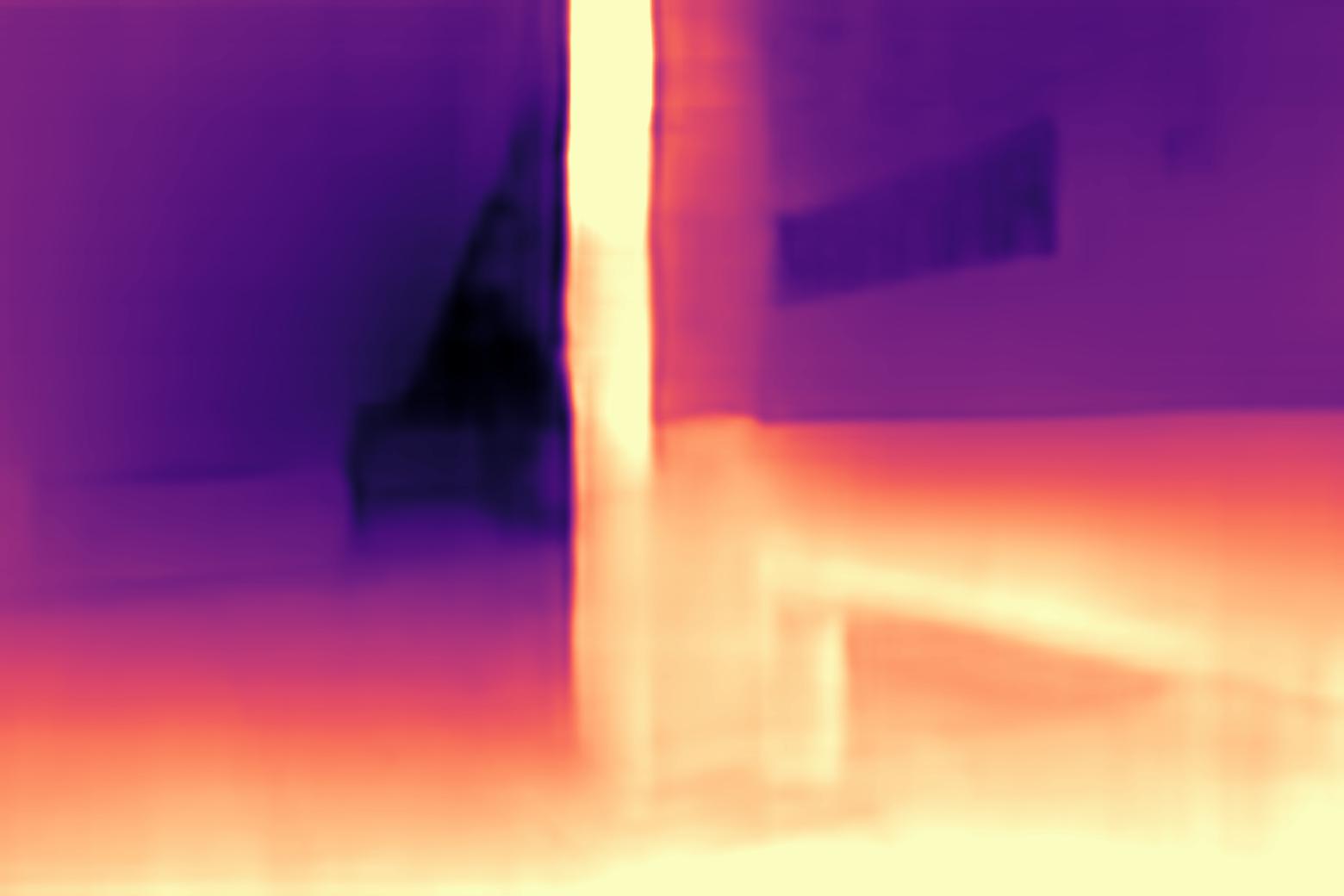}
    \includegraphics[width=\ssizew]{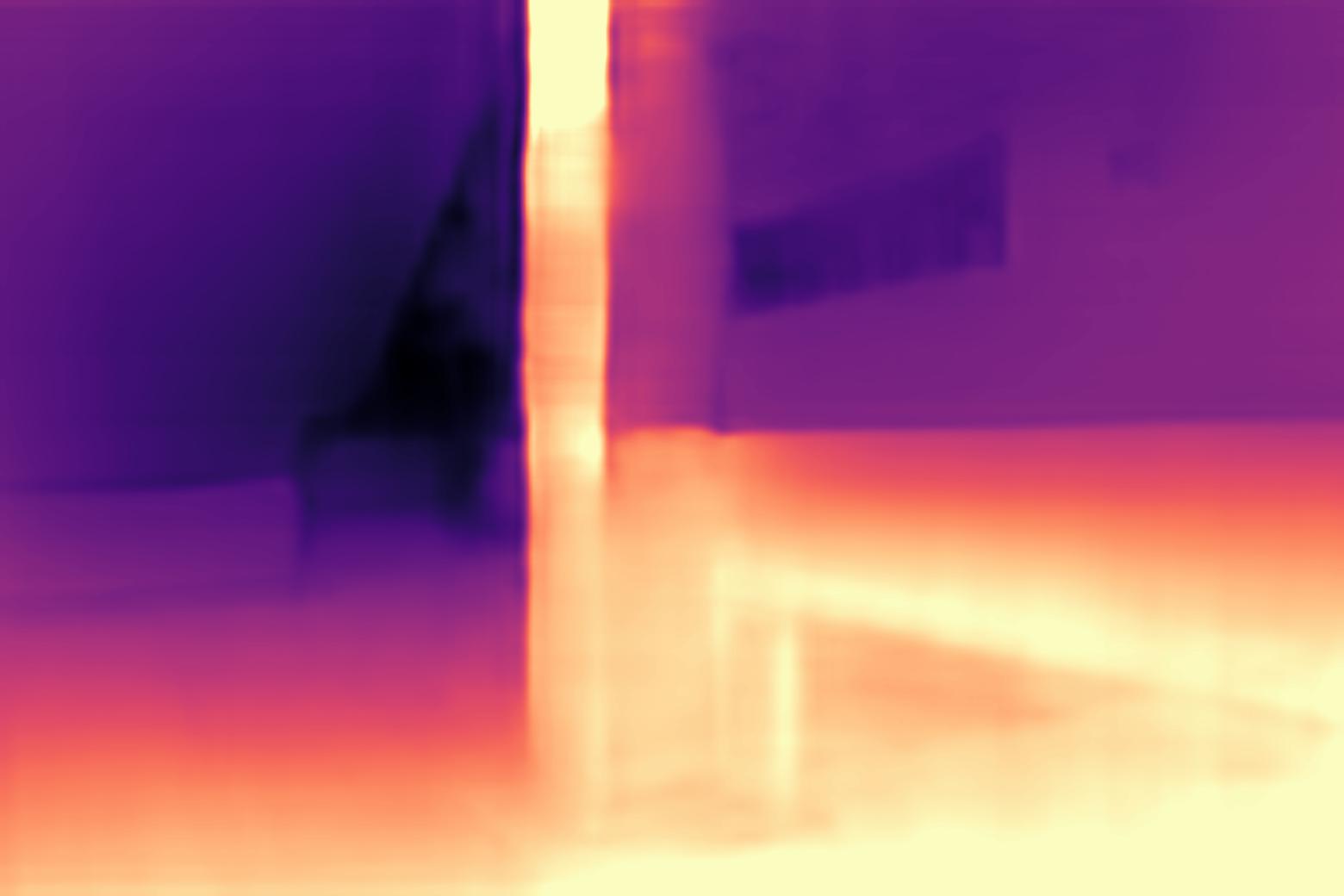}

    \caption{\textcolor{red}{Comparison of generalization results between $\mathrm{P^2Net}$ and our method in scene No.7. Top to bottom: RGB images, $\mathrm{P^2Net}$\citep{Yu2020} and ours.}}
    \label{fig:fig_7scenes_cmp}
\end{figure}
\subsection{Ablation studies}

We conduct ablation studies on NYU Depth V2 dataset to verify the effectiveness of optical flow consistency loss, multi-scale feature map synthesis loss and patch-based photometric loss for the optical flow estimation network.

\subsubsection{ Effects of optical flow consistency loss and multi-scale feature map synthesis loss}

In \autoref{tab:tab5}, we report the evaluation result without any proposed losses in the first row, then only add optical flow consistency loss for training and report the result in the second row. The third row shows our complete model’s result with the addition of multi-scale feature map synthesis loss. \textcolor{red}{The first three rows show the results without post-processing while the last three rows with post-processing.} The experiments can validate the effectiveness of our proposed losses.

\begin{table}[H]

\footnotesize
\centering
\begin{adjustbox}{center}
\setlength\tabcolsep{3pt}
\begin{tabular}[c]{c  >{\centering\arraybackslash}m{2.2cm} >{\centering\arraybackslash}m{2.2cm} l l l l l l}
\hline
\makecell[b]{\textcolor{red}{PP}} & \makecell[bl]{Optical flow\\ consistency\\ loss} & \makecell[bl]{Feature map\\synthesis loss} & \makecell[b]{REL$\downarrow$} & \makecell[b]{RMS$\downarrow$} & \makecell[b]{log10$\downarrow$} & \makecell[b]{$\delta\!<\!1.25$$\uparrow$} & \makecell[b]{$\delta\!<\!1.25^2$$\uparrow$} &\makecell[b]{$\delta\!<\!1.25^3$$\uparrow$}  \\
\hline
\textcolor{red}{$\times$}& $\times$ & $\times$ & 0.159 & 0.599 & 0.068 & 0.772 & 0.942 & 0.984 \\
\textcolor{red}{$\times$}& \checkmark & $\times$ & 0.158 & 0.586 & 0.067 & 0.778 & 0.946 & 0.986 \\
\textcolor{red}{$\times$}& \checkmark & \checkmark & \textbf{0.158} & \textbf{0.583} & \textbf{0.067} & \textbf{0.779} & \textbf{0.947} & \textbf{0.987} \\
\hline
\textcolor{red}{\checkmark} &\textcolor{red}{$\times$} &\textcolor{red}{$\times$} & \textcolor{red}{0.157} & \textcolor{red}{0.592} & \textcolor{red}{0.067} & \textcolor{red}{0.777} & \textcolor{red}{0.944} & \textcolor{red}{0.985} \\
\textcolor{red}{\checkmark} & \textcolor{red}{\checkmark} & \textcolor{red}{$\times$} & \textcolor{red}{0.156} & \textcolor{red}{0.577} & \textcolor{red}{0.066} & \textcolor{red}{0.783} & \textcolor{red}{0.947} & \textcolor{red}{0.986} \\
\textcolor{red}{\checkmark} &\textcolor{red}{\checkmark} & \textcolor{red}{\checkmark} & \textcolor{red}{\textbf{0.153}} & \textcolor{red}{\textbf{0.569}} & \textcolor{red}{\textbf{0.065}} & \textcolor{red}{\textbf{0.787}} & \textcolor{red}{\textbf{0.950}} & \textcolor{red}{\textbf{0.987}} \\
\hline
\end{tabular}
\end{adjustbox}
\caption{Ablation studies results of our losses on the NYU Depth V2 dataset. \textcolor{red}{PP denotes the results with post-processing.} $\downarrow$ means the lower the better, $\uparrow$  means the higher the better.}
\label{tab:tab5}
\vspace{-13mm}
\end{table}

\subsubsection{Effects of patch-based photometric loss for the optical flow estimation network}

We train our depth estimation model respectively using original optical flow estimation network pretrained on MPI Sintel dataset, finetuned original network and finetuned network with our loss. We report evaluation results in \autoref{tab:tab6}. Only finetuning original optical flow estimation network on NYU Depth V2 is not even as good as no finetuning. It can be demonstrated that the patch-based photometric loss of optical flow estimation network has positive effects on monocular depth estimation. 

\begin{center}
\begin{table}[H]


\footnotesize
\centering
\setlength\tabcolsep{4pt}
\begin{tabular}[c]{l l l l l l l}
\hline
Weights & REL$\downarrow$ & RMS$\downarrow$ & \makecell{log10$\downarrow$} & \makecell{$\delta\!<\!1.25$$\uparrow$} & \makecell{$\delta\!<\!1.25^2$$\uparrow$} &\makecell{$\delta\!<\!1.25^3$$\uparrow$}  \\
\hline
Pretrained original network & 0.159 & 0.591 & 0.068 & 0.775 & 0.946 & 0.986 \\
Finetuned original network & 0.160 & 0.596 & 0.068 & 0.773 & 0.945 & 0.986 \\
Our finetuned network & \textbf{0.158} & \textbf{0.583} & \textbf{0.067} & \textbf{0.779} & \textbf{0.947} & \textbf{0.987} \\
\hline
\end{tabular}
\caption{Monocular depth estimation results with different optical flow estimation network weights. $\downarrow$ means the lower the better, $\uparrow$  means the higher the better.}
\label{tab:tab6}
\vspace{-13mm}
\end{table}

\end{center}

\begin{figure}[H]
    \centering
    \includegraphics[width=0.242\textwidth]{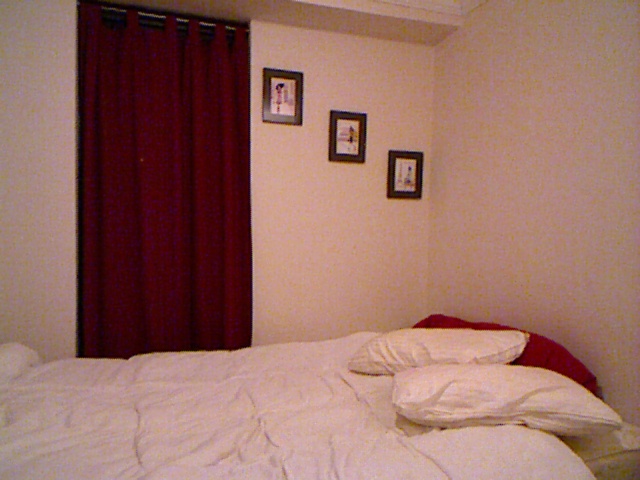}
    \includegraphics[width=0.242\textwidth]{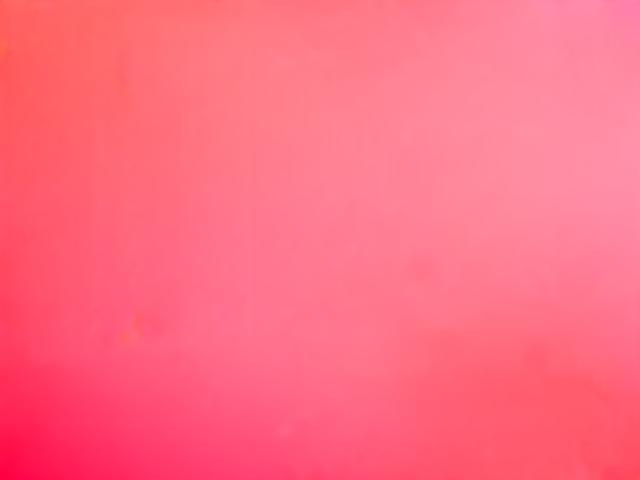}
    \includegraphics[width=0.242\textwidth]{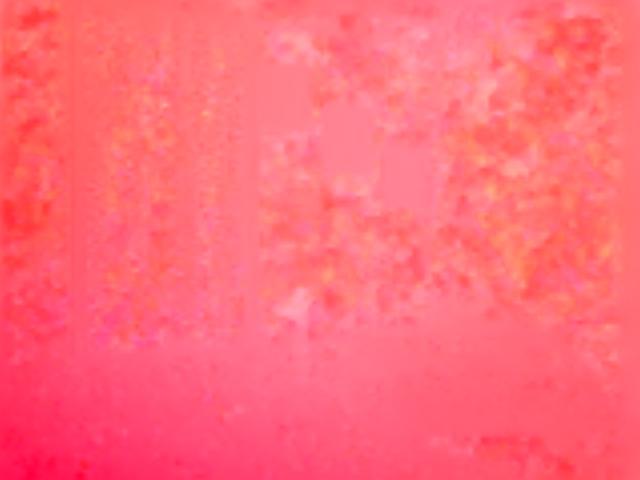}
    \includegraphics[width=0.242\textwidth]{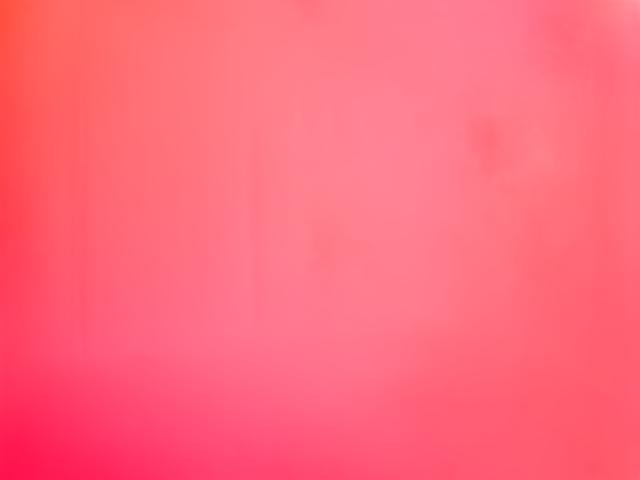}
    \includegraphics[width=0.242\textwidth]{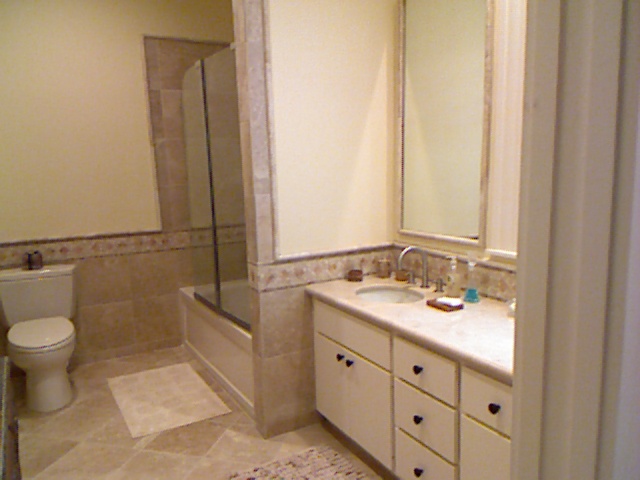}
    \includegraphics[width=0.242\textwidth]{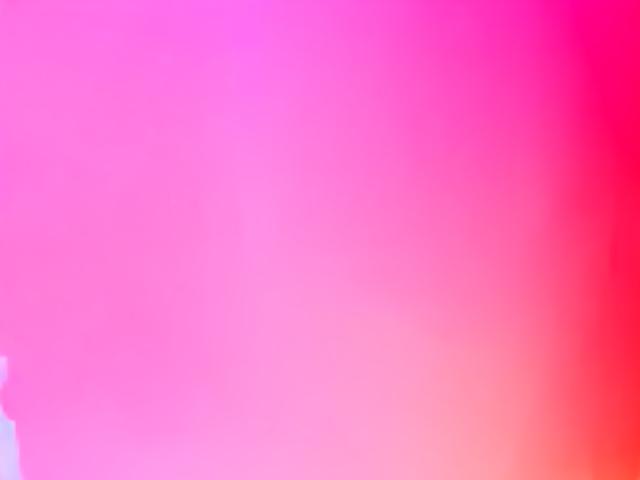}
    \includegraphics[width=0.242\textwidth]{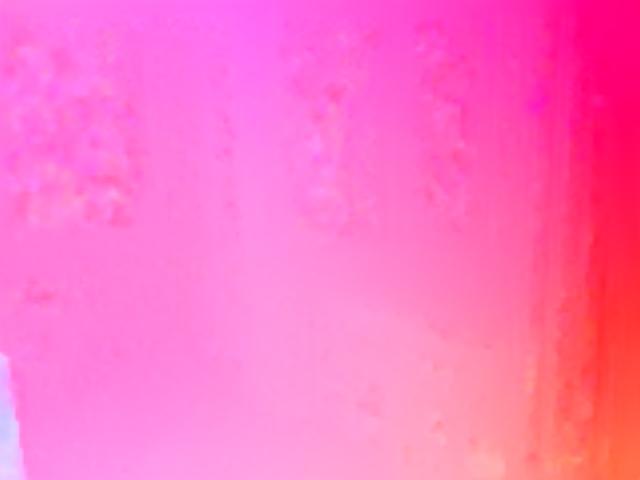}
    \includegraphics[width=0.242\textwidth]{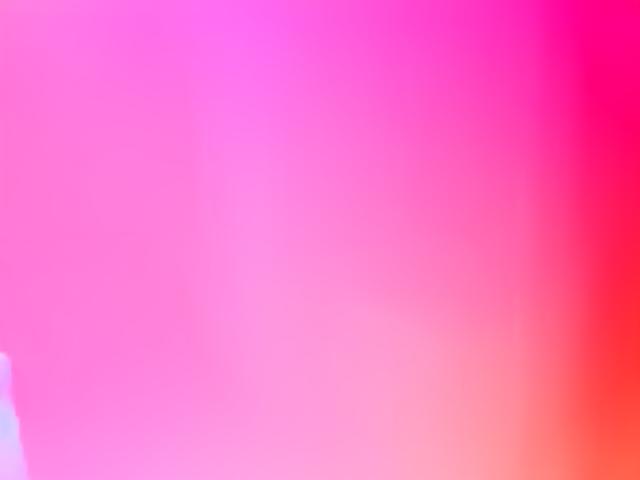}
    \includegraphics[width=0.242\textwidth]{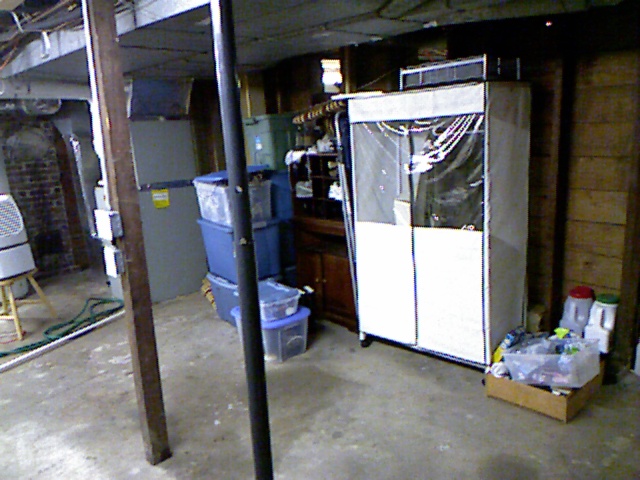}
    \includegraphics[width=0.242\textwidth]{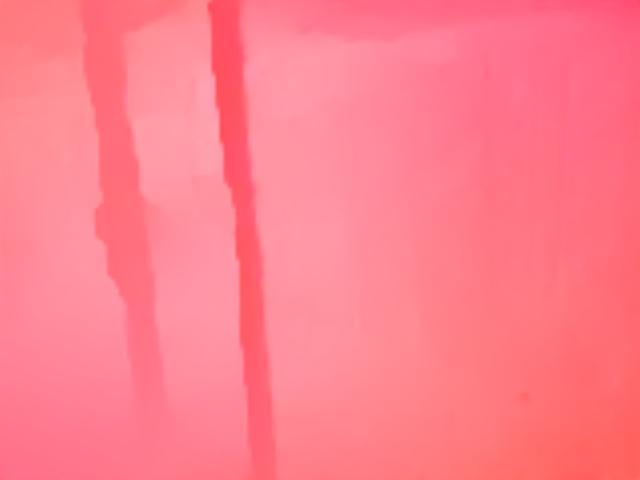}
    \includegraphics[width=0.242\textwidth]{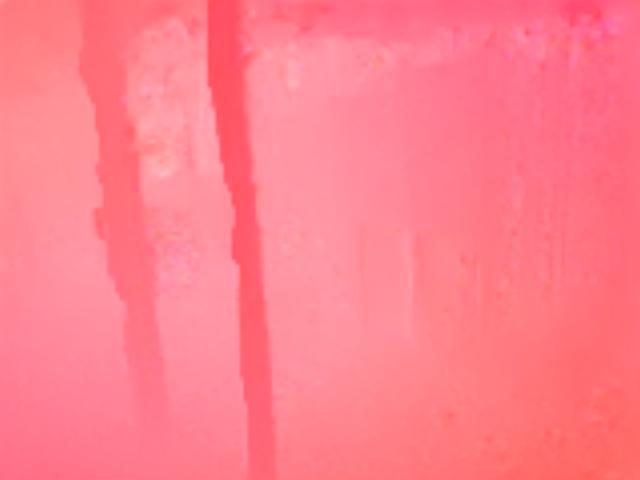}
    \includegraphics[width=0.242\textwidth]{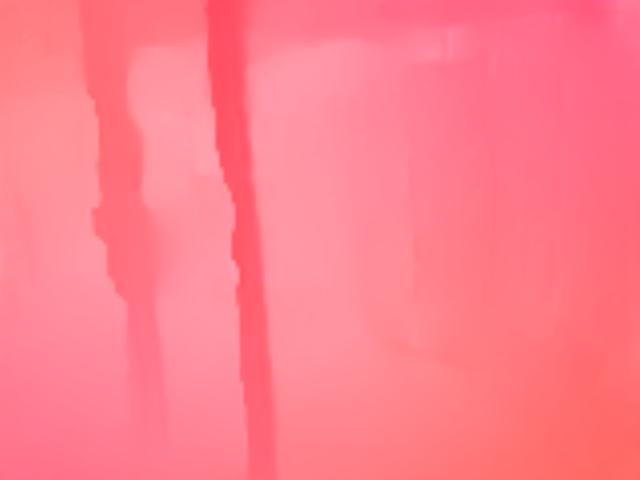}
    \includegraphics[width=0.242\textwidth]{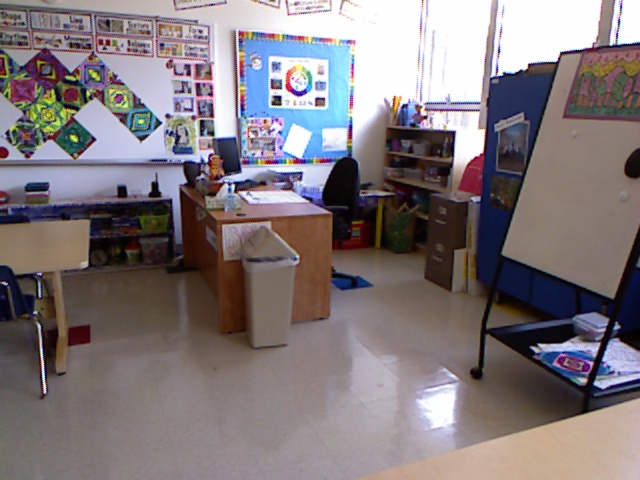}
    \includegraphics[width=0.242\textwidth]{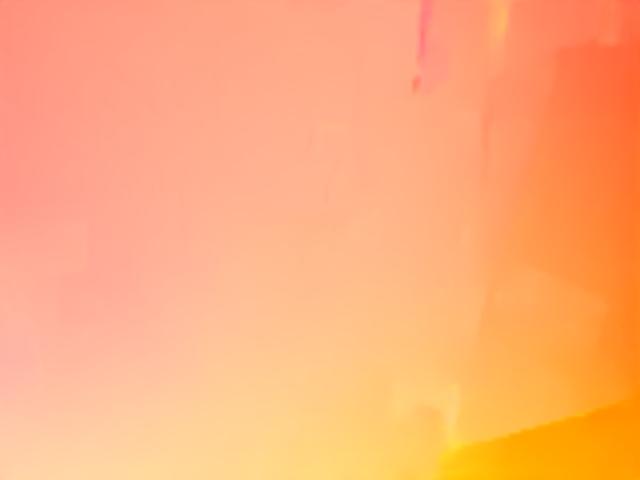}
    \includegraphics[width=0.242\textwidth]{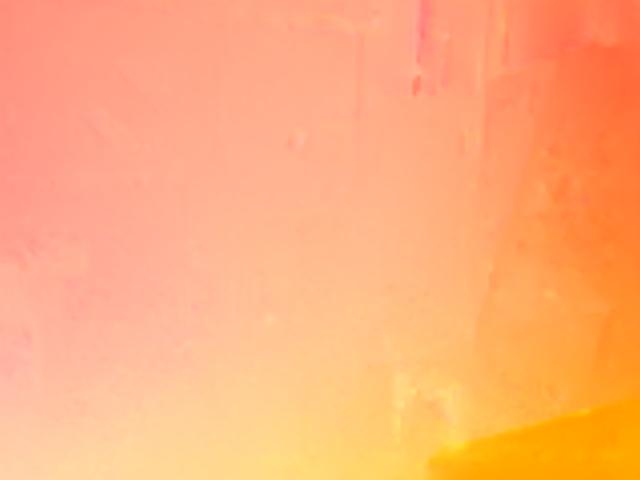}
    \includegraphics[width=0.242\textwidth]{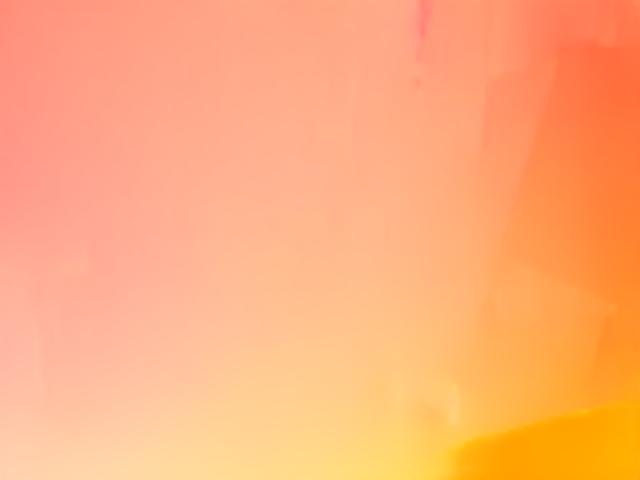}\\
    \vspace{-0.25cm}
    \begin{subfigure}[b]{0.242\textwidth}
        \caption{\textcolor{red}{RGB Image}}
    \end{subfigure}
    \begin{subfigure}[b]{0.242\textwidth}
        \caption{\textcolor{red}{Pretrained Original}}
    \end{subfigure}
    \begin{subfigure}[b]{0.242\textwidth}
        \caption{\textcolor{red}{Finetuned Original}}
    \end{subfigure}
    \begin{subfigure}[b]{0.242\textwidth}
        \caption{\textcolor{red}{Finetuned Ours}}
    \end{subfigure}
    \caption{Optical flow finetuning results. Left to right: RGB image, pretrained original optical flow estimation network on Sintel, finetuned original network, our finetuned network.}
    \label{fig:fig_8}
\end{figure}

For optical flow estimation results, since there is no ground truth of optical flow in NYU Depth V2 dataset, we cannot evaluate quantitatively. Therefore, we present qualitative optical flow results to illustrate the effect of improving the photometric loss qualitatively. \autoref{fig:fig_8} shows optical flow finetuning results with pretrained network on MPI Sintel, finetuned network before and after adjusting the loss function. For some low-textured areas such as walls, doors and the whiteboard in scenes, the original optical flow estimation network computes optical flow not smoothly and correctly, i.e., the regions looking a bit chaotic in maps. The improved network is finetuned better in these low-textured regions.

\section{Conclusions}

In this work, we propose to improve the performance of indoor monocular depth estimation by introducing optical flow learning. To handle the low-texture issue, original photometric loss of the optical flow estimation network is optimized to patch-based photometric loss. This helps reduce the participation of pixels in low-textured regions in training and obtain better optical flow prediction results. The finetuned flow estimation network produces high-accuracy optical flow to supervise rigid flow. Feature maps with fewer low gradient variance regions are generated from finetuned flow network. The multi-scale feature maps are warped with rigid flow to compute feature map synthesis loss. All these contributions are demonstrated effective in experiments on NYU Depth V2 dataset. We also conduct zero-shot generalization experiments on the \textcolor{red}{7-Scenes and} Campus Indoor dataset. The experiments \textcolor{red}{on both datasets} achieve good generalization results. \textcolor{red}{In the future, we will apply the proposed losses in more advanced approaches to improve the robustness of our work. Simultaneous training of optical flow and depth will be considered to explore the effect on depth estimation. For feature enhancement, Transformer architecture can be adopted. Other kinds of losses beneficial for indoor depth estimation will also be designed.}

\section*{Acknowledgments}
This work was supported by the Application Innovation Project of CASC [grant number 6230109004].

\bibliographystyle{elsarticle-harv} 
\bibliography{bib}

\begin{thebibliography}{71}
\expandafter\ifx\csname natexlab\endcsname\relax\def\natexlab#1{#1}\fi
\providecommand{\url}[1]{\texttt{#1}}
\providecommand{\href}[2]{#2}
\providecommand{\path}[1]{#1}
\providecommand{\DOIprefix}{doi:}
\providecommand{\ArXivprefix}{arXiv:}
\providecommand{\URLprefix}{URL: }
\providecommand{\Pubmedprefix}{pmid:}
\providecommand{\doi}[1]{\href{http://dx.doi.org/#1}{\path{#1}}}
\providecommand{\Pubmed}[1]{\href{pmid:#1}{\path{#1}}}
\providecommand{\bibinfo}[2]{#2}
\ifx\xfnm\relax \def\xfnm[#1]{\unskip,\space#1}\fi
\bibitem[{Agarwal and Arora(2022)}]{Agarwal2022}
\bibinfo{author}{Agarwal, A.}, \bibinfo{author}{Arora, C.}, \bibinfo{year}{2022}.
\newblock \bibinfo{title}{Depthformer: Multiscale vision transformer for monocular depth estimation with global local information fusion}, in: \bibinfo{booktitle}{2022 IEEE International Conference on Image Processing (ICIP)}, \bibinfo{publisher}{IEEE}. pp. \bibinfo{pages}{3873--3877}.
\bibitem[{Agarwal and Arora(2023)}]{Agarwal2023}
\bibinfo{author}{Agarwal, A.}, \bibinfo{author}{Arora, C.}, \bibinfo{year}{2023}.
\newblock \bibinfo{title}{Attention attention everywhere: Monocular depth prediction with skip attention}, in: \bibinfo{booktitle}{Proceedings of the IEEE/CVF Winter Conference on Applications of Computer Vision}, pp. \bibinfo{pages}{5861--5870}.
\bibitem[{Bhat et~al.(2021)Bhat, Alhashim and Wonka}]{Bhat2021}
\bibinfo{author}{Bhat, S.F.}, \bibinfo{author}{Alhashim, I.}, \bibinfo{author}{Wonka, P.}, \bibinfo{year}{2021}.
\newblock \bibinfo{title}{Adabins: Depth estimation using adaptive bins}, in: \bibinfo{booktitle}{Proceedings of the IEEE/CVF Conference on Computer Vision and Pattern Recognition}, pp. \bibinfo{pages}{4009--4018}.
\bibitem[{Bhat et~al.(2022)Bhat, Alhashim and Wonka}]{Bhat2022}
\bibinfo{author}{Bhat, S.F.}, \bibinfo{author}{Alhashim, I.}, \bibinfo{author}{Wonka, P.}, \bibinfo{year}{2022}.
\newblock \bibinfo{title}{Localbins: Improving depth estimation by learning local distributions}, in: \bibinfo{booktitle}{Computer Vision–ECCV 2022: 17th European Conference, Tel Aviv, Israel, October 23–27, 2022, Proceedings, Part I}, \bibinfo{publisher}{Springer}. pp. \bibinfo{pages}{480--496}.
\bibitem[{Bian et~al.(2021a)Bian, Zhan, Wang, Chin, Shen and Reid}]{Bian2021_RN54}
\bibinfo{author}{Bian, J.W.}, \bibinfo{author}{Zhan, H.}, \bibinfo{author}{Wang, N.}, \bibinfo{author}{Chin, T.J.}, \bibinfo{author}{Shen, C.}, \bibinfo{author}{Reid, I.}, \bibinfo{year}{2021}a.
\newblock \bibinfo{title}{Auto-rectify network for unsupervised indoor depth estimation}.
\newblock \bibinfo{journal}{IEEE Transactions on Pattern Analysis and Intelligence, Machine} \bibinfo{volume}{44}, \bibinfo{pages}{9802--9813}.
\bibitem[{Bian et~al.(2021b)Bian, Zhan, Wang, Li, Zhang, Shen, Cheng and Reid}]{Bian2021_RN52}
\bibinfo{author}{Bian, J.W.}, \bibinfo{author}{Zhan, H.}, \bibinfo{author}{Wang, N.}, \bibinfo{author}{Li, Z.}, \bibinfo{author}{Zhang, L.}, \bibinfo{author}{Shen, C.}, \bibinfo{author}{Cheng, M.M.}, \bibinfo{author}{Reid, I.}, \bibinfo{year}{2021}b.
\newblock \bibinfo{title}{Unsupervised scale-consistent depth learning from video}.
\newblock \bibinfo{journal}{International Journal of Computer Vision} \bibinfo{volume}{129}, \bibinfo{pages}{2548--2564}.
\bibitem[{Butler et~al.(2012)Butler, Wulff, Stanley and Black}]{Butler2012}
\bibinfo{author}{Butler, D.J.}, \bibinfo{author}{Wulff, J.}, \bibinfo{author}{Stanley, G.B.}, \bibinfo{author}{Black, M.J.}, \bibinfo{year}{2012}.
\newblock \bibinfo{title}{A naturalistic open source movie for optical flow evaluation}, in: \bibinfo{booktitle}{Computer Vision–ECCV 2012: 12th European Conference on Computer Vision, Florence, Italy, October 7-13, 2012, Proceedings, Part VI 12}, \bibinfo{publisher}{Springer}. pp. \bibinfo{pages}{611--625}.
\bibitem[{Chopra et~al.(2005)Chopra, Hadsell and LeCun}]{Chopra2005}
\bibinfo{author}{Chopra, S.}, \bibinfo{author}{Hadsell, R.}, \bibinfo{author}{LeCun, Y.}, \bibinfo{year}{2005}.
\newblock \bibinfo{title}{Learning a similarity metric discriminatively, with application to face verification}, in: \bibinfo{booktitle}{2005 IEEE Computer Society Conference on Computer Vision and Pattern Recognition (CVPR'05)}, \bibinfo{publisher}{IEEE}. pp. \bibinfo{pages}{539--546}.
\bibitem[{Cordts et~al.(2016)Cordts, Omran, Ramos, Rehfeld, Enzweiler, Benenson, Franke, Roth and Schiele}]{Cordts2016}
\bibinfo{author}{Cordts, M.}, \bibinfo{author}{Omran, M.}, \bibinfo{author}{Ramos, S.}, \bibinfo{author}{Rehfeld, T.}, \bibinfo{author}{Enzweiler, M.}, \bibinfo{author}{Benenson, R.}, \bibinfo{author}{Franke, U.}, \bibinfo{author}{Roth, S.}, \bibinfo{author}{Schiele, B.}, \bibinfo{year}{2016}.
\newblock \bibinfo{title}{The cityscapes dataset for semantic urban scene understanding}, in: \bibinfo{booktitle}{Proceedings of the IEEE conference on computer vision and pattern recognition}, pp. \bibinfo{pages}{3213--3223}.
\bibitem[{Dosovitskiy et~al.(2015)Dosovitskiy, Fischer, Ilg, Hausser, Hazirbas, Golkov, Van Der~Smagt, Cremers and Brox}]{Dosovitskiy2015}
\bibinfo{author}{Dosovitskiy, A.}, \bibinfo{author}{Fischer, P.}, \bibinfo{author}{Ilg, E.}, \bibinfo{author}{Hausser, P.}, \bibinfo{author}{Hazirbas, C.}, \bibinfo{author}{Golkov, V.}, \bibinfo{author}{Van Der~Smagt, P.}, \bibinfo{author}{Cremers, D.}, \bibinfo{author}{Brox, T.}, \bibinfo{year}{2015}.
\newblock \bibinfo{title}{Flownet: Learning optical flow with convolutional networks}, in: \bibinfo{booktitle}{Proceedings of the IEEE international conference on computer vision}, pp. \bibinfo{pages}{2758--2766}.
\bibitem[{Eigen and Fergus(2015)}]{Eigen2015}
\bibinfo{author}{Eigen, D.}, \bibinfo{author}{Fergus, R.}, \bibinfo{year}{2015}.
\newblock \bibinfo{title}{Predicting depth, surface normals and semantic labels with a common multi-scale convolutional architecture}, in: \bibinfo{booktitle}{Proceedings of the IEEE international conference on computer vision}, pp. \bibinfo{pages}{2650--2658}.
\bibitem[{Eigen et~al.(2014)Eigen, Puhrsch and Fergus}]{Eigen2014}
\bibinfo{author}{Eigen, D.}, \bibinfo{author}{Puhrsch, C.}, \bibinfo{author}{Fergus, R.}, \bibinfo{year}{2014}.
\newblock \bibinfo{title}{Depth map prediction from a single image using a multi-scale deep network}.
\newblock \bibinfo{journal}{Advances in neural information processing systems} \bibinfo{volume}{27}.
\bibitem[{Engel et~al.(2017)Engel, Koltun and Cremers}]{Engel2017}
\bibinfo{author}{Engel, J.}, \bibinfo{author}{Koltun, V.}, \bibinfo{author}{Cremers, D.}, \bibinfo{year}{2017}.
\newblock \bibinfo{title}{Direct sparse odometry}.
\newblock \bibinfo{journal}{IEEE transactions on pattern analysis and intelligence, machine} \bibinfo{volume}{40}, \bibinfo{pages}{611--625}.
\bibitem[{Fu et~al.(2018)Fu, Gong, Wang, Batmanghelich and Tao}]{Fu2018}
\bibinfo{author}{Fu, H.}, \bibinfo{author}{Gong, M.}, \bibinfo{author}{Wang, C.}, \bibinfo{author}{Batmanghelich, K.}, \bibinfo{author}{Tao, D.}, \bibinfo{year}{2018}.
\newblock \bibinfo{title}{Deep ordinal regression network for monocular depth estimation}, in: \bibinfo{booktitle}{Proceedings of the IEEE conference on computer vision and pattern recognition}, pp. \bibinfo{pages}{2002--2011}.
\bibitem[{Garg et~al.(2016)Garg, Bg, Carneiro and Reid}]{Garg2016}
\bibinfo{author}{Garg, R.}, \bibinfo{author}{Bg, V.K.}, \bibinfo{author}{Carneiro, G.}, \bibinfo{author}{Reid, I.}, \bibinfo{year}{2016}.
\newblock \bibinfo{title}{Unsupervised cnn for single view depth estimation: Geometry to the rescue}, in: \bibinfo{booktitle}{Computer Vision–ECCV 2016: 14th European Conference, Amsterdam, The Netherlands, October 11-14, 2016, Proceedings, Part VIII 14}, \bibinfo{publisher}{Springer}. pp. \bibinfo{pages}{740--756}.
\bibitem[{Godard et~al.(2017)Godard, Mac~Aodha and Brostow}]{Godard2017}
\bibinfo{author}{Godard, C.}, \bibinfo{author}{Mac~Aodha, O.}, \bibinfo{author}{Brostow, G.J.}, \bibinfo{year}{2017}.
\newblock \bibinfo{title}{Unsupervised monocular depth estimation with left-right consistency}, in: \bibinfo{booktitle}{Proceedings of the IEEE conference on computer vision and pattern recognition}, pp. \bibinfo{pages}{270--279}.
\bibitem[{Godard et~al.(2019)Godard, Mac~Aodha, Firman and Brostow}]{Godard2019}
\bibinfo{author}{Godard, C.}, \bibinfo{author}{Mac~Aodha, O.}, \bibinfo{author}{Firman, M.}, \bibinfo{author}{Brostow, G.J.}, \bibinfo{year}{2019}.
\newblock \bibinfo{title}{Digging into self-supervised monocular depth estimation}, in: \bibinfo{booktitle}{Proceedings of the IEEE/CVF international conference on computer vision}, pp. \bibinfo{pages}{3828--3838}.
\bibitem[{He et~al.(2016)He, Zhang, Ren and Sun}]{He2016}
\bibinfo{author}{He, K.}, \bibinfo{author}{Zhang, X.}, \bibinfo{author}{Ren, S.}, \bibinfo{author}{Sun, J.}, \bibinfo{year}{2016}.
\newblock \bibinfo{title}{Deep residual learning for image recognition}, in: \bibinfo{booktitle}{Proceedings of the IEEE conference on computer vision and pattern recognition}, pp. \bibinfo{pages}{770--778}.
\bibitem[{Hu et~al.(2019)Hu, Ozay, Zhang and Okatani}]{Hu2019}
\bibinfo{author}{Hu, J.}, \bibinfo{author}{Ozay, M.}, \bibinfo{author}{Zhang, Y.}, \bibinfo{author}{Okatani, T.}, \bibinfo{year}{2019}.
\newblock \bibinfo{title}{Revisiting single image depth estimation: Toward higher resolution maps with accurate object boundaries}, in: \bibinfo{booktitle}{2019 IEEE winter conference on applications of computer vision (WACV)}, \bibinfo{publisher}{IEEE}. pp. \bibinfo{pages}{1043--1051}.
\bibitem[{Ilg et~al.(2017)Ilg, Mayer, Saikia, Keuper, Dosovitskiy and Brox}]{Ilg2017}
\bibinfo{author}{Ilg, E.}, \bibinfo{author}{Mayer, N.}, \bibinfo{author}{Saikia, T.}, \bibinfo{author}{Keuper, M.}, \bibinfo{author}{Dosovitskiy, A.}, \bibinfo{author}{Brox, T.}, \bibinfo{year}{2017}.
\newblock \bibinfo{title}{Flownet 2.0: Evolution of optical flow estimation with deep networks}, in: \bibinfo{booktitle}{Proceedings of the IEEE conference on computer vision and pattern recognition}, pp. \bibinfo{pages}{2462--2470}.
\bibitem[{Janai et~al.(2018)Janai, Guney, Ranjan, Black and Geiger}]{Janai2018}
\bibinfo{author}{Janai, J.}, \bibinfo{author}{Guney, F.}, \bibinfo{author}{Ranjan, A.}, \bibinfo{author}{Black, M.}, \bibinfo{author}{Geiger, A.}, \bibinfo{year}{2018}.
\newblock \bibinfo{title}{Unsupervised learning of multi-frame optical flow with occlusions}, in: \bibinfo{booktitle}{Proceedings of the European conference on computer vision (ECCV)}, pp. \bibinfo{pages}{690--706}.
\bibitem[{Ji et~al.(2021)Ji, Li, Bhanu and Xu}]{Ji2021}
\bibinfo{author}{Ji, P.}, \bibinfo{author}{Li, R.}, \bibinfo{author}{Bhanu, B.}, \bibinfo{author}{Xu, Y.}, \bibinfo{year}{2021}.
\newblock \bibinfo{title}{Monoindoor: Towards good practice of self-supervised monocular depth estimation for indoor environments}, in: \bibinfo{booktitle}{Proceedings of the IEEE/CVF International Conference on Computer Vision}, pp. \bibinfo{pages}{12787--12796}.
\bibitem[{Jun et~al.(2022)Jun, Lee, Lee and Kim}]{Jun2022}
\bibinfo{author}{Jun, J.}, \bibinfo{author}{Lee, J.H.}, \bibinfo{author}{Lee, C.}, \bibinfo{author}{Kim, C.S.}, \bibinfo{year}{2022}.
\newblock \bibinfo{title}{Depth map decomposition for monocular depth estimation}, in: \bibinfo{booktitle}{Computer Vision–ECCV 2022: 17th European Conference, Tel Aviv, Israel, October 23–27, 2022, Proceedings, Part II}, \bibinfo{publisher}{Springer}. pp. \bibinfo{pages}{18--34}.
\bibitem[{Kong and Yang(2022)}]{kong2022mdflow}
\bibinfo{author}{Kong, L.}, \bibinfo{author}{Yang, J.}, \bibinfo{year}{2022}.
\newblock \bibinfo{title}{Mdflow: Unsupervised optical flow learning by reliable mutual knowledge distillation}.
\newblock \bibinfo{journal}{IEEE Transactions on Circuits and Systems for Video Technology} \bibinfo{volume}{33}, \bibinfo{pages}{677--688}.
\bibitem[{Laina et~al.(2016)Laina, Rupprecht, Belagiannis, Tombari and Navab}]{Laina2016}
\bibinfo{author}{Laina, I.}, \bibinfo{author}{Rupprecht, C.}, \bibinfo{author}{Belagiannis, V.}, \bibinfo{author}{Tombari, F.}, \bibinfo{author}{Navab, N.}, \bibinfo{year}{2016}.
\newblock \bibinfo{title}{Deeper depth prediction with fully convolutional residual networks}, in: \bibinfo{booktitle}{2016 Fourth international conference on 3D vision (3DV)}, \bibinfo{publisher}{IEEE}. pp. \bibinfo{pages}{239--248}.
\bibitem[{Li et~al.(2021)Li, Huang, Liu, Zou and Yu}]{Li2021}
\bibinfo{author}{Li, B.}, \bibinfo{author}{Huang, Y.}, \bibinfo{author}{Liu, Z.}, \bibinfo{author}{Zou, D.}, \bibinfo{author}{Yu, W.}, \bibinfo{year}{2021}.
\newblock \bibinfo{title}{Structdepth: Leveraging the structural regularities for self-supervised indoor depth estimation}, in: \bibinfo{booktitle}{Proceedings of the IEEE/CVF International Conference on Computer Vision}, pp. \bibinfo{pages}{12663--12673}.
\bibitem[{Li et~al.(2015)Li, Shen, Dai, Van Den~Hengel and He}]{Li2015}
\bibinfo{author}{Li, B.}, \bibinfo{author}{Shen, C.}, \bibinfo{author}{Dai, Y.}, \bibinfo{author}{Van Den~Hengel, A.}, \bibinfo{author}{He, M.}, \bibinfo{year}{2015}.
\newblock \bibinfo{title}{Depth and surface normal estimation from monocular images using regression on deep features and hierarchical crfs}, in: \bibinfo{booktitle}{Proceedings of the IEEE conference on computer vision and pattern recognition}, pp. \bibinfo{pages}{1119--1127}.
\bibitem[{Li et~al.(2017)Li, Klein and Yao}]{Li2017}
\bibinfo{author}{Li, J.}, \bibinfo{author}{Klein, R.}, \bibinfo{author}{Yao, A.}, \bibinfo{year}{2017}.
\newblock \bibinfo{title}{A two-streamed network for estimating fine-scaled depth maps from single rgb images}, in: \bibinfo{booktitle}{Proceedings of the IEEE International Conference on Computer Vision}, pp. \bibinfo{pages}{3372--3380}.
\bibitem[{Li et~al.(2022)Li, Ji, Xu and Bhanu}]{Li2022}
\bibinfo{author}{Li, R.}, \bibinfo{author}{Ji, P.}, \bibinfo{author}{Xu, Y.}, \bibinfo{author}{Bhanu, B.}, \bibinfo{year}{2022}.
\newblock \bibinfo{title}{Monoindoor++: towards better practice of self-supervised monocular depth estimation for indoor environments}.
\newblock \bibinfo{journal}{IEEE Transactions on Circuits and Systems for Video Technology} \bibinfo{volume}{33}, \bibinfo{pages}{830--846}.
\bibitem[{Liu et~al.(2018)Liu, Yang, Ceylan, Yumer and Furukawa}]{Liu2018}
\bibinfo{author}{Liu, C.}, \bibinfo{author}{Yang, J.}, \bibinfo{author}{Ceylan, D.}, \bibinfo{author}{Yumer, E.}, \bibinfo{author}{Furukawa, Y.}, \bibinfo{year}{2018}.
\newblock \bibinfo{title}{Planenet: Piece-wise planar reconstruction from a single rgb image}, in: \bibinfo{booktitle}{Proceedings of the IEEE Conference on Computer Vision and Pattern Recognition}, pp. \bibinfo{pages}{2579--2588}.
\bibitem[{Liu et~al.(2015)Liu, Shen, Lin and Reid}]{Liu2015}
\bibinfo{author}{Liu, F.}, \bibinfo{author}{Shen, C.}, \bibinfo{author}{Lin, G.}, \bibinfo{author}{Reid, I.}, \bibinfo{year}{2015}.
\newblock \bibinfo{title}{Learning depth from single monocular images using deep convolutional neural fields}.
\newblock \bibinfo{journal}{IEEE transactions on pattern analysis and machine intelligence} \bibinfo{volume}{38}, \bibinfo{pages}{2024--2039}.
\bibitem[{Liu et~al.(2020)Liu, Zhang, He, Liu, Wang, Tai, Luo, Wang, Li and Huang}]{Liu2020}
\bibinfo{author}{Liu, L.}, \bibinfo{author}{Zhang, J.}, \bibinfo{author}{He, R.}, \bibinfo{author}{Liu, Y.}, \bibinfo{author}{Wang, Y.}, \bibinfo{author}{Tai, Y.}, \bibinfo{author}{Luo, D.}, \bibinfo{author}{Wang, C.}, \bibinfo{author}{Li, J.}, \bibinfo{author}{Huang, F.}, \bibinfo{year}{2020}.
\newblock \bibinfo{title}{Learning by analogy: Reliable supervision from transformations for unsupervised optical flow estimation}, in: \bibinfo{booktitle}{Proceedings of the IEEE/CVF conference on computer vision and pattern recognition}, pp. \bibinfo{pages}{6489--6498}.
\bibitem[{Liu et~al.(2014)Liu, Salzmann and He}]{Liu2014}
\bibinfo{author}{Liu, M.}, \bibinfo{author}{Salzmann, M.}, \bibinfo{author}{He, X.}, \bibinfo{year}{2014}.
\newblock \bibinfo{title}{Discrete-continuous depth estimation from a single image}, in: \bibinfo{booktitle}{Proceedings of the IEEE Conference on Computer Vision and Pattern Recognition}, pp. \bibinfo{pages}{716--723}.
\bibitem[{Marsal et~al.(2023)Marsal, Chabot, Loesch and Sahbi}]{marsal2023brightflow}
\bibinfo{author}{Marsal, R.}, \bibinfo{author}{Chabot, F.}, \bibinfo{author}{Loesch, A.}, \bibinfo{author}{Sahbi, H.}, \bibinfo{year}{2023}.
\newblock \bibinfo{title}{Brightflow: Brightness-change-aware unsupervised learning of optical flow}, in: \bibinfo{booktitle}{Proceedings of the IEEE/CVF Winter Conference on Applications of Computer Vision}, pp. \bibinfo{pages}{2061--2070}.
\bibitem[{Meister et~al.(2018)Meister, Hur and Roth}]{Meister2018}
\bibinfo{author}{Meister, S.}, \bibinfo{author}{Hur, J.}, \bibinfo{author}{Roth, S.}, \bibinfo{year}{2018}.
\newblock \bibinfo{title}{Unflow: Unsupervised learning of optical flow with a bidirectional census loss}, in: \bibinfo{booktitle}{Proceedings of the AAAI conference on artificial intelligence}.
\bibitem[{Ning et~al.(2023)Ning, Li, Zhang, Wang, Geng, Dai, He and Hu}]{ning2023all}
\bibinfo{author}{Ning, J.}, \bibinfo{author}{Li, C.}, \bibinfo{author}{Zhang, Z.}, \bibinfo{author}{Wang, C.}, \bibinfo{author}{Geng, Z.}, \bibinfo{author}{Dai, Q.}, \bibinfo{author}{He, K.}, \bibinfo{author}{Hu, H.}, \bibinfo{year}{2023}.
\newblock \bibinfo{title}{All in tokens: Unifying output space of visual tasks via soft token}, in: \bibinfo{booktitle}{Proceedings of the IEEE/CVF International Conference on Computer Vision}, pp. \bibinfo{pages}{19900--19910}.
\bibitem[{Patil et~al.(2022)Patil, Sakaridis, Liniger and Van~Gool}]{Patil2022}
\bibinfo{author}{Patil, V.}, \bibinfo{author}{Sakaridis, C.}, \bibinfo{author}{Liniger, A.}, \bibinfo{author}{Van~Gool, L.}, \bibinfo{year}{2022}.
\newblock \bibinfo{title}{P3depth: Monocular depth estimation with a piecewise planarity prior}, in: \bibinfo{booktitle}{Proceedings of the IEEE/CVF Conference on Computer Vision and Pattern Recognition}, pp. \bibinfo{pages}{1610--1621}.
\bibitem[{Piccinelli et~al.(2023)Piccinelli, Sakaridis and Yu}]{Piccinelli2023}
\bibinfo{author}{Piccinelli, L.}, \bibinfo{author}{Sakaridis, C.}, \bibinfo{author}{Yu, F.}, \bibinfo{year}{2023}.
\newblock \bibinfo{title}{idisc: Internal discretization for monocular depth estimation}, in: \bibinfo{booktitle}{Proceedings of the IEEE/CVF Conference on Computer Vision and Pattern Recognition}, pp. \bibinfo{pages}{21477--21487}.
\bibitem[{Ranftl et~al.(2021)Ranftl, Bochkovskiy and Koltun}]{Ranftl2021}
\bibinfo{author}{Ranftl, R.}, \bibinfo{author}{Bochkovskiy, A.}, \bibinfo{author}{Koltun, V.}, \bibinfo{year}{2021}.
\newblock \bibinfo{title}{Vision transformers for dense prediction}, in: \bibinfo{booktitle}{Proceedings of the IEEE/CVF International Conference on Computer Vision}, pp. \bibinfo{pages}{12179--12188}.
\bibitem[{Ranftl et~al.(2020)Ranftl, Lasinger, Hafner, Schindler and Koltun}]{Ranftl2020}
\bibinfo{author}{Ranftl, R.}, \bibinfo{author}{Lasinger, K.}, \bibinfo{author}{Hafner, D.}, \bibinfo{author}{Schindler, K.}, \bibinfo{author}{Koltun, V.}, \bibinfo{year}{2020}.
\newblock \bibinfo{title}{Towards robust monocular depth estimation: Mixing datasets for zero-shot cross-dataset transfer}.
\newblock \bibinfo{journal}{IEEE transactions on pattern analysis and intelligence, machine} \bibinfo{volume}{44}, \bibinfo{pages}{1623--1637}.
\bibitem[{Ren et~al.(2017)Ren, Yan, Ni, Liu, Yang and Zha}]{Ren2017}
\bibinfo{author}{Ren, Z.}, \bibinfo{author}{Yan, J.}, \bibinfo{author}{Ni, B.}, \bibinfo{author}{Liu, B.}, \bibinfo{author}{Yang, X.}, \bibinfo{author}{Zha, H.}, \bibinfo{year}{2017}.
\newblock \bibinfo{title}{Unsupervised deep learning for optical flow estimation}, in: \bibinfo{booktitle}{Proceedings of the AAAI conference on artificial intelligence}.
\bibitem[{Saxena et~al.(2005)Saxena, Chung and Ng}]{Saxena2005}
\bibinfo{author}{Saxena, A.}, \bibinfo{author}{Chung, S.}, \bibinfo{author}{Ng, A.}, \bibinfo{year}{2005}.
\newblock \bibinfo{title}{Learning depth from single monocular images}.
\newblock \bibinfo{journal}{Advances in neural information processing systems} \bibinfo{volume}{18}.
\bibitem[{Schonberger and Frahm(2016)}]{schonberger2016colmap}
\bibinfo{author}{Schonberger, J.L.}, \bibinfo{author}{Frahm, J.M.}, \bibinfo{year}{2016}.
\newblock \bibinfo{title}{Structure-from-motion revisited}, in: \bibinfo{booktitle}{Proceedings of the IEEE conference on computer vision and pattern recognition}, pp. \bibinfo{pages}{4104--4113}.
\bibitem[{Shao et~al.(2024a)Shao, Pei, Chen, Li, Liu and Li}]{shao2023urcdc}
\bibinfo{author}{Shao, S.}, \bibinfo{author}{Pei, Z.}, \bibinfo{author}{Chen, W.}, \bibinfo{author}{Li, R.}, \bibinfo{author}{Liu, Z.}, \bibinfo{author}{Li, Z.}, \bibinfo{year}{2024}a.
\newblock \bibinfo{title}{Urcdc-depth: Uncertainty rectified cross-distillation with cutflip for monocular depth estimation}.
\newblock \bibinfo{journal}{IEEE Transactions on Multimedia} \bibinfo{volume}{26}, \bibinfo{pages}{3341--3353}.
\bibitem[{Shao et~al.(2024b)Shao, Pei, Wu, Liu, Chen and Li}]{shao2024iebins}
\bibinfo{author}{Shao, S.}, \bibinfo{author}{Pei, Z.}, \bibinfo{author}{Wu, X.}, \bibinfo{author}{Liu, Z.}, \bibinfo{author}{Chen, W.}, \bibinfo{author}{Li, Z.}, \bibinfo{year}{2024}b.
\newblock \bibinfo{title}{Iebins: Iterative elastic bins for monocular depth estimation}.
\newblock \bibinfo{journal}{Advances in Neural Information Processing Systems} \bibinfo{volume}{36}.
\bibitem[{Shotton et~al.(2013)Shotton, Glocker, Zach, Izadi, Criminisi and Fitzgibbon}]{Shotton2013}
\bibinfo{author}{Shotton, J.}, \bibinfo{author}{Glocker, B.}, \bibinfo{author}{Zach, C.}, \bibinfo{author}{Izadi, S.}, \bibinfo{author}{Criminisi, A.}, \bibinfo{author}{Fitzgibbon, A.}, \bibinfo{year}{2013}.
\newblock \bibinfo{title}{Scene coordinate regression forests for camera relocalization in rgb-d images}, in: \bibinfo{booktitle}{Proceedings of the IEEE conference on computer vision and pattern recognition}, pp. \bibinfo{pages}{2930--2937}.
\bibitem[{Silberman et~al.(2012)Silberman, Hoiem, Kohli and Fergus}]{Silberman2012}
\bibinfo{author}{Silberman, N.}, \bibinfo{author}{Hoiem, D.}, \bibinfo{author}{Kohli, P.}, \bibinfo{author}{Fergus, R.}, \bibinfo{year}{2012}.
\newblock \bibinfo{title}{Indoor segmentation and support inference from rgbd images}.
\newblock \bibinfo{journal}{ECCV} \bibinfo{volume}{7576}, \bibinfo{pages}{746--760}.
\bibitem[{Song et~al.(2023)Song, Hu, Liang, Shi, Xie, Lu and Hei}]{Song2023}
\bibinfo{author}{Song, X.}, \bibinfo{author}{Hu, H.}, \bibinfo{author}{Liang, L.}, \bibinfo{author}{Shi, W.}, \bibinfo{author}{Xie, G.}, \bibinfo{author}{Lu, X.}, \bibinfo{author}{Hei, X.}, \bibinfo{year}{2023}.
\newblock \bibinfo{title}{Unsupervised monocular estimation of depth and visual odometry using attention and depth-pose consistency loss}.
\newblock \bibinfo{journal}{IEEE Transactions on Multimedia} \bibinfo{volume}{26}, \bibinfo{pages}{3517--3529}.
\bibitem[{Sun et~al.(2018)Sun, Yang, Liu and Kautz}]{Sun2018}
\bibinfo{author}{Sun, D.}, \bibinfo{author}{Yang, X.}, \bibinfo{author}{Liu, M.Y.}, \bibinfo{author}{Kautz, J.}, \bibinfo{year}{2018}.
\newblock \bibinfo{title}{Pwc-net: Cnns for optical flow using pyramid, warping, and cost volume}, in: \bibinfo{booktitle}{Proceedings of the IEEE conference on computer vision and pattern recognition}, pp. \bibinfo{pages}{8934--8943}.
\bibitem[{Uhrig et~al.(2017)Uhrig, Schneider, Schneider, Franke, Brox and Geiger}]{Uhrig2017}
\bibinfo{author}{Uhrig, J.}, \bibinfo{author}{Schneider, N.}, \bibinfo{author}{Schneider, L.}, \bibinfo{author}{Franke, U.}, \bibinfo{author}{Brox, T.}, \bibinfo{author}{Geiger, A.}, \bibinfo{year}{2017}.
\newblock \bibinfo{title}{Sparsity invariant cnns}, in: \bibinfo{booktitle}{2017 international conference on 3D Vision (3DV)}, \bibinfo{publisher}{IEEE}. pp. \bibinfo{pages}{11--20}.
\bibitem[{Vaswani et~al.(2017)Vaswani, Shazeer, Parmar, Uszkoreit, Jones, Gomez, Kaiser and Polosukhin}]{Vaswani2017}
\bibinfo{author}{Vaswani, A.}, \bibinfo{author}{Shazeer, N.}, \bibinfo{author}{Parmar, N.}, \bibinfo{author}{Uszkoreit, J.}, \bibinfo{author}{Jones, L.}, \bibinfo{author}{Gomez, A.N.}, \bibinfo{author}{Kaiser, {\L}.}, \bibinfo{author}{Polosukhin, I.}, \bibinfo{year}{2017}.
\newblock \bibinfo{title}{Attention is all you need}.
\newblock \bibinfo{journal}{Advances in neural information processing systems} \bibinfo{volume}{30}.
\bibitem[{Wang et~al.(2018a)Wang, Buenaposada, Zhu and Lucey}]{Wang2018_RN51}
\bibinfo{author}{Wang, C.}, \bibinfo{author}{Buenaposada, J.M.}, \bibinfo{author}{Zhu, R.}, \bibinfo{author}{Lucey, S.}, \bibinfo{year}{2018}a.
\newblock \bibinfo{title}{Learning depth from monocular videos using direct methods}, in: \bibinfo{booktitle}{Proceedings of the IEEE conference on computer vision and pattern recognition}, pp. \bibinfo{pages}{2022--2030}.
\bibitem[{Wang et~al.(2018b)Wang, Yang, Yang, Zhao, Wang and Xu}]{Wang2018_RN94}
\bibinfo{author}{Wang, Y.}, \bibinfo{author}{Yang, Y.}, \bibinfo{author}{Yang, Z.}, \bibinfo{author}{Zhao, L.}, \bibinfo{author}{Wang, P.}, \bibinfo{author}{Xu, W.}, \bibinfo{year}{2018}b.
\newblock \bibinfo{title}{Occlusion aware unsupervised learning of optical flow}, in: \bibinfo{booktitle}{Proceedings of the IEEE conference on computer vision and pattern recognition}, pp. \bibinfo{pages}{4884--4893}.
\bibitem[{Wu et~al.(2022)Wu, Wang, Hall, Neumann and Su}]{Wu2022}
\bibinfo{author}{Wu, C.Y.}, \bibinfo{author}{Wang, J.}, \bibinfo{author}{Hall, M.}, \bibinfo{author}{Neumann, U.}, \bibinfo{author}{Su, S.}, \bibinfo{year}{2022}.
\newblock \bibinfo{title}{Toward practical monocular indoor depth estimation}, in: \bibinfo{booktitle}{Proceedings of the IEEE/CVF Conference on Computer Vision and Pattern Recognition}, pp. \bibinfo{pages}{3814--3824}.
\bibitem[{Wulff et~al.(2012)Wulff, Butler, Stanley and Black}]{Wulff2012}
\bibinfo{author}{Wulff, J.}, \bibinfo{author}{Butler, D.J.}, \bibinfo{author}{Stanley, G.B.}, \bibinfo{author}{Black, M.J.}, \bibinfo{year}{2012}.
\newblock \bibinfo{title}{Lessons and insights from creating a synthetic optical flow benchmark}, in: \bibinfo{booktitle}{Computer Vision–ECCV 2012. Workshops and Demonstrations: Florence, Italy, October 7-13, 2012, Proceedings, Part II 12}, \bibinfo{publisher}{Springer}. pp. \bibinfo{pages}{168--177}.
\bibitem[{Xie et~al.(2016)Xie, Girshick and Farhadi}]{Xie2016}
\bibinfo{author}{Xie, J.}, \bibinfo{author}{Girshick, R.}, \bibinfo{author}{Farhadi, A.}, \bibinfo{year}{2016}.
\newblock \bibinfo{title}{Deep3d: Fully automatic 2d-to-3d video conversion with deep convolutional neural networks}, in: \bibinfo{booktitle}{Computer Vision–ECCV 2016: 14th European Conference, Amsterdam, The Netherlands, October 11–14, 2016, Proceedings, Part IV 14}, \bibinfo{publisher}{Springer}. pp. \bibinfo{pages}{842--857}.
\bibitem[{Xu et~al.(2023)Xu, Zhang, Cai, Rezatofighi, Yu, Tao and Geiger}]{Xu2023}
\bibinfo{author}{Xu, H.}, \bibinfo{author}{Zhang, J.}, \bibinfo{author}{Cai, J.}, \bibinfo{author}{Rezatofighi, H.}, \bibinfo{author}{Yu, F.}, \bibinfo{author}{Tao, D.}, \bibinfo{author}{Geiger, A.}, \bibinfo{year}{2023}.
\newblock \bibinfo{title}{Unifying flow, stereo and depth estimation}.
\newblock \bibinfo{journal}{IEEE Transactions on Pattern Analysis and Machine Intelligence} \bibinfo{volume}{45}, \bibinfo{pages}{13941--13958}.
\bibitem[{Yin et~al.(2019)Yin, Liu, Shen and Yan}]{Yin2019}
\bibinfo{author}{Yin, W.}, \bibinfo{author}{Liu, Y.}, \bibinfo{author}{Shen, C.}, \bibinfo{author}{Yan, Y.}, \bibinfo{year}{2019}.
\newblock \bibinfo{title}{Enforcing geometric constraints of virtual normal for depth prediction}, in: \bibinfo{booktitle}{Proceedings of the IEEE/CVF International Conference on Computer Vision}, pp. \bibinfo{pages}{5684--5693}.
\bibitem[{Yin and Shi(2018)}]{Yin2018}
\bibinfo{author}{Yin, Z.}, \bibinfo{author}{Shi, J.}, \bibinfo{year}{2018}.
\newblock \bibinfo{title}{Geonet: Unsupervised learning of dense depth, optical flow and camera pose}, in: \bibinfo{booktitle}{Proceedings of the IEEE conference on computer vision and pattern recognition}, pp. \bibinfo{pages}{1983--1992}.
\bibitem[{Yu et~al.(2020)Yu, Jin and Gao}]{Yu2020}
\bibinfo{author}{Yu, Z.}, \bibinfo{author}{Jin, L.}, \bibinfo{author}{Gao, S.}, \bibinfo{year}{2020}.
\newblock \bibinfo{title}{$\mathrm{P^2Net}$: Patch-match and plane-regularization for unsupervised indoor depth estimation}, in: \bibinfo{booktitle}{Computer Vision–ECCV 2020: 16th European Conference, Glasgow, UK, August 23–28, 2020, Proceedings, Part XXIV}, \bibinfo{publisher}{Springer}. pp. \bibinfo{pages}{206--222}.
\bibitem[{Yu et~al.(2019)Yu, Zheng, Lian, Zhou and Gao}]{Yu2019}
\bibinfo{author}{Yu, Z.}, \bibinfo{author}{Zheng, J.}, \bibinfo{author}{Lian, D.}, \bibinfo{author}{Zhou, Z.}, \bibinfo{author}{Gao, S.}, \bibinfo{year}{2019}.
\newblock \bibinfo{title}{Single-image piece-wise planar 3d reconstruction via associative embedding}, in: \bibinfo{booktitle}{Proceedings of the IEEE/CVF Conference on Computer Vision and Pattern Recognition}, pp. \bibinfo{pages}{1029--1037}.
\bibitem[{Yuan et~al.(2022)Yuan, Gu, Dai, Zhu and Tan}]{Yuan2022}
\bibinfo{author}{Yuan, W.}, \bibinfo{author}{Gu, X.}, \bibinfo{author}{Dai, Z.}, \bibinfo{author}{Zhu, S.}, \bibinfo{author}{Tan, P.}, \bibinfo{year}{2022}.
\newblock \bibinfo{title}{New crfs: Neural window fully-connected crfs for monocular depth estimation}.
\newblock \bibinfo{journal}{arXiv preprint arXiv:.01502} .
\bibitem[{Zhan et~al.(2018)Zhan, Garg, Weerasekera, Li, Agarwal and Reid}]{Zhan2018}
\bibinfo{author}{Zhan, H.}, \bibinfo{author}{Garg, R.}, \bibinfo{author}{Weerasekera, C.S.}, \bibinfo{author}{Li, K.}, \bibinfo{author}{Agarwal, H.}, \bibinfo{author}{Reid, I.}, \bibinfo{year}{2018}.
\newblock \bibinfo{title}{Unsupervised learning of monocular depth estimation and visual odometry with deep feature reconstruction}, in: \bibinfo{booktitle}{Proceedings of the IEEE conference on computer vision and pattern recognition}, pp. \bibinfo{pages}{340--349}.
\bibitem[{Zhang et~al.(2023)Zhang, Yang, Mi, Zheng and Yao}]{Zhang2023}
\bibinfo{author}{Zhang, S.}, \bibinfo{author}{Yang, L.}, \bibinfo{author}{Mi, M.B.}, \bibinfo{author}{Zheng, X.}, \bibinfo{author}{Yao, A.}, \bibinfo{year}{2023}.
\newblock \bibinfo{title}{Improving deep regression with ordinal entropy}.
\newblock \bibinfo{journal}{arXiv preprint arXiv:.08915} .
\bibitem[{Zhang et~al.(2022)Zhang, Gong, Li, Zhang, Jiang and Zhao}]{Zhang2022}
\bibinfo{author}{Zhang, Y.}, \bibinfo{author}{Gong, M.}, \bibinfo{author}{Li, J.}, \bibinfo{author}{Zhang, M.}, \bibinfo{author}{Jiang, F.}, \bibinfo{author}{Zhao, H.}, \bibinfo{year}{2022}.
\newblock \bibinfo{title}{Self-supervised monocular depth estimation with multiscale perception}.
\newblock \bibinfo{journal}{IEEE transactions on image processing} \bibinfo{volume}{31}, \bibinfo{pages}{3251--3266}.
\bibitem[{Zhao et~al.(2023)Zhao, Poggi, Tosi, Zhou, Sun, Tang and Mattoccia}]{Zhao2023}
\bibinfo{author}{Zhao, C.}, \bibinfo{author}{Poggi, M.}, \bibinfo{author}{Tosi, F.}, \bibinfo{author}{Zhou, L.}, \bibinfo{author}{Sun, Q.}, \bibinfo{author}{Tang, Y.}, \bibinfo{author}{Mattoccia, S.}, \bibinfo{year}{2023}.
\newblock \bibinfo{title}{Gasmono: Geometry-aided self-supervised monocular depth estimation for indoor scenes}, in: \bibinfo{booktitle}{Proceedings of the IEEE/CVF International Conference on Computer Vision}, pp. \bibinfo{pages}{16209--16220}.
\bibitem[{Zhao et~al.(2020)Zhao, Liu, Shu and Liu}]{Zhao2020}
\bibinfo{author}{Zhao, W.}, \bibinfo{author}{Liu, S.}, \bibinfo{author}{Shu, Y.}, \bibinfo{author}{Liu, Y.J.}, \bibinfo{year}{2020}.
\newblock \bibinfo{title}{Towards better generalization: Joint depth-pose learning without posenet}, in: \bibinfo{booktitle}{Proceedings of the IEEE/CVF Conference on Computer Vision and Pattern Recognition}, pp. \bibinfo{pages}{9151--9161}.
\bibitem[{Zhong et~al.(2019)Zhong, Ji, Wang, Dai and Li}]{Zhong2019}
\bibinfo{author}{Zhong, Y.}, \bibinfo{author}{Ji, P.}, \bibinfo{author}{Wang, J.}, \bibinfo{author}{Dai, Y.}, \bibinfo{author}{Li, H.}, \bibinfo{year}{2019}.
\newblock \bibinfo{title}{Unsupervised deep epipolar flow for stationary or dynamic scenes}, in: \bibinfo{booktitle}{Proceedings of the IEEE/CVF conference on computer vision and pattern recognition}, pp. \bibinfo{pages}{12095--12104}.
\bibitem[{Zhou et~al.(2019)Zhou, Wang, Qin and Zeng}]{Zhou2019}
\bibinfo{author}{Zhou, J.}, \bibinfo{author}{Wang, Y.}, \bibinfo{author}{Qin, K.}, \bibinfo{author}{Zeng, W.}, \bibinfo{year}{2019}.
\newblock \bibinfo{title}{Moving indoor: Unsupervised video depth learning in challenging environments}, in: \bibinfo{booktitle}{Proceedings of the IEEE/CVF International Conference on Computer Vision}, pp. \bibinfo{pages}{8618--8627}.
\bibitem[{Zhou et~al.(2017)Zhou, Brown, Snavely and Lowe}]{Zhou2017}
\bibinfo{author}{Zhou, T.}, \bibinfo{author}{Brown, M.}, \bibinfo{author}{Snavely, N.}, \bibinfo{author}{Lowe, D.G.}, \bibinfo{year}{2017}.
\newblock \bibinfo{title}{Unsupervised learning of depth and ego-motion from video}, in: \bibinfo{booktitle}{Proceedings of the IEEE conference on computer vision and pattern recognition}, pp. \bibinfo{pages}{1851--1858}.
\bibitem[{Zou et~al.(2018)Zou, Luo and Huang}]{Zou2018}
\bibinfo{author}{Zou, Y.}, \bibinfo{author}{Luo, Z.}, \bibinfo{author}{Huang, J.B.}, \bibinfo{year}{2018}.
\newblock \bibinfo{title}{Df-net: Unsupervised joint learning of depth and flow using cross-task consistency}, in: \bibinfo{booktitle}{Proceedings of the European conference on computer vision (ECCV)}, pp. \bibinfo{pages}{36--53}.

\end{thebibliography}

\end{document}